\documentclass[10pt,journal,compsoc]{IEEEtran}
\pdfoutput=1
\usepackage[utf8]{inputenc}
\usepackage{amsmath,graphicx}
\usepackage{epstopdf}
\usepackage{amsfonts,amsthm}
\usepackage{algorithm,algorithmic}
\usepackage{setspace}
\usepackage{cite}
\usepackage[colorlinks,linkcolor=blue,anchorcolor=blue,citecolor=blue]{hyperref}
\usepackage{subfig}
\usepackage{subfloat}
\usepackage{multirow,booktabs}
\usepackage{fontenc}

\usepackage{amsmath,amssymb,mathrsfs}
\usepackage[misc]{ifsym}
\usepackage{url}
\usepackage{bm}
\usepackage{bbding}
\usepackage{mathtools}
\usepackage{makecell}

\usepackage{hyperref}
\hypersetup{hypertex=true,
	colorlinks=true,
	linkcolor=red,
	anchorcolor=green,
	citecolor=blue}

\newtheorem{Proposition}{Proposition}

\newtheorem{definition}{Definition}
\newtheorem{theorem}{Theorem}

\newtheorem{remark}{Remark}

\usepackage{array}
\usepackage{cases}
\newcolumntype{"}{@{\hskip\tabcolsep\vrule width 1pt\hskip\tabcolsep}}
\usepackage{pifont}
\usepackage{color}

\if CLASSOPTIONcompsoc
  \usepackage[nocompress]{cite}
\else
  \usepackage{cite}
\fi
\ifCLASSINFOpdf
\else
\fi

\begin{document}

%
\title{Guaranteed Tensor Recovery Fused Low-rankness and Smoothness}
%
%
%
%

\author{
Hailin~Wang,~Jiangjun~Peng,~Wenjin~Qin,~Jianjun~Wang~and~Deyu~Meng,~\IEEEmembership{Member,~IEEE}
\thanks{H. Wang, J. Peng (co-first author) are with the School of Mathematics and Statistics, Xi'an Jiaotong University, Xi'an 710049, Shanxi, China (email: wanghailin97@163.com, andrew.pengjj@gmail.com).} \thanks{W. Qin and J. Wang are with the School of Mathematics and Statistics, Southwest University, Chongqing, 400715, China (email: qinwenjin2021@163.com, wjj@swu.edu.cn).}
\thanks{D. Meng (corresponding author) is with the School of Mathematics and Statistics and Ministry of Education Key Lab of Intelligent Networks and Network Security, Xi'an Jiaotong University, Xian, Shanxi, China and Macau Institute of Systems Engineering, Macau University of Science and Technology, Macau, China. (email: dymeng@mail.xjtu.edu.cn).}
}

\markboth{Journal of \LaTeX\ Class Files,~Vol.~14, No.~8, August~2015}%
{Shell \MakeLowercase{\textit{et al.}}: Bare Demo of IEEEtran.cls for Computer Society Journals}
%



\IEEEtitleabstractindextext{%
\begin{abstract}
Vast visual data like multi-spectral images and multi-frame videos are essentially with the tensor format. However, due to the defects of signal acquisition equipments, the practically collected tensor data are always with evident degradations like corruptions or missing values. The tensor data recovery task has thus attracted much research attention in recent years. Solving such an ill-posed problem generally requires to explore intrinsic prior structures underlying tensor data, and formulate them as certain forms of regularization terms for guiding a sound estimate of the restored tensor. Recent research have made significant progress by adopting two insightful tensor priors, i.e., global low-rankness (\textbf{L}) and local smoothness (\textbf{S}) across different tensor modes, which are always encoded as a sum of two separate regularization terms into the recovery models. However, unlike the primary theoretical developments on low-rank tensor recovery, these joint ``\textbf{L}+\textbf{S}'' models have no theoretical exact-recovery guarantees yet, making the methods lack reliability in real practice.
To this crucial issue, in this work, we build a unique regularization term, which essentially encodes both \textbf{L} and \textbf{S} priors of a tensor simultaneously. Especially, by equipping this single regularizer into the recovery models, we can rigorously prove the exact recovery guarantees for two typical tensor recovery tasks, i.e., tensor completion (TC) and tensor robust principal component analysis (TRPCA). To the best of our knowledge, this should be the first exact-recovery results among all related ``\textbf{L}+\textbf{S}'' methods for tensor recovery. We further propose ADMM algorithms for solving the proposed models, and prove their fine convergence properties. Significant recovery accuracy improvements over many other SOTA methods in several TC and TRPCA tasks with various kinds of visual tensor data are observed in extensive experiments. Typically, our method achieves a workable performance when the missing rate is extremely large, e.g., $99.5\%$, for the color image inpainting task, while all its peers totally fail in such challenging case. Source code is released at \url{https://github.com/wanghailin97}.
\end{abstract}

}

\maketitle

\IEEEdisplaynontitleabstractindextext

%
\IEEEpeerreviewmaketitle

\IEEEraisesectionheading{\section{Introduction}\label{sec:introduction}}

%
%
%
%

\IEEEPARstart{T}{ensors}, or multidimensional arrays, are the natural representation format of a wide range of real-world data, e.g., multi-frame/spectral/view data, network flow data, etc. Compared with representation in vector/matrix structure, tensor tends to more faithfully and accurately deliver intrinsic multidimensional structural information underlying data, and thus show more potential usefulness in the recent years across various multifarious fields, such as statistics \cite{mccullagh2018tensor}, signal processing \cite{cichocki2015tensor}, data mining \cite{papalexakis2016tensors}, machine learning \cite{signoretto2014learning} and computer vision \cite{panagakis2021tensor}.

However, due to the defects of signal acquisition equipments, such as sensor sensitivity, photon effects and calibration error, tensor data collected in real world are always with evident degradations like corruptions or missing values. Tensor recovery has thus become one of the fundamental problems in tensor research. Mathematically speaking, this is a typical inverse problem that aims to recover an unknown tensor $\mathcal{T}\in\mathbb{R}^{n_1\times n_2\times \cdots \times n_d}$ with certain structural priori assumptions from the observation $\mathcal{Y} = \Phi(\mathcal{T})$, where $\Phi(\cdot)$ is the operator modeling certain degradation process. Two typical degradations are information loss and noise disturbance, corresponding to two common tensor recovery tasks, \textit{tensor completion} (TC) \cite{liu2012tensor} and \textit{tensor robust component principal analysis} (TRPCA) \cite{huang2015provable}, respectively.

Solving such an ill-posed problem generally depends on characterizing intrinsic prior structures underlying tensor data, and encoding them as certain regularization terms for guiding a sound estimate of the restored tensor. One of the most employed priors is the \textit{low-rankness} (denoted as ``\textbf{L}''). This structural prior considers that a tensor resides in a proper low-dimensional subspace over the entire range of its certain tensor mode, revealing its information correlation at a global scale along this tensor mode. This prior then leads to the following low-rank tensor recovery model,
\begin{equation}\label{eq.1}
\min_{\mathcal{T}} \mathfrak{R}(\mathcal{T}) \ \ \text{s.t.} \ \ \mathcal{Y} = \Phi(\mathcal{T}),
\end{equation}
where $\mathfrak{R}(\cdot)$ denotes the regularizer measuring tensor low-rankness. Different from that defined on matrices, there are various notions of tensor rank defined on the basis of different tensor decompositions. The most classic ones contain CP \cite{hitchcock1927expression}, Tucker \cite{tucker1963implications}, and HOSVD \cite{de2000multilinear}. In the last few years, several new low-rank tensor approximation frameworks were proposed, such as tensor train \cite{oseledets2011tensor-train}, tensor ring \cite{zhao2016tensor}, and \textit{tensor singular value decomposition} (t-SVD) \cite{kilmer2011factorization}. Among them, t-SVD presents the first closed multiplicative operation, named t-product, on tensor rank, and establishes a complete tensor decomposition algebraic framework. Especially, the recent work \cite{kilmer2021tensor} revolutionarily proved the optimal representation and compression theories of t-SVD, making it more notable in characterizing the intrinsic low-rank structure underlying tensors. Therefore, the model (\ref{eq.1}) under t-SVD has captured lots of interest recently \cite{zhang2016exact,lu2019tensor,zhang2020low,hou2021robust,zhang2021low}. Especially, in recent works \cite{huang2015provable,zhang2016exact,lu2019tensor,zhang2020low}, accurate recovery theory has been proved to guarantee such \textbf{L}-prior model able to achieve an exact recovery of an original true tensor, further validating its reliability for tensors purely possessing such structural prior.

Besides the low-rankness prior, the \textit{smoothness} (denoted as ``\textbf{S}'') prior is also widely adopted in tensor recovery tasks. This prior reflects a general structural property of a practical visual tensor. The adjacent pixels along  a tensor mode tend to be changed continuously, representing a type of information similarity at a relatively local scale. To guarantee a good performance, most related works incorporate such \textbf{S} prior into the \textbf{L}-prior model, and then use the following model with two regularization items for tensor recovery:
\begin{equation}\label{eq.2}
\min_{\mathcal{T}} \mathfrak{R}(\mathcal{T})+\alpha\mathfrak{S}(\mathcal{T}) \ \ \text{s.t.} \ \ \mathcal{Y} = \Phi(\mathcal{T}),
\end{equation}
where $\mathfrak{S}(\cdot)$ denotes the regularizer for measuring the \textbf{S} prior, and $\alpha>0$ is a balancing parameter. In the recent years, a series of works \cite{ji2016tensor,li2017low,he2015total,yokota2016smooth,wang2017hyperspectral,chen2018tensor,zhang2019hyperspectral,ko2020fast,qiu2021robust} have emerged with the form of (\ref{eq.2}) applied in visual tensor data restoration tasks, and achieve excellent performance beyond purely \textbf{L}-prior methods. This reflects the universality of such mixture priors possessed by visual tensor data.

Although the recovery performance have a pleasing lifting by adopting the \textbf{L}+\textbf{S}-prior model (\ref{eq.2}), it has not succeeded in a fundamental aspect of tensor recovery from the \textbf{L}-prior model. That is, the theoretical exact recovery guarantee has still not been proved for the related methods. Comparatively, as aforementioned, the pure \textbf{L}-prior model (\ref{eq.1}) is provable with exact recovery \cite{zhang2016exact,lu2019tensor,zhang2020low}. This issue of theory vacancy inclines to make the related methods lack reliability in applications. Besides, the performance of (\ref{eq.2}) are highly affected by the trade-off parameter imposed between \textbf{L} and \textbf{S} regularizers. For a practically collected visual tensor, the \textbf{L} and \textbf{S} priors are usually coupled with each other, each representing a type of information redundancy property. This always makes it fairly difficult to build a general rule for finely tuning this balancing parameter in real scenarios.

Against these issues, in this study we propose a theoretically exact-recovery guaranteed method without balancing parameter for joint low-rank and smooth tensor recovery. The main contributions can be summarized as follows.

1. Under the high-order t-SVD framework \cite{qin2022low}, a new regularizer named \textit{tensor correlated total variation} (or t-CTV briefly) is introduced to characterize both \textbf{L}+\textbf{S} priors of a tensor with a unique term. The underlying mathematical principles indicate that this single regularizer can promote the two priors simultaneously.

2. The t-CTV is applied to two typical tensor recovery tasks, i.e., TC and TRPCA. With some mild assumptions of tensor incoherence conditions, the exact recovery theories can be proved for the conducted t-CTV-TC and t-CTV-TRPCA models. To the best of our knowledge, these should be the first theoretical exact-recovery guarantees among all related joint \textbf{L}+\textbf{S} tensor recovery studies. Furthermore, we obtain a lower bound of sampling complexity in an interpretable manner that manifests our t-CTV based model possessing more powerful recovery ability than classic models purely considering \textbf{L} and \textbf{S}, as well as \textbf{L}+\textbf{S} priors.

3. Efficient algorithms based on ADMM are designed for solving the corresponding proposed models, with closed-form updating equation for each involved variable. Besides, the computational complexity and convergence of the proposed algorithms are well analyzed.

\begin{figure}[t]
\centering
\includegraphics[width=1\linewidth]{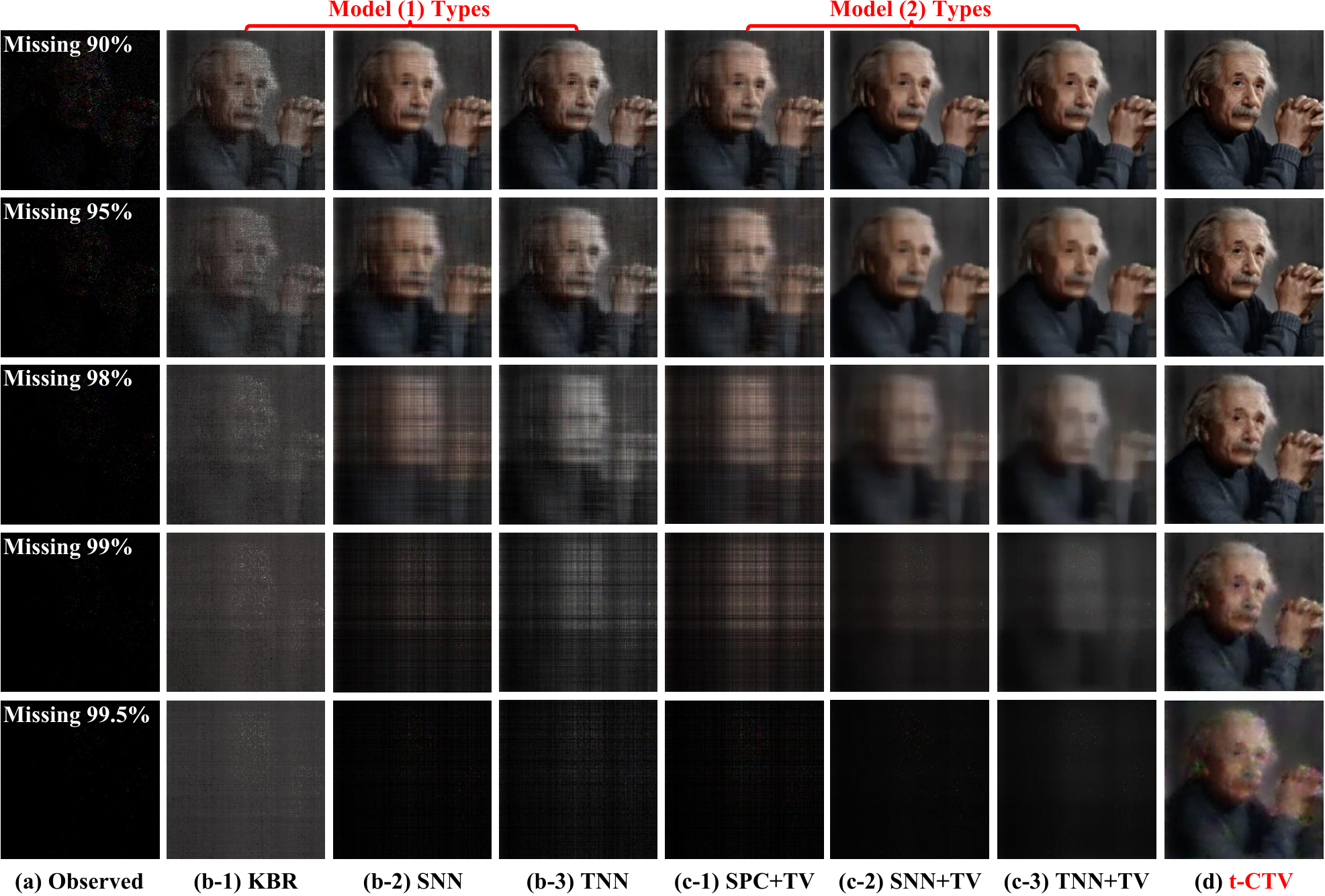}
\vspace{-0.6cm}
\caption{Illustration of recovery performance of all competing methods in color image inpainting. (a) Observed ``Einstein'' image with $90\%, 95\%, 98\%, 99\%$ and $99.5\%$ missing pixels (from up to bottom); (b) Recovery obtained by several SOTA methods built under model (\ref{eq.1}): (b-1) KBR\cite{xie2017kronecker}, (b-2) SNN\cite{liu2012tensor}, (b-3) TNN\cite{qin2022low};
(c) Recovery obtained by several SOTA methods built under model (\ref{eq.2}): (c-1) SPC+TV\cite{yokota2016smooth}, (c-2) SNN+TV\cite{li2017low}, (c-3)  TNN+TV\cite{qiu2021robust};
(d) Recovery by our method. It can be seen that our method can still work while all its peers largely failed in the last extreme missing case.}\label{fig.1}
\vspace{-0.2cm}
\end{figure}

4. The proposed t-CTV based algorithms are verified to be with an evidently stronger capacity in tensor recovery than the baseline \textbf{L}-prior model (\ref{eq.1}) and \textbf{L}+\textbf{S}-prior model (\ref{eq.2}) by comprehensive simulated experiments. Extensive applications in various visual tensor  recovery tasks demonstrate that our algorithms markedly enhances the recovery accuracy compared with many SOTA methods, including those considering \textbf{L}, \textbf{S} and \textbf{L}+\textbf{S}-priors.
Typically, our method achieves a significant improvement in color image inpainting task when missing-pixel rates are up to $90\%, 95\%$ and even $99.5\%$, as clearly depicted in Fig.\;\ref{fig.1}.

The remainder of this work is organized as follows. Section \ref{sec.2} reviews a series of related works. Section \ref{sec.3} introduces the fundamental high-order t-SVD framework, based on which the t-CTV regularizer is defined and analyzed in Section \ref{sec.4}. The main models and theories are given in Section \ref{sec.5}. Sections \ref{sec.6} and \ref{sec.7} present the algorithms and experiments, respectively. Finally, we conclude our work in Section \ref{sec.8}. All proof details are given in supplementary materials.

\section{Related Works}\label{sec.2}

\subsection{Low-rank tensor recovery}
We take TC and TRPCA, two of the most commonly studied tensor recovery tasks, as examples to introduce the related \textbf{L}-prior works. Specifically, TC refers to the problem of recovering a low-rank tensor $\mathcal{T}_0$ from its partial entries, i.e.,
\begin{equation}\label{eq.3}
\min_{\mathcal{T}} \mathfrak{R}(\mathcal{T}) \ \ \text{s.t.} \ \ \mathscr{P}_{\Omega}(\mathcal{T}) = \mathscr{P}_{\Omega}(\mathcal{T}_0),
\end{equation}
where $\Omega$ is the known elements' index and $\mathscr{P}(\cdot)$ is the projection operator. TRPCA aims to achieve recovery from grossly corrupted observations, modeled as:
\begin{equation}\label{eq.4}
\min_{\mathcal{T,E}} \mathfrak{R}(\mathcal{T})+\lambda\|\mathcal{E}\|_1 \ \ \text{s.t.} \ \ \mathcal{M} = \mathcal{T} + \mathcal{E},
\end{equation}
where $\mathcal{E}$ represents the noise component characterized by the $L_1$-norm $\|\cdot\|_1$ and $\lambda>0$ denotes the trade-off parameter.

Low-rank tensor recovery is originally developed from low-rank matrix recovery \cite{candes2009exact,candes2011robust}, but fraught with more challenges since the low-rank representation of a tensor is not unique. The early works are mainly established upon Tucker decomposition. Liu et al. \cite{liu2012tensor} first proposed the \textit{sum of nuclear norm} (SNN) and implemented it in TC tasks as an extension of matrix's nuclear norm method to tensors, which greatly promotes the development of tensor recovery in visual data processing. Later on, Goldfarb and Qin \cite{goldfarb2014robust} extended TRPCA problem in aspect of algorithm and Huang et al. \cite{huang2015provable} filled the gap in theory. Unfortunately, it's proved that SNN is not the tightest convex relaxation \cite{romera2013new}, thus leading to suboptimality of the solution \cite{mu2014square}. Several literatures attempted to ameliorate this issue. E.g., Mu et al. \cite{mu2014square} proposed a better convexification based on a balanced matricization strategy and proved a lower bound in sampling complexity. Xie et al. \cite{xie2017kronecker} further transformed the Tucker decomposition as a representation form of Kronecker basis and proposed the KBR tensor sparsity, resulting significant improvements both in TC and TRPCA tasks. Some other works are proposed with nonconvex relaxation and \textit{matrix factorization} (MF) on Tucker mode-wise matrices \cite{zhang2018nonconvex,xu2013parallel}. Except for the Tucker decomposition based research, in \cite{zhao2015bayesian1}, Zhao et al. proposed a fully Bayesian probabilistic CP factorization for TC problem, and several works further studied robust tensor CP decomposition problem \cite{anandkumar2016tensor,zhao2015bayesian2}. Besides, there also exist a series of works using certain type of tensor network decompositions \cite{oseledets2011tensor-train,zhao2016tensor}, see \cite{bengua2017efficient,rauhut2017low,wang2017efficient,zheng2021fully}.

Another class of works are based on t-SVD \cite{kilmer2011factorization}. It provides a more appropriate extension of matrix SVD. Compared with Tucker decomposition, t-SVD does not need matricization, and more importantly, equips the optimality property in low-rank representation as well as the matrix SVD \cite{kilmer2021tensor}. Formally, it decomposes a tensor into a tensor-tensor product of two orthogonal tensors and a f-diagonal tensor (also called as singular value tensor), and thus induces the t-SVD rank, also called tubal rank \cite{kilmer2013third} for third-order tensors, defined as the number of nonzero tubes of the singular value tensor. Based on that, a heuristic \textit{tensor nuclear norm} (TNN) as the convex surrogate of tubal rank is defined in \cite{semerci2014tensor}, and then Zhang and Aeron \cite{zhang2016exact} used it in TC tasks with theoretical guarantee. Then, a new form of TNN was proposed with the tightest convex envelope property and investigated in several tensor recovery problems \cite{lu2019tensor,lu2019low}. Recently, substantial investigations on t-SVD based tensor-related problems have been constructed, such as \cite{liu2019low,jiang2019robust,wang2020estimating,zhang2020low,wang2021generalized,hou2021robust}. Although these methods have attracted much attention, they cannot be applied to tensors of arbitrary order directly. Very recently, as an extension of the series of works by Kilmer et al. \cite{kilmer2011factorization,martin2013order,kernfeld2015tensor,kilmer2021tensor}, in \cite{qin2022low}, Qin et al. made a successful advancement with transform induced high-order t-SVD, exhibiting fine potentials in tensor recovery.

\subsection{Joint low-rank and smooth tensor recovery}
We then review related methods for jointly considering \textbf{L} and \textbf{S} priors for tensor recovery.

The \textbf{S}-prior structure is well possessed by a variety of visual tensor data, and generally modeled by \textit{total variation} (TV). It can be simply divided into anisotropic TV (TV-1) and isotropic TV (TV-2) cases, defined by the absolute distance and square distance on the difference of the neighbor elements, respectively \cite{rudin1992nonlinear}. In practice, researchers always design certain variations of TV form corresponding to the type of data. For a simple image, one often use spatial TV (STV) to characterize the piecewise smooth structure along its spatial dimensions \cite{chambolle1997image,beck2009fast,wang2008new}. For hyperspectral images (HSIs), there also exists smoothness along its spectral direction, thus leading to the spectral-spatial TV (SSTV) \cite{yuan2012hyperspectral,chang2015anisotropic}. Similarly, the temporal-spatial TV is also formulated for videos \cite{tom2020simultaneous}. For other TV variations, see \cite{valkonen2013total,holt2014total,duran2016collaborative}.

Recently, the TV approaches have been widely adopted in the visual tensor data recovery tasks, most of which are used via a joint model with low-rank and TV regularizers. Typically, Ji et al. \cite{ji2016tensor} investigated the TC problem by embedding the STV into the low-rank matrix factorization to all-mode matricization framework, While Li et al. \cite{li2017low} used the SNN to characterize the \textbf{L} prior, and the simple TV-1 on all-mode unfolding matrices for \textbf{S} prior. Other typical works along this line include \cite{yokota2016smooth,he2015total,wang2017hyperspectral,chen2018tensor,zhang2019hyperspectral,ko2020fast}, partially summarized in Table \ref{table.1}. These methods all employ a sum of \textbf{L} and \textbf{S} regularization terms, and have not proved exact-recovery results from a theoretical perspective. As a contrast, the basic tensor recovery models by purely using \textbf{L} regularizer can often establish the related theory guarantees, e.g., \cite{huang2015provable,zhang2016exact,lu2019tensor,zhang2020low,zhang2021low,hou2021robust,wang2021generalized}. By the way, it should be noted that there are several theoretical works related to smooth data recovery, e.g., \cite{needell2013near,needell2013stable,cai2015guarantees,li2019guaranteed}, but the corresponding modeling and analyzing are only considered in vector or matrix space which cannot be directly used for joint low-rank and smooth tensor recovery.

\begin{table}[t]
\centering
\setstretch{0.9}
\renewcommand{\arraystretch}{1.15}
\setlength\tabcolsep{5.0pt}
\small
\caption{Summary of some related works on tensor recovery with joint \textbf{L}+\textbf{S} priors}\label{table.1}
\vspace{-0.2cm}
\newcolumntype{"}{@{\hskip\tabcolsep\vrule width 1pt\hskip\tabcolsep}}
  \setlength{\abovecaptionskip}{5pt}
  \setlength{\belowcaptionskip}{5pt}
\begin{tabular}{l"c|c|c}
\Xhline{1pt}
Literature & Problem & Model & Theory \\
\Xhline{1pt}
Ji et al. \cite{ji2016tensor} & TC & MF + STV & \ding{56}\\
\hline
Li et al. \cite{li2017low} & TC & SNN + TV-1 & \ding{56}\\
\hline
Ko et al. \cite{ko2020fast} & TC & TT + TV-2 & \ding{56}\\
\hline
Yokota et al. \cite{yokota2016smooth} & TC & CP + TV-1/TV-2 & \ding{56}\\
\Xhline{1pt}
He et al. \cite{he2015total}  & TRPCA & MF + STV & \ding{56}\\
\hline
Wang et al. \cite{wang2017hyperspectral}  & TRPCA & SNN + SSTV & \ding{56}\\
\hline
Chen et al. \cite{chen2018tensor}  & TRPCA & TNN + HTV & \ding{56}\\
\hline
Zhang et al. \cite{zhang2019hyperspectral} & TRPCA & NLTRD + SSTV & \ding{56} \\
\Xhline{1pt}
\textbf{This work} & TC$\&$TRPCA & t-CTV & \ding{52}\\
\Xhline{1pt}
\end{tabular}
\vspace{-0.5cm}
\end{table}

Against the aforementioned issues, this work proposes a unique tensor recovery regularizer encoding both \textbf{L} and \textbf{S} priors simultaneously. It adopts a type of fused prior modeling manner under the advanced high-order t-SVD algebraic framework, wherein the considered TC and TRPCA models both have exact recoverability guarantees in theory beyond current \textbf{L}+\textbf{S} tensor recovery models. The concise form of this specifically designed regularizer also naturally helps get rid of the difficulty of tuning the trade-off parameter imposed between two regularizers as conventional.

\section{High-order t-SVD Framework}\label{sec.3}
We first introduce high-order t-SVD framework briefly and provide its optimal representation theory in this section. Please refer to \cite{kilmer2011factorization,kilmer2021tensor,martin2013order,kernfeld2015tensor,qin2022low} for more details.

For an order-$d$ tensor $\mathcal{T}$ sized $n_1\times n_2\times\cdots\times n_d$, $\mathcal{T}(i_1,i_2,\cdots,i_d)$ denotes its $(i_1,i_2,\cdots,i_d)$-th element, and $\mathcal{T}(:,:,i_3,\cdots,i_d)$ is called as its $(i_3,\cdots,i_d)$-th face slice containing the first two modes which is also written as $\mathrm{T}^{(i_3,\cdots,i_d)}$. Then $\operatorname{bdiag}(\mathcal{T})$ sized $n_1\prod_{j=3}^d n_j\times n_2\prod_{j=3}^d n_j$ is the block diagonal matrix constructed by all face slices, which is also written as $\overline{\mathcal{T}}$.

For the t-SVD framework with invertible transform $\mathfrak{L}$ \cite{qin2022low}, a high-order tensor $\mathcal{T}$'s transform form is given by $\mathcal{T}_\mathfrak{L}:=\mathfrak{L}(\mathcal{T})=\mathcal{T}\times_3 \mathrm{U}_{n_3}\times_4\cdots\times_d \mathrm{U}_{n_d}$, where $\times_j$ denotes mode-$j$ product ($\mathcal{Y}= \mathcal{X}\times_j\mathrm{M}$ means each $\mathcal{Y}(\cdots,i_{j-1},:,i_{j+1},\cdots)$ equals $\mathrm{M}\cdot\mathcal{X}(\cdots,i_{j-1},:,i_{j+1},\cdots)$ )and $\mathrm{U}_{n_j}$ is transform matrix sized $n_j\times n_j$, $j=3,\cdots,d$, such as the \textit{discrete fourier transform} (DFT) and \textit{discrete cosine transform} (DCT) matrices. Its inverse operation is $\mathfrak{L}^{-1}(\mathcal{T}):=\mathcal{T}\times_3 \mathrm{U}_{n_3}^{-1}\times_4\cdots\times_d \mathrm{U}_{n_d}^{-1}$ and satisfies $\mathfrak{L}^{-1}(\mathfrak{L}(\mathcal{T}))=\mathcal{T}$. The transform matrices $\{\mathrm{U}_{n_j}\}_{j=3}^d$ of $\mathfrak{L}$ are assumed to satisfy
\begin{equation}\label{eq.4a}
(\mathrm{U}_{n_d}^*\otimes\cdots\otimes\mathrm{U}_{n_3}^*)\cdot(\mathrm{U}_{n_d}\otimes\cdots\otimes\mathrm{U}_{n_3})
=\ell\cdot I_{n_3\cdots n_d},
\end{equation}
where $(\cdot)^*$ is conjugate transpose, $\otimes$ denotes the Kronecker product, $I$ denotes unitary matrix and $\ell>0$ is a specific scale factor corresponding to the transform, e.g., $\ell = \prod_{j=3}^d n_j$ for DFT matrix $\mathrm{F}_{n_j}$ since $\mathrm{F}_{n_j}^*\mathrm{F}_{n_j} = n_j I_{n_j}$, and $\ell = 1$ for DCT matrix $\mathrm{C}_{n_j}$ since $\mathrm{C}_{n_j}^*\mathrm{C}_{n_j}=I_{n_j}$, $j=3,\cdots,d$.

\begin{definition}[tensor-tensor product\cite{qin2022low}]
For order-$d$ tensors $\mathcal{A}\in\mathbb{R}^{n_1\times l\times n_3\times\cdots\times n_d}$ and $\mathcal{B}\in\mathbb{R}^{l\times n_2\times n_3\times\cdots\times n_d}$, its transform $\mathfrak{L}$ based product is defined as
\begin{equation}\label{eq.5}
\mathcal{A}*_{\mathfrak{L}}\mathcal{B}=\mathfrak{L}^{-1}(\mathfrak{L}(\mathcal{A})\Delta\mathfrak{L}(\mathcal{B})),
\end{equation}
where $\Delta$ denotes the face-wise product ($\mathcal{Z}=\mathcal{X}\Delta\mathcal{Y}\Leftrightarrow \mathrm{Z}^{(i_3,\cdots,i_d)} =\mathrm{X}^{(i_3,\cdots,i_d)}\mathrm{Y}^{(i_3,\cdots,i_d)}$ for all face slices).
\end{definition}

The tensor-tensor product leads that $\mathcal{C}=\mathcal{A}*_{\mathfrak{L}}\mathcal{B}$ is equivalent to $\overline{\mathcal{C}_\mathfrak{L}}=\overline{\mathcal{A}_\mathfrak{L}}\cdot\overline{\mathcal{B}_\mathfrak{L}}$. This conducts a simple implement of (\ref{eq.5}) by $\prod_{j=3}^d{n_j}$ times pairwise matrix product in the transform domain and then run inverse transformation.

\begin{definition}[transpose \cite{qin2022low}]
For $\mathcal{T}\in\mathbb{R}^{n_1\times n_2\times n_3\times\cdots\times n_d}$, its transpose $\mathcal{T}^\mathrm{T}\in\mathbb{R}^{n_2\times n_1\times n_3\times\cdots\times n_d}$ satisfies that $\mathcal{T}^\mathrm{T}_\mathfrak{L}(:,:,i_3,\cdots,i_d)=\mathcal{T}_\mathfrak{L}(:,:,i_3,\cdots,i_d)^\mathrm{T}$ for all face slices.
\end{definition}

\begin{definition}[identity tensor\cite{qin2022low}]
An order-$d$ tensor $\mathcal{I}_n\in\mathbb{R}^{n\times n\times n_3\times\cdots\times  n_d}$ is called as identity tensor if it satisfies $\mathcal{I}_\mathfrak{L}(:,:,i_3,\cdots,i_d)=I_n$ for all face slices.
\end{definition}

\begin{definition}[orthogonal tensor\cite{qin2022low}]
An order-$d$ tensor $\mathcal{U}\in\mathbb{R}^{n\times n\times n_3\times\cdots\times  n_d}$ is orthogonal if $\mathcal{U}^\mathrm{T}*_\mathfrak{L}\mathcal{U}
=\mathcal{U}*_\mathfrak{L}\mathcal{U}^\mathrm{T}=\mathcal{I}_n$.
\end{definition}

\begin{definition}[f-diagonal tensor\cite{qin2022low}]
An order-$d$ tensor $\mathcal{T}\in\mathbb{R}^{n\times n\times n_3\times\cdots\times  n_d}$ is f-diagonal if all its face slices is diagonal.
\end{definition}

\begin{theorem}[t-SVD\cite{qin2022low}]
For any order-$d$ tensor $\mathcal{T}\in\mathbb{R}^{n_1\times n_2\times\cdots\times n_d}$, it can be decomposed as
\begin{equation}\label{eq.6}
\mathcal{T} = \mathcal{U}*_\mathfrak{L}\mathcal{S}*_\mathfrak{L}\mathcal{V}^\mathrm{T},
\end{equation}
where $\mathcal{U}\in\mathbb{R}^{n_1\times n_1\times\cdots\times n_d}$ and $\mathcal{V}\in\mathbb{R}^{n_2\times n_2\times\cdots\times n_d}$ are orthogonal, and $\mathcal{S}\in\mathbb{R}^{n_1\times n_2\times\cdots\times n_d}$ is a f-diagonal tensor.
\end{theorem}

Similar to the normal t-SVD with DFT transform \cite{kilmer2011factorization,martin2013order}, the above general invertible linear transforms induced t-SVD can be realized by performing SVD on each slice of $\mathcal{T}_\mathfrak{L}$ in the transform domain and then inverting the corresponding components back to the original domain. Besides, the t-SVD has a skinny form \cite{qin2022low} with the following t-SVD rank, and some concepts can be further established.

\begin{definition}[t-SVD rank \cite{qin2022low}]
For $\mathcal{T}\in\mathbb{R}^{n_1\times n_2\times\cdots\times n_d}$ with t-SVD $\mathcal{T} = \mathcal{U}*_\mathfrak{L}\mathcal{S}*_\mathfrak{L}\mathcal{V}^\mathrm{T}$, its t-SVD rank is defined as
\begin{equation}\label{eq.7}
\operatorname{rank}_{\operatorname{t-SVD}}(\mathcal{T}):=\sharp\{i: \mathcal{S}(i,i,:,\cdots,:)\neq\textbf{0}\},
\end{equation}
where $\sharp$ denotes the cardinality of a set.
\end{definition}

\begin{definition}[tensor nuclear norm\cite{qin2022low}]
For order-$d$ tensor $\mathcal{T}\in\mathbb{R}^{n_1\times n_2\times\cdots\times n_d}$ under the t-SVD framework with transform $\mathfrak{L}$, its tensor nuclear norm (TNN) is defined as
\begin{equation}\label{eq.8}
\|\mathcal{T}\|_{\circledast,\mathfrak{L}}:= \frac{1}{\ell}\sum_{i_3=1}^{n_3}\cdots\sum_{i_d=1}^{n_d}\|\mathcal{T}_\mathfrak{L}(:,:,i_3,\cdots,i_d)\|_*,
\end{equation}
where $\|\cdot\|_*$ denotes the nuclear norm of a matrix.
\end{definition}

\begin{theorem}[order-$d$ t-SVT\cite{qin2022low}]\label{th.2}
Given $\mathcal{T}\in\mathbb{R}^{n_1\times n_2\times\cdots\times n_d}$ with t-SVD $\mathcal{T} = \mathcal{U}*_\mathfrak{L}\mathcal{S}*_\mathfrak{L}\mathcal{V}^\mathrm{T}$, its tensor singular value thresholding (t-SVT) is defined by $\operatorname{t-SVT}_\tau(\mathcal{T}):=\mathcal{U}*_\mathfrak{L}\mathcal{S}_\tau *_\mathfrak{L}\mathcal{V}^\mathrm{T}$, where $\mathcal{S}_\tau = \mathfrak{L}^{-1}((\mathcal{S}_\mathfrak{L}-\tau)_+)$, $t_+=\max(0,t)$, which obeys
\begin{equation}\label{eq.9}
\operatorname{t-SVT}_\tau(\mathcal{T})=\arg\min_{\mathcal{X}} \tau\|\mathcal{X}\|_{\circledast,\mathfrak{L}}+\frac{1}{2}\|\mathcal{X}-\mathcal{T}\|_\mathrm{F}^2.
\end{equation}
\end{theorem}

Very recently, Kilmer et al. \cite{kilmer2021tensor} proved that tensor format equips the optimality over its flattening matrix in low-rank approximation via the normal order-3 t-SVD. Here, we present a generic result, which reveals the necessity and natural advantage of the direct study on high-order tensors.

\begin{theorem}[optimality principle]\label{th.3}
Given the t-SVD of $\mathcal{T}\in\mathbb{R}^{n_1\times n_2\times\cdots\times n_d}$ by $\mathcal{T} = \mathcal{U}*_\mathfrak{L}\mathcal{S}*_\mathfrak{L}\mathcal{V}^\mathrm{T}$, then,
$
\mathcal{T}_k:=\sum_{i=1}^k\mathcal{U}(:,i,:,\cdots,:)
*_\mathfrak{L}\mathcal{S}(i,i,:,\cdots,:)
*_\mathfrak{L}\mathcal{V}(:,i,:,\cdots,:)^\mathrm{T}
$
is the best Frobenius norm approximation over all t-SVD rank-$k$ tensor. Moreover, suppose $\mathrm{M}$ is any unfolding matrix of $\mathcal{T}$, we have,
\begin{equation}\label{eq.10}
\operatorname{rank}_{\operatorname{t-SVD}}(\mathcal{T})\leq\operatorname{rank}(\mathrm{M}),
\end{equation}
and
\begin{equation}\label{eq.11}
\|\mathcal{T}-\mathcal{T}_k\|_\mathrm{F}\leq\|\mathrm{M}-\mathrm{M}_k\|_\mathrm{F},
\end{equation}
where $\mathrm{M}_k$ is the rank-$k$ approximation of the matrix $\mathrm{M}$.
\end{theorem}

\section{Tensor Correlated Total Variation}\label{sec.4}

\begin{figure*}[t]
\centering\vspace{-0.1cm}
\includegraphics[width=0.98\linewidth]{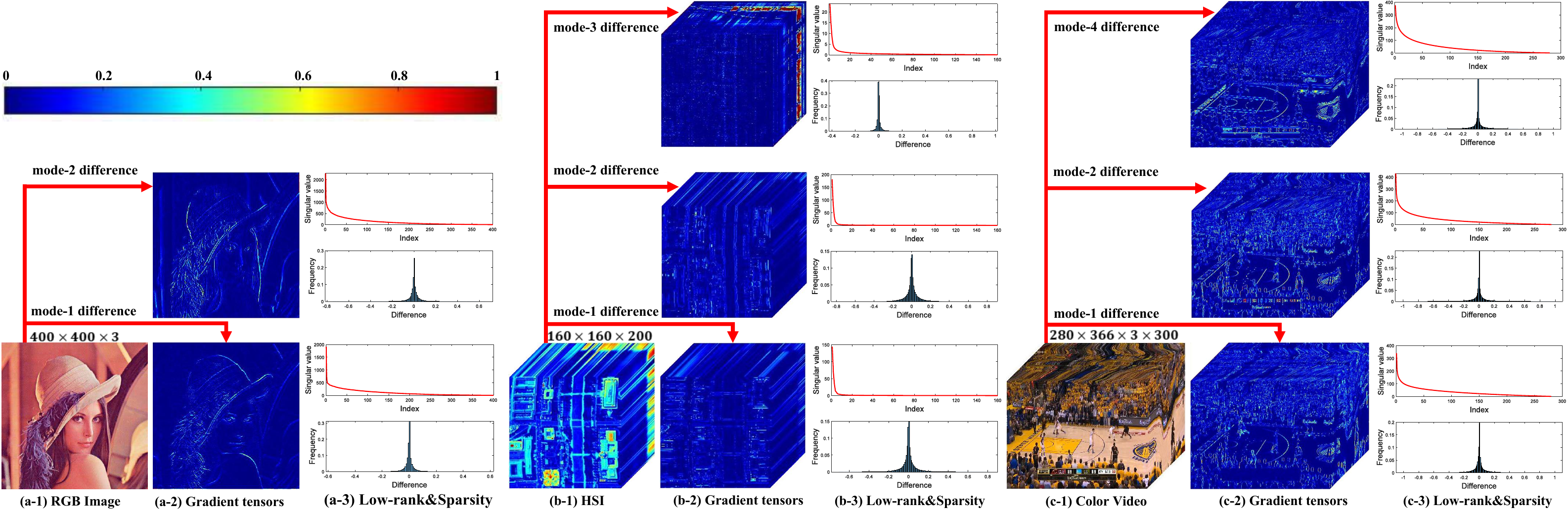}
\vspace{-0.2cm}
\caption{Illustrations of simultaneous  \textbf{L} and \textbf{S} prior structures in correlated gradient tensors. (a-1), (b-1), (c-1): three typical types of visual tensor data: RGB image, HSI and color video; (a-2), (b-2), (c-2): their correlated gradient tensors; (a-3), (b-3), (c-3): the corresponding curves of tensor singular values (upper) and frequency histograms of all their elements (below).}\label{fig.2}
\vspace{-0.5cm}
\end{figure*}

Considering a tensor $\mathcal{T}\in\mathbb{R}^{n_1\times\cdots\times n_d}$ with joint \textbf{L} and \textbf{S} priors, its low-rank property can be well characterized via the rank surrogates under the aforementioned high-order t-SVD decomposition, like the TNN. Its smoothness, on the other hand, is often captured by the low energy of $\mathcal{T}$'s gradient tensors using certain TV (semi)-norm. We first formally define such gradient tensor as follows:
\begin{definition}[gradient tensor]
For $\mathcal{T}\in\mathbb{R}^{n_1\times\cdots\times n_d}$, its gradient tensor along the $k$-th mode is defined by
\begin{equation}\label{eq.12}
\mathcal{G}_k := \nabla_k(\mathcal{T})=\mathcal{T}\times_k\mathrm{D}_{n_k}, \ k = 1,2,\cdots,d,
\end{equation}
where $\mathrm{D}_{n_k}$ is a row circulant matrix of $(-1,1,0,\cdots,0)$.
\end{definition}

The anisotropic TV (TV-1) and the isotropic TV (TV-2) are defined as
$
\|\mathcal{T}\|_{\operatorname{TV-1}}
:=\sum_{k\in\Gamma}\|\mathcal{G}_k\|_1$
and
$
\|\mathcal{T}\|_{\operatorname{TV-2}}
:=\sum_{k\in\Gamma}\|\mathcal{G}_k\|_\mathrm{F},
$
respectively, where $\Gamma$ is a priori set consisting of certain directions that $\mathcal{T}$ equips smooth continuity along these modes. For instances, images are often assumed with local smoothness along the spatial directions, i.e., $\Gamma=\{1,2\}$. HSIs and color videos have further spectral and temporal smoothness respectively, except spatial smoothness, thus $\Gamma$ can be set as $\{1,2,3\}$ and $\{1,2,4\}$ respectively. For easy notation, the two forms of TV norm are all denoted as
$\|\mathcal{T}\|_{\operatorname{TV}}$ throughout the paper.

Considering a structured tensor with simultaneous \textbf{L} and \textbf{S} priors, it is natural for existing research to generally use a sum of two separate regularizers for encoding such two priors, easily following the \textbf{L}+\textbf{S} models. Different from the previous methods, our aim is to represent both \textbf{L}+\textbf{S} priors on the gradient tensors, and especially specify a unique regularization term for concisely delivering both prior information simultaneously. This also finely complies with the fact that the two priors are always coexist with each other in natural visual tensor data, but not independently occur as the conventional models with separate \textbf{L} and \textbf{S} regularizers implicitly imply. This can be evidently observed from Fig.\;\ref{fig.2}, which illustrates such phenomenon visually on several typical types of visual tensor data. We name this proposed regularizer as \textit{tensor correlated total variation} (t-CTV\footnote{The idea of fusing \textbf{L} and \textbf{S} priors into a unique regularizer on gradient images is firstly proposed in our previous work \cite{peng2022exact}, and mainly specified for hyper-spectral images. Differently, this work considers more general tensor cases, proposes more formal and adaptable t-CTV definition, and has a wider range of application tasks.}), defined in the following:

\begin{definition}[t-CTV]
For $\mathcal{T}\in\mathbb{R}^{n_1\times\cdots\times n_d}$,  denote $\Gamma$ as a priori set consisting of directions along which $\mathcal{T}$ equips \textbf{L}+ \textbf{S} priors, and $\mathcal{G}_k, k\in\Gamma$ as its correlated gradient tensors. We define a tensor correlated total variation (t-CTV) norm\footnote{It is easy to verify that the t-CTV is a well-defined tensor (semi)-norm since the TNN item $\|\cdot\|_{\circledast,\mathfrak{L}}$ is well-defined.} as
\begin{equation}\label{eq.13}
\|\mathcal{T}\|_{\operatorname{t-CTV}}
:=\frac{1}{\gamma}\sum_{k\in\Gamma}\|\mathcal{G}_k\|_{\circledast,\mathfrak{L}},
\end{equation}
where $\gamma:=\sharp\{\Gamma\}$ equals the cardinality of $\Gamma$.
\end{definition}

\begin{figure}[tp]
\centering
\vspace{-0.4cm}
\begin{tabular}{@{}c@{}@{}c@{}@{}c@{}@{}c@{}}
\centering
\subfloat{\includegraphics[width=22mm, height = 15mm]{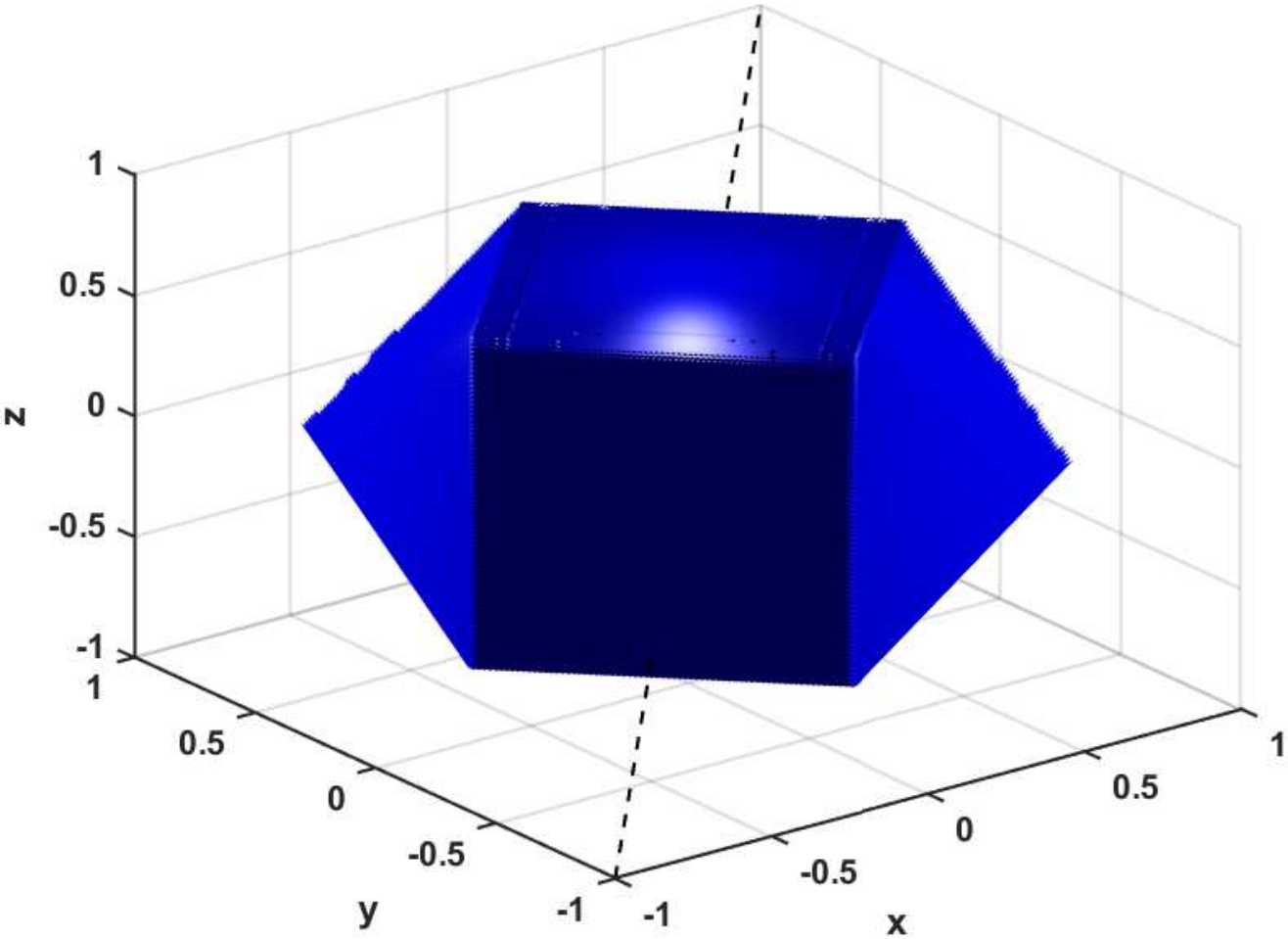}}&
\subfloat{\includegraphics[width=22mm, height = 15mm]{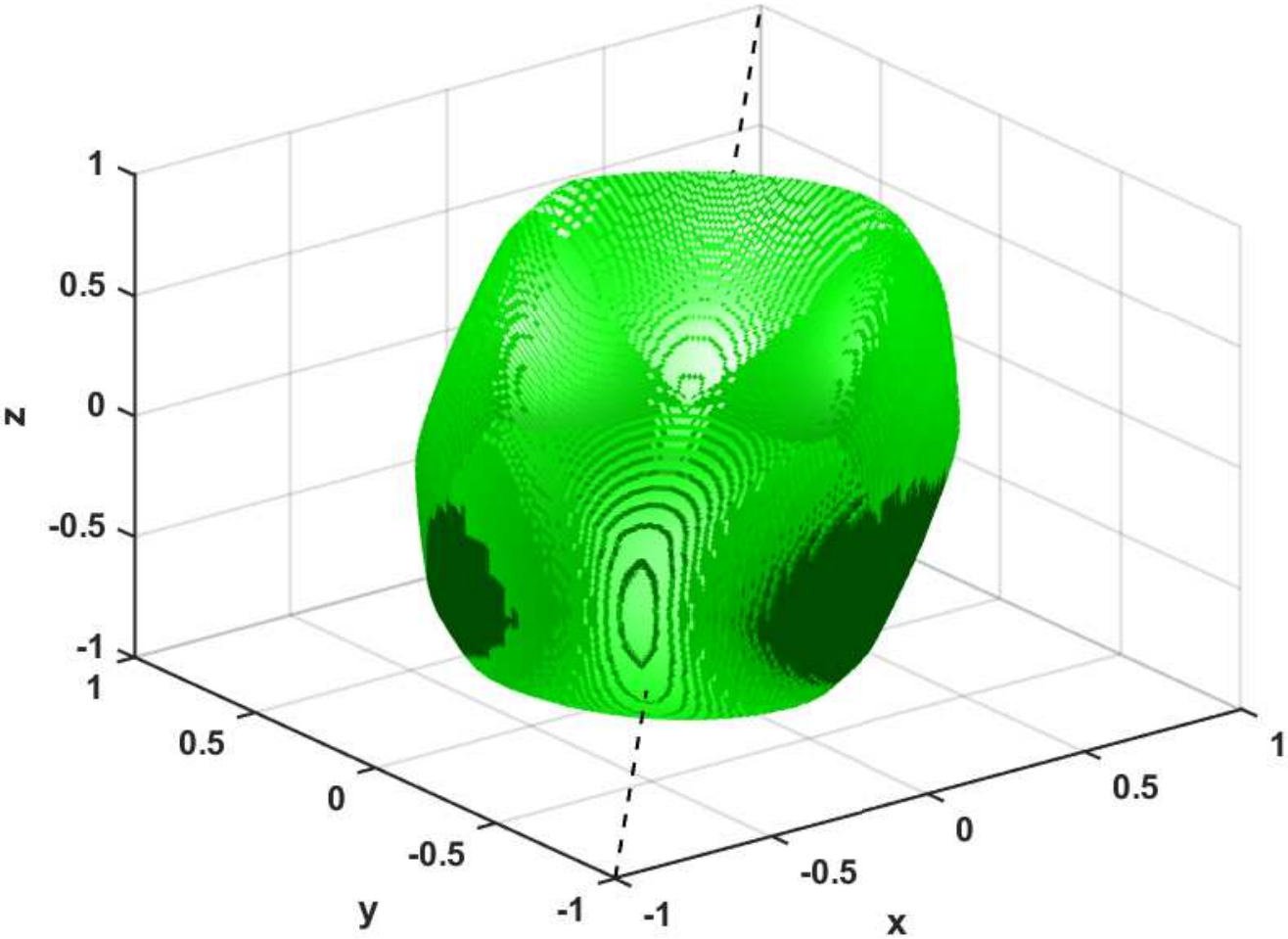}}&
\subfloat{\includegraphics[width=22mm, height = 15mm]{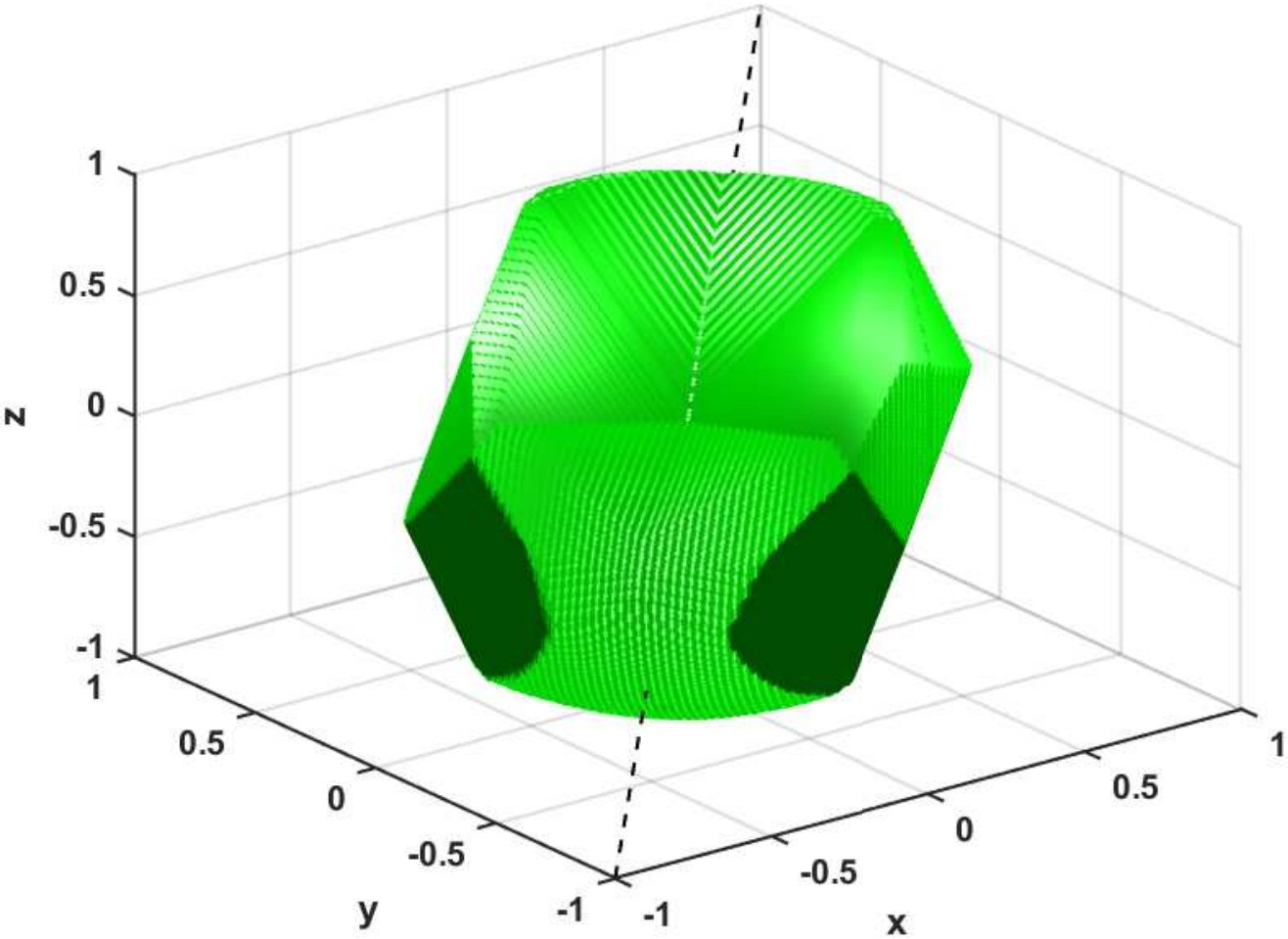}}&
\subfloat{\includegraphics[width=22mm, height = 15mm]{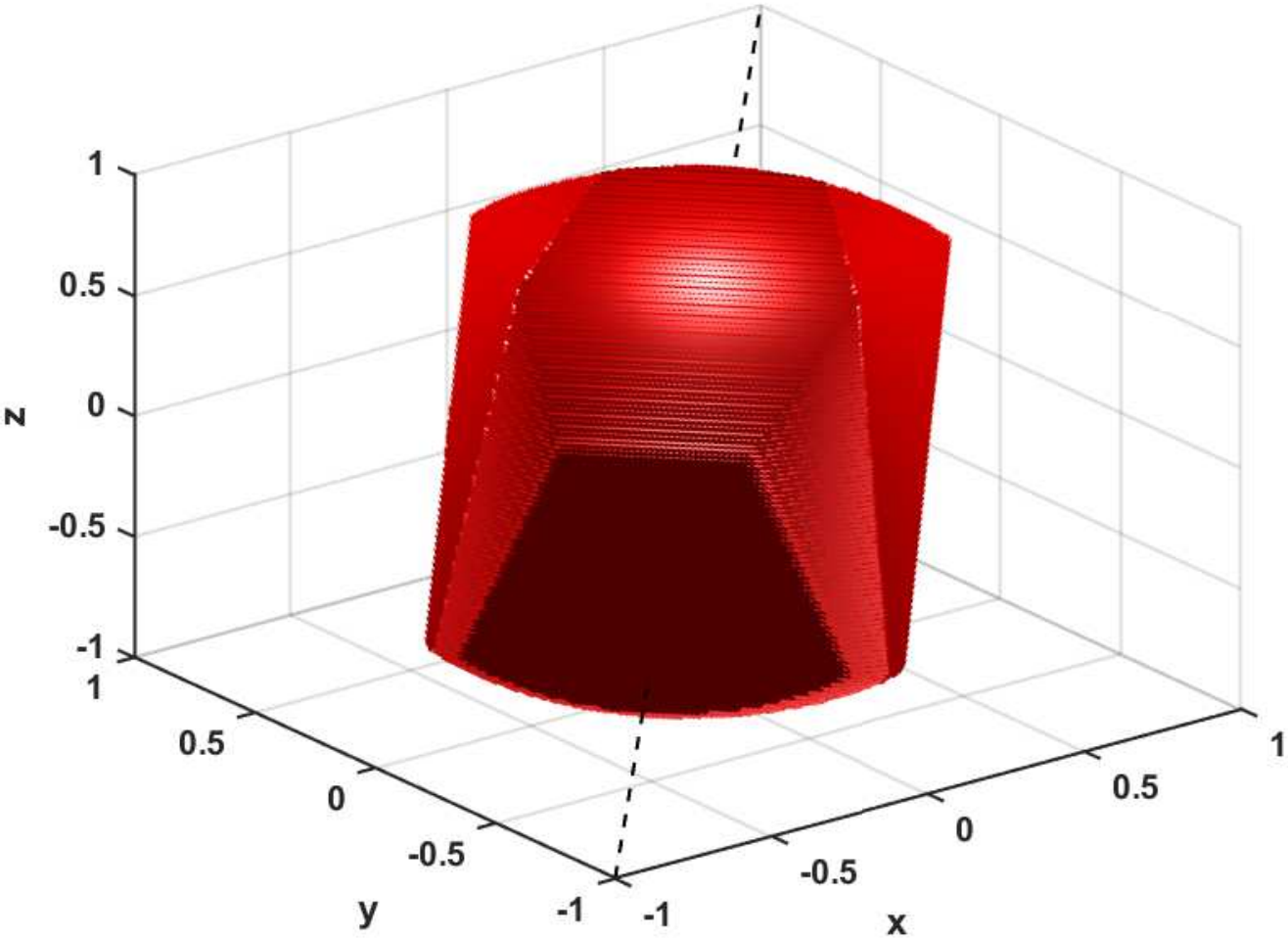}}\\
\specialrule{0em}{-6pt}{4pt}
\small (a) TNN &  \small (b) TV-1 & \small (c) TV-2 & \small  (d) t-CTV \\
\specialrule{0em}{-6pt}{4pt}
\end{tabular}
\vspace{-0.2cm}
\caption{Manifolds of the TNN, TV and t-CTV norm.}\label{fig.2a}
\vspace{-0.5cm}
\end{figure}

From the perspective of \textbf{L}-prior encoding, the t-CTV constrains the TNN metric of correlated gradient tensors $\mathcal{G}_k$ intuitively, which naturally promotes $\mathcal{G}_k$'s low-rankness property, and then enhances the similar prior structure of the original tensor, as validated in the following remark:

\begin{remark}\label{the.4}
For $\mathcal{T}\in\mathbb{R}^{n_1\times\cdots\times n_d}$ with t-SVD rank $R$, it can be verified that
$
R-1\leq\operatorname{rank}_{\operatorname{t-SVD}}(\mathcal{G}_k)\leq R,
$
where $\mathcal{G}_k$ is the gradient tensor along $k$-th mode. This means that the low-rankness between the original and gradient tensors are consistent, indicating that the t-CTV can indirectly induce the expected \textbf{L}-prior structure of the original tensor like a low-rank regularizer.
\end{remark}

Besides, from the perspective of \textbf{S}-prior encoding, since the t-CTV is a norm defined in the gradient domain, it measures as an energy control term that tends the discrete first-order derivatives of the tensor data along certain smoothness prior modes being small numerical values. This is similar to the TV norm.

\begin{remark}\label{the.4a}
For $\mathcal{T}\in\mathbb{R}^{n_1\times\cdots\times n_d}$ with t-SVD rank $R$, it can be verified that
$\|\mathcal{T}\|_{\operatorname{TV}}
\lesssim\footnote{$a\lesssim b$ means that $a\leq Cb$, where $C$ is a fixed absolute constant.}\|\mathcal{T}\|_{\operatorname{t-CTV}}
\lesssim\sqrt{R}\|\mathcal{T}\|_{\operatorname{TV}}
$,
meaning that the t-CTV and TV are compatible in sense of norm. From the viewpoint of energy minimization, both t-CTV and TV norm tend to be smaller when $\mathcal{T}$ becomes smoother, and take the minimum value of zero only if $\mathcal{T}$ is absolutely flat, indicating that the t-CTV can also indirectly induce the expected \textbf{S}-prior structure of the targeted tensor like a TV regularizer.
\end{remark}

Combining the above two points, it is expected that the t-CTV can finely encode both the \textbf{L} and \textbf{S} priors. To more intuitively observe its capability on constraining the two priors, we plot the manifolds\footnote{The manifold polytope constrains the solution space based corresponding regularization norm. Here, these manifolds are plotted on an $2\times2\times2$ tensor with slices $[0\;x;y\;z]$ and $[z\;y;x\;0]$ for easy visualization.} of TNN, TV and t-CTV norm in Fig.\;\ref{fig.2a}. One can see that the manifold of t-CTV has evident similarity to that of TV norm in the whole shape, and also exists some close characteristics with TNN like the two tangent planes seen from the front, showing the close connections in constraining the solution space in terms of low-rankness and smoothness, respectively.

Besides, it should be noted that although the t-CTV is not the unique encoding term for \textbf{L} nor \textbf{S} prior, it still achieves much better performance for joint low-rank and smooth tensor recovery than existing pure \textbf{L}, \textbf{S} and \textbf{L}+\textbf{S} models, which will be demonstrated in the following experiments.

\section{Tensor Recovery via t-CTV Minimization}\label{sec.5}

\subsection{Models}
Suppose $\mathcal{T}_0\in\mathbb{R}^{n_1\times\cdots\times n_d}$ is the underlying unknown tensor with joint \textbf{L} and \textbf{S} structural assumption. For the TC problem, we adopt the commonly used random Bernoulli sampling scheme, $\Omega\sim\operatorname{Ber}(p)$. Using t-CTV to characterize the \textbf{L} and \textbf{S} priors, the conducted t-CTV based TC model is
\begin{equation}\label{eq.16}
\min_{\mathcal{T}} \|\mathcal{T}\|_{\operatorname{t-CTV}} \ \ \text{s.t.} \ \ \mathscr{P}_{\Omega}(\mathcal{T}) = \mathscr{P}_{\Omega}(\mathcal{T}_0).
\end{equation}
For the TRPCA problem, the observation of $\mathcal{T}_0$ is a corruption with outlier or noise $\mathcal{E}_0$, denoted as
$
\mathcal{M} = \mathcal{T}_0+\mathcal{E}_0.
$
Then the t-CTV based TRPCA problem is modeled as
\begin{equation}\label{eq.17}
\min_{\mathcal{T},\mathcal{E}} \|\mathcal{T}\|_{\operatorname{t-CTV}}+\lambda\|\mathcal{E}\|_1 \ \ \text{s.t.} \ \ \mathcal{M} = \mathcal{T}+\mathcal{E}.
\end{equation}
The following recovery guarantees aim to ensure that $\mathcal{T}_0$, $(\mathcal{T}_0,\mathcal{E}_0)$ are the exact solutions of (\ref{eq.16}) and  (\ref{eq.17}), respectively, with some mild conditions.

\subsection{Incoherence Conditions}

Note that t-CTV is actually constructed by the tensor nuclear norm imposed on related gradient tensors under the high order t-SVD framework. Thus the theoretical analytical framework of classic low-rank tensor recovery can be readily borrowed. The incoherence condition is one of the most vital theoretical tools in low-rank recovery \cite{zhang2016exact}\cite{qin2022low}\cite{candes2009exact}\cite{candes2010power}. Below, we define the gradient tensor incoherence conditions, whose purpose is the same with that in the classic low-rank analysis. That is to impose or constraint the low-rankness structure of the underlying tensor, avoiding the pathological issue or hopeless case, i.e., most elements of $T_0$ are zero but it is low-rank. More descriptions are in \cite{zhang2016exact}\cite{qin2022low}\cite{candes2009exact}\cite{candes2010power}.

\begin{definition}
For $\mathcal{T}\in\mathbb{R}^{n_1\times\cdots\times n_d}$ with t-SVD rank $R$, and any $k\in\Gamma$, assume that the gradient tensors $\mathcal{G}_i$s have the skinny t-SVD $\mathcal{G}_k=\mathcal{U}_k *_\mathfrak{L}\mathcal{S}_k *_\mathfrak{L}\mathcal{V}_k^\mathrm{T}$, and then $\mathcal{T}$ is said to satisfy the gradient tensor incoherence conditions with parameter $\mu>0$ if
\begin{equation}\label{eq.18}
\max_{i_1 = 1,\cdots,n_1} \|\mathcal{U}_k^\mathrm{T}*_\mathfrak{L}\mathring{\mathfrak{e}}_1^{(i_1)}\|_\mathrm{F}^2
\leq{\mu R}/{n_1 \ell},
\end{equation}
\begin{equation}\label{eq.19}
\max_{i_2 = 1,\cdots,n_2} \|\mathcal{V}_k^\mathrm{T}*_\mathfrak{L}\mathring{\mathfrak{e}}_2^{(i_2)}\|_\mathrm{F}^2
\leq{\mu R}/{n_2 \ell},
\end{equation}
and
\begin{equation}\label{eq.20} \|\mathcal{U}_k*_\mathfrak{L}\mathcal{V}_k^\mathrm{T}\|_\infty^2
\leq{\mu R}/{n_1 n_2\ell^2},
\end{equation}
where $\mathring{\mathfrak{e}}_1^{(i_1)}$ is the order-$d$ tensor mode-1 basis sized $n_1\times1\times n_3\times\cdots\times n_d$, whose $(i_1,1,1,\cdots,1)$-th entry equals 1 and the rest equal 0, and $\mathring{\mathfrak{e}}_2^{(i_2)}:=(\mathring{\mathfrak{e}}_1^{(i_2)})^\mathrm{T}$ is the mode-2 basis.
\end{definition}

The difference between the above incoherence conditions and previous ones is that it is imposed on the gradient tensors instead of the original tensor. Note that \cite{candes2010power} has proved that almost all uniformly bounded low-rank matrices equip incoherence property well. Since difference operation on a tensor changes neither its boundness nor low-rankness, the incoherence conditions naturally hold on its gradient maps. The first two conditions (\ref{eq.18}), (\ref{eq.19}) and the third one (\ref{eq.20}) are normally called the standard and joint incoherence conditions, associated with TC and TRPCA problem, respectively.

\subsection{Main Results}
We firstly show the exact recovery guarantee of the proposed t-CTV-TC model (\ref{eq.16}).

\begin{theorem}
Consider t-CTV based TC model (\ref{eq.16}). Suppose that $\mathcal{T}_0$ obeys the standard gradient tensor incoherence conditions (\ref{eq.18})-(\ref{eq.19}) and $\Omega\sim\operatorname{Ber}(p)$. Then, there exist universal constants $c_0, c_1, c_2>0$ such that $\mathcal{T}_0$ is the unique solution to model (\ref{eq.16}) with probability at least $1-c_1\gamma(n_{(1)}n_3\cdots n_d)^{-c_2}$, provided that
\begin{equation}\label{eq.21}
p\geq c_0\mu R(\log(n_{(1)}\ell))^2/n_{(2)}\ell,
\end{equation}
where $\ell$ is the specific scale factor given in (\ref{eq.4a}), $n_{(1)}:=\max\{n_1,n_2\}$ and $n_{(2)}:=\min\{n_1,n_2\}$.
\end{theorem}

The above result shows that minimizing t-CTV norm can achieve exact tensor completion with considerable probability. The corresponding sampling complexity needs to be around $p\cdot\prod_{i=1}^d n_i \approx O(\mu R n_{(1)}n_3\cdots n_d\log^2(n_{(1)}\ell)/\ell)$. Such a bound can be acceptable if considering only the low-rankness since it differs from the degrees of freedom of an arbitrary order-$d$ tensor with t-SVD rank $R$, i.e., $O(R n_{(1)}n_3\cdots n_d)$, by a logarithmic factor. But, for a joint \textbf{L} and \textbf{S} prior structured tensor, its degree of freedom should be lower than that of a pure low-rank tensor with the same rank, which indicates that its sampling complexity for recovery could be further improved. Below, we introduce an important proposition on the lower bound for general regularization norm based TC model inspired from \cite{oymak2015simultaneously}.

\begin{Proposition}\label{pro.1}
Let $\mathcal{T}_0\in\mathbb{R}^{n_1\times\cdots\times n_d}$ with multi-structural prior simultaneously. Consider the following general TC model
\begin{equation}\label{eq.22}
\min_{\mathcal{T}} f(\mathcal{T}):=\sum w_i\|\mathcal{T}\|_{(i)} \ \ \text{s.t.} \ \ \mathscr{P}_{\Omega}(\mathcal{T}) = \mathscr{P}_{\Omega}(\mathcal{T}_0),
\end{equation}
where $\|\cdot\|_{(i)}$ denotes a regularization norm (such as TNN, TV, and t-CTV norm) modeling certain prior with Lipschitz constant $L_i$, $w_i>0$ is the wight parameter. Suppose $\Omega\sim\operatorname{Ber}(p)$ and $m$ is the number of sampling entries. Then, there exist constant $c_0,c_1>0$ such that $\mathcal{T}_0$ \textbf{is not} the unique solution of (\ref{eq.22}) with probability at least $1-\exp(-\frac{c_1m}{n_1\cdots n_d\|\bar{\mathcal{T}}_0\|_\infty^2})$, provided that
\begin{equation}\label{eq.23}
m\leq m_{\operatorname{low}}:=c_0\kappa_{\operatorname{min}}^2n_1\cdots n_d,
\end{equation}
where $\kappa_{\operatorname{min}}=\min\{\kappa_i=\|\bar{\mathcal{T}_0}\|_{(i)}/L_i\}$ and $\bar{\mathcal{T}}_0=\mathcal{T}_0/\|\mathcal{T}_0\|_\mathrm{F}$.
\end{Proposition}

Proposition \ref{pro.1} states an interesting and enlightening result that the linear combination of multi-structures promoted regularization norm can do no better than using only the best single one. It can be only determined by one single regularizer for multi-structural tensor completion in terms of the boundary of failure recovery. That provides a strong evidence that to fully explore the multiple priors, we might necessarily need to design certain new regularizer to character the joint structures, as also claimed in \cite{tomioka2013convex,oymak2015simultaneously,mu2014square}. This actually motivates us for designing the t-CTV term to reprace previous \textbf{L}+\textbf{S} terms with summarization of two elements. Now, we show that the t-CTV model (\ref{eq.16}) owns an interpretable lower sampling bound and is superior over those of pure \textbf{L}, \textbf{S} and joint \textbf{L}+\textbf{S} prior models.

\begin{theorem}\label{th.6}
For order-$d$ tensor $\mathcal{T}_0\in\mathbb{R}^{N\times\cdots\times N}$ with \textbf{L} and \textbf{S} prior structures simultaneously, denote its t-SVD rank as $R$ and gradient tensor $\mathcal{G}_k$'s sparsity (number of nonzero entries) as $S_k$, and $S=\min_{k\in\Gamma}\{S_k\}$. Then, the corresponding lower bounds of the following \textbf{L} and/or \textbf{S} models satisfy:
\begin{center}
\begin{tabular}{c|c|c}
\Xhline{1pt}
Model & $f(\mathcal{T})=$ & $m_{\operatorname{low}}\lesssim$ \\
\Xhline{1pt}
\textbf{L} & $\|\mathcal{T}\|_{\circledast,\mathfrak{L}}$ & $N^{d}\cdot\frac{R}{N}$ \\
\hline
\textbf{S} & $\|\mathcal{T}\|_{\operatorname{TV}}$ &  $N^d\cdot\frac{S}{N^{d}}$\\
\hline
\textbf{L}+\textbf{S} & $\|\mathcal{T}\|_{\circledast,\mathfrak{L}}+\alpha\|\mathcal{T}\|_{\operatorname{TV}}$
& $N^d\cdot\min\{\frac{R}{N},\frac{S}{N^{d}}\}$\\
\Xhline{1pt}
\textbf{t-CTV} & $\|\mathcal{T}\|_{\operatorname{t-CTV}}$ & $N^d\cdot\frac{R}{N}\cdot\frac{S}{N^d}$\\
\Xhline{1pt}
\end{tabular}
\end{center}
\end{theorem}

As presented in Theorem \ref{th.6}, t-CTV's lower bound, i.e.,
\begin{equation}
N^d\cdot\frac{R}{N}\cdot\frac{S}{N^d},
\end{equation}
is a product interaction form of the \textbf{L} and \textbf{S} metrics in an interpretable manner, that is rigorously less than that of the pure \textbf{L} and \textbf{S} regularization models and the widely adopted \textbf{L}+\textbf{S} ones if the underlying tensor is low-rank and smooth simultaneously. This certainly explains why our method achieves nice performance in even extremely small sampling rate, as shown in Fig.\;\ref{fig.1}. The lower the rank ratio $R/N$ is, and the lower the smooth ratio $S/N^d$ is, the lower the informatic bound of our t-CTV model should be. Whereas, existing \textbf{L}+\textbf{S} modeling manner cannot so intrinsically fuse the two priors from this theoretical perspective.

We then analyze of the t-CTV based TRPCA problem (\ref{eq.17}), whose recoverability can be also guaranteed.

\begin{theorem}\label{th.7}
Consider t-CTV based TRPCA model (\ref{eq.17}). Suppose that $\mathcal{T}_0$ obeys the standard and joint gradient tensor incoherence conditions (\ref{eq.18})-(\ref{eq.20}) and $\mathcal{E}_0$'s support set, denoted as $\Omega_0$, is uniformly distributed among all sets of cardinality $m$. Then, there exist universal constants $c_1, c_2>0$ such that $(\mathcal{T}_0,\mathcal{E}_0)$ is the unique solution to model (\ref{eq.17}) when $\lambda =1/\sqrt{n_{(1)}\ell}$ with probability at least $1-c_1\gamma(n_{(1)}n_3\cdots n_d)^{-c_2}$, provided that
\begin{equation}\label{eq.24}
\operatorname{rank}_{\operatorname{t-SVD}}(\mathcal{T}_0)\leq \frac{\rho_r n_{(2)}\ell}{\mu\log^2(n_{(1)}\ell)} \ \ \text{and} \ \
m\leq \rho_s n_1\cdots n_d,
\end{equation}
where $\rho_r, \rho_s>0$ are some numerical constants.
\end{theorem}

Theorem \ref{th.7} states that the t-CTV induced TRPCA model (\ref{eq.17}) is able to exactly recover a joint \textbf{L} and \textbf{S} structured tensor and a sparse noise with high probability. The corresponding parameters $\rho_r$ and $\rho_s$ determine the rank of $\mathcal{T}_0$ and the sparsity of $\mathcal{E}_0$, respectively. Moreover, the analysis identifies that the model (\ref{eq.17}) is parameter-free where the trade-off parameter $\lambda=1/\sqrt{n_{(1)}\ell}$ is universal, making it easier to implement in applications.

\begin{remark}
The above theoretical analysis provides solid support of the recoverability of t-CTV modeling in tensor recovery. The proof is mainly executed by possibly finely embedding the t-CTV model into the classic sparse modeling framework. Yet it should be indicated that such proof is not that easy since the dependent high-order t-SVD framework differs from common matrix algebra in many aspects. E.g., the linear conversion via tensor-tensor product is totally different from that in matrix space, and there exist difference operation and the low-rankness of gradient tensors required to be proved (see Appendix A, B and C). More importantly, the key dual certification process is different from previous works, where one needs to consider the difference transformation (see Lemmas 5 and 13 in Appendixes D and F, respectively).
\end{remark}

\begin{algorithm}[tbp]\label{alg.a1}\vspace{-1mm}
\setstretch{0.5}
\renewcommand{\algorithmicrequire}{ \textbf{Input}:}
\renewcommand{\algorithmicensure}{ \textbf{Output}:}
\caption{ADMM for solving t-CTV-TC model (\ref{eq.16})}
\begin{algorithmic}[1]
\REQUIRE observation $\mathscr{P}_{\Omega}(\mathcal{T}_0)$, priori set $\Gamma$ and transform $\mathfrak{L}$.
\STATE Initialize $\mathcal{G}_k^0=\nabla_k(\mathscr{P}_{\Omega}(\mathcal{T}_0))$, $\mathcal{E}^0=\Upsilon^0=\Lambda_k^0=\mathcal{O}$.
\STATE \textbf{while} not converge \textbf{do}
\STATE \quad Update $\mathcal{T}^{t+1}$ by (\ref{eq.29});
\STATE \quad Update $\mathcal{G}_k^{t+1}$ by (\ref{eq.30}) for each $k\in\Gamma$;
\STATE \quad Update $\mathcal{K}^{t+1}$ by (\ref{eq.31});
\STATE \quad Update multipliers $\Lambda_k^{t+1},k\in\Gamma$ and $\Upsilon^{t+1}$ by (\ref{eq.32});
\STATE \quad Let $\mu_{t+1}=\rho\mu_t$; $t = t +1$.
\STATE \textbf{end while}
\ENSURE recovered tensor $\hat{\mathcal{T}}=\mathcal{T}^{t+1}$.
\end{algorithmic}
\end{algorithm}

\section{Optimization Algorithms}\label{sec.6}
This section derives algorithms for solving the t-CTV based TC and TRPCA problem via the Alternating Direction Method of Multipliers (ADMM) framework \cite{boyd2011distributed}.
\subsection{Optimization to t-CTV TC}
First, the t-CTV-TC model (\ref{eq.16}) can be reformulated as
\begin{align}\label{eq.25}
\begin{split}
&\min_{\mathcal{T},\mathcal{G}_k,\mathcal{K}} \ \ \frac{1}{\gamma}\sum_{k\in\Gamma}\|\mathcal{G}_k\|_{\circledast,\mathfrak{L}}
+\delta_{\mathcal{K},\Omega}\\
&\text{s.t.} \ \  \mathcal{G}_k = \nabla_k(\mathcal{T}), \mathcal{T}+\mathcal{K} = \mathscr{P}_{\Omega}(\mathcal{T}_0),
\end{split}
\end{align}
where the auxiliary variable $\mathcal{G}_k$ separates the difference operation $\nabla_k(\cdot)$ and $\mathcal{K}$ compensates missing entries of $\mathcal{T}$ that is restricted in $\Omega^\bot$ using the indicative function defined as
\begin{align}\label{eq.26}
\begin{split}
\delta_{\mathcal{K},\Omega}=
\left \{
\begin{array}{ll}
0, &\mathscr{P}_{\Omega}(\mathcal{K})=\mathcal{O},\\
+\infty, &\text{otherwise}.
\end{array}
\right.
\end{split}
\end{align}
Then the augmented Lagrangian function of (\ref{eq.25}) is
\begin{flalign*}
\begin{split}
& \ \ \ \ \mathcal{L}(\mathcal{T},\{\mathcal{G}_k,k\in\Gamma\},\mathcal{K},\{\Lambda_k,k\in\Gamma\},\Upsilon)=\\
&\sum_{k\in\Gamma}(\frac{1}{\gamma}\|\mathcal{G}_k\|_{\circledast,\mathfrak{L}}
+\langle\Lambda_k, \nabla_k(\mathcal{T})-\mathcal{G}_k\rangle
+\frac{\mu_t}{2}\|\nabla_k(\mathcal{T})-\mathcal{G}_k\|_\mathrm{F}^2)\\
&
+\delta_{\mathcal{K},\Omega}+\langle\Upsilon, \mathscr{P}_{\Omega}(\mathcal{T}_0)-\mathcal{T}-\mathcal{K}\rangle
+\frac{\mu_t}{2}\|\mathscr{P}_{\Omega}(\mathcal{T}_0)-\mathcal{T}-\mathcal{E}\|_\mathrm{F}^2,
\end{split}&
\end{flalign*}
where $\mu_t$ is a positive scalar, and $\Lambda_k$ and $\Upsilon$ are Lagrange multipliers. It can be further expressed as
\begin{flalign}\label{eq.27}
\begin{split}
& \ \ \mathcal{L}=\sum_{k\in\Gamma}(\frac{1}{\gamma}\|\mathcal{G}_k\|_{\circledast,\mathfrak{L}}
+\frac{\mu_t}{2}\|\nabla_k(\mathcal{T})-\mathcal{G}_k+{\Lambda_k}/{\mu_t}\|_\mathrm{F}^2)\\
& \ \ \ \ \ \ \
+\delta_{\mathcal{K},\Omega}+\frac{\mu_t}{2}\|\mathscr{P}_{\Omega}(\mathcal{T}_0)-\mathcal{T}-\mathcal{K}+{\Upsilon}/{\mu_t}\|_\mathrm{F}^2+C,
\end{split}&
\end{flalign}
where $C$ is only the multipliers dependent squared items. Below, we show how to solve its sub-problems for each involved variable.

\textit{1) Updating $\mathcal{T}^{t+1}$}: Taking the derivative in (\ref{eq.27}) with respect to $\mathcal{T}$, it gets the following linear system
\begin{flalign}\label{eq.28}
\begin{split}
&\ \ \ \ (\mathcal{I}+\sum\nabla_k^{\mathrm{T}}\nabla_k)(\mathcal{T})\\
&= \mathscr{P}_{\Omega}(\mathcal{T}_0)-\mathcal{K}^t+\Upsilon^t/\mu_t +\sum\nabla_k^{\mathrm{T}}(\mathcal{G}_k^t-\Lambda_k^t/\mu_t),
\end{split}
\end{flalign}
where $\nabla_k^\mathrm{T}(\cdot)$ denotes the transpose operator of $\nabla_k(\cdot)$. Note that the difference operation on tensors has been proved to be linear via tensor-tensor product (see Appendix C, D). Following \cite{wang2008new}, we can apply multi-dimensional FFT, which diagonalizes $\nabla_k(\cdot)$'s corresponding difference tensors $\mathcal{D}_k$, enabling to efficiently get the optimal solution of (\ref{eq.28}) based on the convolution theorem of Fourier transforms, i.e.,
\begin{equation}\label{eq.29}
\mathcal{T}^{t+1} = \mathcal{F}^{-1}\left(
\frac{
\mathcal{F}(
\mathscr{P}_{\Omega}(\mathcal{T}_0)-\mathcal{K}^t+\Upsilon^t/\mu_t
)+\mathcal{H}
}
{
\mathbf{1}+\sum_{k\in\Gamma}\mathcal{F}(\mathcal{D}_k)^*
\odot\mathcal{F}(\mathcal{D}_k)
}
\right),
\end{equation}
where
$
\mathcal{H} = \sum_{k\in\Gamma}\mathcal{F}(\mathcal{D}_k)^*\odot
\mathcal{F}(\mathcal{G}_k^t-\Lambda_k^t/\mu_t),
$
$\mathbf{1}$ is a tensor with all entries as 1, $\odot$ is componentwise multiplication, and the division is componentwise as well.

\textit{2) Updating $\mathcal{G}_k^{t+1},k\in\Gamma$}: For each $k\in\Gamma$, extracting all items containing $\mathcal{G}_k$ from (\ref{eq.27}), we can get that
\begin{flalign*}
\begin{split}
\mathcal{G}_k^{t+1} = \arg\min_{\mathcal{G}_k}
\frac{1}{\gamma}\|\mathcal{G}_n\|_{\circledast,\mathfrak{L}}
+\frac{\mu_t}{2}\|\nabla_k(\mathcal{T}^{t+1})-\mathcal{G}_k+\frac{\Lambda_k^t}{\mu_t}\|_F^2.
\end{split}&
\end{flalign*}
The close-form solution of this sub-problem is given as
\begin{align}\label{eq.30}
\mathcal{G}_k^{t+1} = \operatorname{t-SVT}_{1/\gamma\mu_t}(\nabla_k(\mathcal{T}^{t+1})+{\Lambda_k^t}/{\mu_t})
\end{align}
via the order-$d$ t-SVT as stated in Theorem \ref{th.2}.

\textit{3) Updating $\mathcal{K}^{t+1}$}: The optimization of $\mathcal{K}$ is based on that it obeys
$\mathscr{P}_\Omega(\mathcal{K}) = \mathcal{O}$. Thus, it is updated by performing
\begin{equation}\label{eq.31}
\mathcal{K}^{t+1} = \mathscr{P}_{\Omega}(\mathcal{T}_0)-\mathcal{T}^{t+1}+{\Upsilon^t}/{\mu_t}, \
\mathscr{P}_\Omega(\mathcal{K}^{t+1}) = \mathcal{O}.
\end{equation}

\textit{4) Updating $\Lambda_k^{t+1}$ and $\Upsilon^{t+1}$}: Based on the ADMM's rule, these multipliers are updated by the following equations:
\begin{align}\label{eq.32}
\begin{split}
\left \{
\begin{array}{ll}
\Lambda_k^{t+1} = \Lambda_k^t + \mu_t(\nabla_k(\mathcal{T}^{t+1})-\mathcal{G}_k^{t+1}), \ \forall k\in\Gamma,\\
\Upsilon^{t+1} = \Upsilon^t + \mu_t(\mathscr{P}_{\Omega}(\mathcal{T}_0)-\mathcal{T}^{t+1}-\mathcal{K}^{t+1}).
\end{array}
\right.
\end{split}
\end{align}
Last, the penalty parameter $\mu_{t+1}$ is lifted by $\mu_{t+1}=\rho\mu_t$ with some control constant $\rho>1$. The whole ADMM optimization scheme is summarized in Algorithm 1.

\subsection{Optimization to t-CTV TRPCA}
The optimization to t-CTV-TRPCA is quite similar to that to the t-CTV-TC problem. Its model reformulation and augmented Lagrangian function are similar like (\ref{eq.25}) and (\ref{eq.27}), except that auxiliary variable $\mathcal{K}$ are replaced by the sparse component $\mathcal{E}$ in model (\ref{eq.17}) with corresponding regularization $\lambda\|\mathcal{E}\|_1$, and the observation is $\mathcal{M}$ instead of $\mathscr{P}_{\Omega}(\mathcal{T}_0)$. Thus, under similar analysis, the ADMM iteration system with respect to t-CTV-TRPCA is briefly derived as follows:
\begin{align}
\mathcal{T}^{t+1} &= \mathcal{F}^{-1}\left(
\frac{
\mathcal{F}(
\mathcal{M}-\mathcal{E}^t+\Upsilon^t/\mu_t
)+\mathcal{H}
}
{
\mathbf{1}+\sum_{k\in\Gamma}\mathcal{F}(\mathcal{D}_k)^*
\odot\mathcal{F}(\mathcal{D}_k)
}
\right),
\label{eq.33}
\end{align}
\begin{align}
\mathcal{G}_k^{t+1} &= \operatorname{t-SVT}_{1/\gamma\mu_t}(\nabla_k(\mathcal{T}^{t+1})+{\Lambda_k^t}/{\mu_t}),
\label{eq.34}
\end{align}
\begin{align}
\mathcal{E}^{t+1} &= \mathcal{S}_{1/\lambda\mu_t}(\mathcal{M}-\mathcal{T}^{t+1}+\Upsilon^t/\mu_t),
\label{eq.35}
\end{align}
\begin{align}
\Lambda_k^{t+1} &= \Lambda_k^t + \mu_t(\nabla_k(\mathcal{T}^{t+1})-\mathcal{G}_k^{t+1}),
\label{eq.36}
\end{align}
\begin{align}
\Upsilon^{t+1} &= \Upsilon^t + \mu_t(\mathcal{M}-\mathcal{T}^{t+1}-\mathcal{E}^{t+1}),\label{eq.37}
\end{align}
where $\mathcal{S}(\cdot)$ in (\ref{eq.35}) is the soft-thresholding operator. The corresponding ADMM is described in Algorithm 2.

\begin{algorithm}[tbp]\label{alg.a2}
\setstretch{0.5}
\renewcommand{\algorithmicrequire}{ \textbf{Input}:}
\renewcommand{\algorithmicensure}{ \textbf{Output}:}
\caption{ADMM for solving t-CTV-TRPCA model (\ref{eq.17})}
\begin{algorithmic}[1]
\REQUIRE observation $\mathcal{M}$, priori set $\Gamma$ and transform $\mathfrak{L}$.
\STATE Initialize $\mathcal{G}_k^0=\mathcal{E}^0=\Upsilon^0=\Lambda_k^0=\mathcal{O}$.
\STATE \textbf{while} not converge \textbf{do}
\STATE \quad Update $\mathcal{T}^{t+1}$ by (\ref{eq.33});
\STATE \quad Update $\mathcal{G}_k^{t+1}$ by (\ref{eq.34}) for each $k\in\Gamma$;
\STATE \quad Update $\mathcal{K}^{t+1}$ by (\ref{eq.35});
\STATE \quad Update multipliers $\Lambda_k^{t+1},k\in\Gamma$ and $\Upsilon^{t+1}$ by (\ref{eq.36}), (\ref{eq.37});
\STATE \quad Let $\mu_{t+1}=\rho\mu_t$; $t = t +1$.
\STATE \textbf{end while}
\ENSURE recovered tensors $\hat{\mathcal{T}}=\mathcal{T}^{t+1}$ and $\hat{\mathcal{E}} = \mathcal{E}^{t+1}$.
\end{algorithmic}
\end{algorithm}

\subsection{Computational Complexity Analysis}
For Algorithm 1, the computational complexity in each iteration contains four parts, i.e., steps $3\sim6$. First, the time complexity in step 3 that mainly using FFT is $O(n_1\cdots n_d\log(n_1\cdots n_d))$. Second, the time complexity for order-$d$ t-SVT in step 4 is $O(n_1 n_2(n_3\cdots n_d)^2+n_{(1)}n_{(2)}^2n_3\cdots n_d)$, corresponding to the linear transform along mode-3 to mode-$d$ and the matrix SVD, respectively \cite{qin2022low}. Detailed t-SVT algorithmic procedure is given in supplementary material, appendix A. The steps 5 and 6 have the same complexity $O(n_1\cdots n_p)$ with only element-wise computation. In all, the pre-iteration computational complexity of Algorithm 1 is $O(n_1\cdots n_d\log(n_1\cdots n_d)+n_1 n_2(n_3\cdots n_d)^2+n_{(1)}n_{(2)}^2n_3\cdots n_d)$.

It is easy to see that the Algorithm 2 has the same per-iteration computational complexity since the only difference is in step 5, soft-thresholding operator, which only takes $O(n_1\cdots n_d)$ computation cost. Compared with the baseline TNN minimization algorithm in \cite{qin2022low}, it only increases $O(n_1\cdots n_d\log(n_1\cdots n_d))$, i.e., the difference equation computation in the first part. Besides, such computational complexity is in the same order to that of ADMM based TNN+TV method in \cite{qiu2021robust}.

\subsection{Convergence Analysis}

Both Algorithms 1 and 2 are in form of multi-block ADMM, whose convergence cannot be guaranteed directly in normal cases unlike the classic two-block ADMM \cite{chen2016direct}. Fortunately, benefiting from that the linear constraints are separable in our models, we can equivalently transform our algorithms into standard two-block ADMM forms \cite{boyd2011distributed}. Thus the general convergence of two-block ADMM proved in \cite{boyd2011distributed} can be adopted directly. We only analyze Algorithm 1 below. Similar result can be deduced for Algorithm 2.

\begin{remark}
Denote matrices $\mathrm{Y}$, $\mathrm{A}$, $\mathrm{X}_1$, $\mathrm{B}$ respectively as
$$
\begin{bmatrix}
0 \\
\vdots\\
0\\
\overline{\mathfrak{L}(\mathscr{P}_{\Omega}(\mathcal{T}_0))}
\end{bmatrix},
\begin{bmatrix}
\overline{\mathcal{I}_\mathfrak{L}} &  &  &  \\
 & \ddots &  &  \\
 &  & \overline{\mathcal{I}_\mathfrak{L}} &  \\
 &  &  & \overline{\mathcal{I}_\mathfrak{L}}
\end{bmatrix},
\begin{bmatrix}
\overline{\mathfrak{L}(\mathcal{G}_{k_1})} \\
\vdots\\
\overline{\mathfrak{L}(\mathcal{G}_{k_{\gamma}})} \\
\overline{\mathcal{K}_\mathfrak{L}}
\end{bmatrix},
\begin{bmatrix}
\overline{\mathfrak{L}(\mathcal{D}_{k_1})} \\
\vdots\\
\overline{\mathfrak{L}(\mathcal{D}_{k_{\gamma}})} \\
-\overline{\mathcal{I}_\mathfrak{L}}
\end{bmatrix},
$$
and $\mathrm{X}_2:=\overline{\mathcal{T}_\mathfrak{L}}$. Then, t-CTV-TC model's reformulation (\ref{eq.25}) can be converted to a standard two-block ADMM with constraint condition $\mathrm{A}\mathrm{X}_1-\mathrm{B}\mathrm{X}_2=\mathrm{Y}$, and the updating of $(\mathrm{X}_1,\mathrm{X}_2)$ is equal to the iteration system in Algorithm 1 since there exists uniqueness between the original and transform domains. Note that it has been proved that general two-block ADMM with closed, proper and convex objective is convergent in residual, objective and variables, see \cite{boyd2011distributed}. This directly yields the convergence of the ADMM iterates in Algorithm 1.
\end{remark}

\section{Experimental Results}\label{sec.7}

This section conducts experiments on synthetic tensors with joint \textbf{L} and \textbf{S} priors to verify our main theoretical results, and applies to real visual tensors to substantiate the effectiveness of the proposed methods. More results are given in supplementary material. All experiments are implemented on the platform of MATLAB (2021a) with Intel(R) Core(TM) i5-10400F 2.90-GHz CPU and 16GB memory.

\subsection{Simulations}

We use a simple program to generate a tensor with determined t-SVD rank and equipping intrinsic smooth structures. For an order-$d$ tensor $\mathcal{T}$ sized $n_1\times n_2\times\cdots\times n_d$, for its each frontal slice $\mathrm{T}^j\in\mathbb{R}^{n_1\times n_2},j=1,\cdots,n_3\cdots n_d$, we randomly select $s(s\ll n_1 n_2)$ initial points and divide the slice into $s$ regions by nearest neighbor principle with each area with the same value sampled from $N(0,1)$. Then $\mathcal{T}_0$ is obtained by $\mathcal{T}$'s best t-SVD rank $R$ approximation as given in Theorem \ref{th.3}. Since the t-SVD is orientation dependent, this makes $\mathcal{T}_0$ has well smooth property in the first two directions. Such property is similar to the spatial smoothness of visual data.

\begin{figure}[tp]
\renewcommand{\arraystretch}{0.5}
\setlength\tabcolsep{0.5pt}
\centering
\begin{tabular}{cc}
\centering
\includegraphics[width=44mm, height = 30mm]{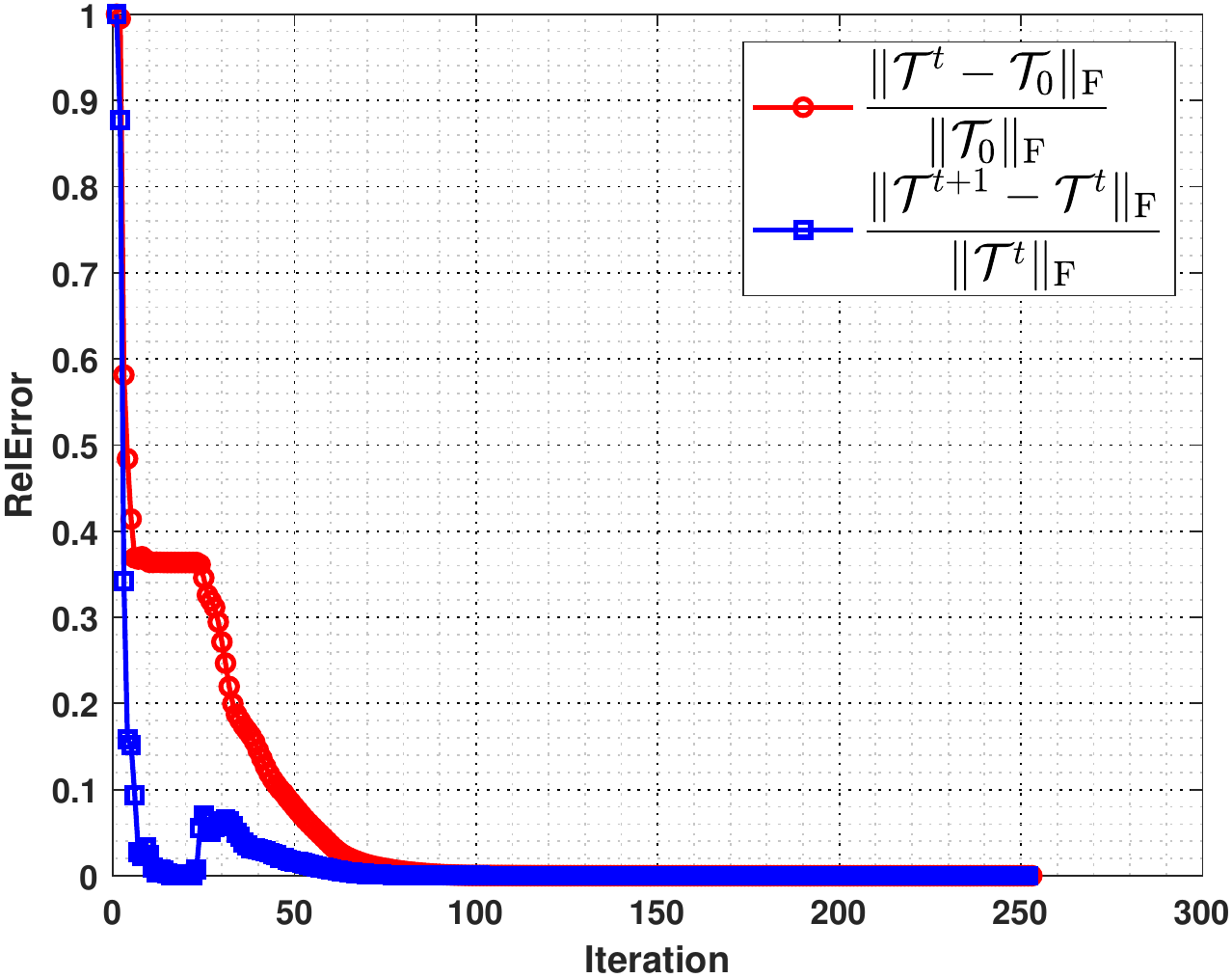}&
\includegraphics[width=44mm, height = 30mm]{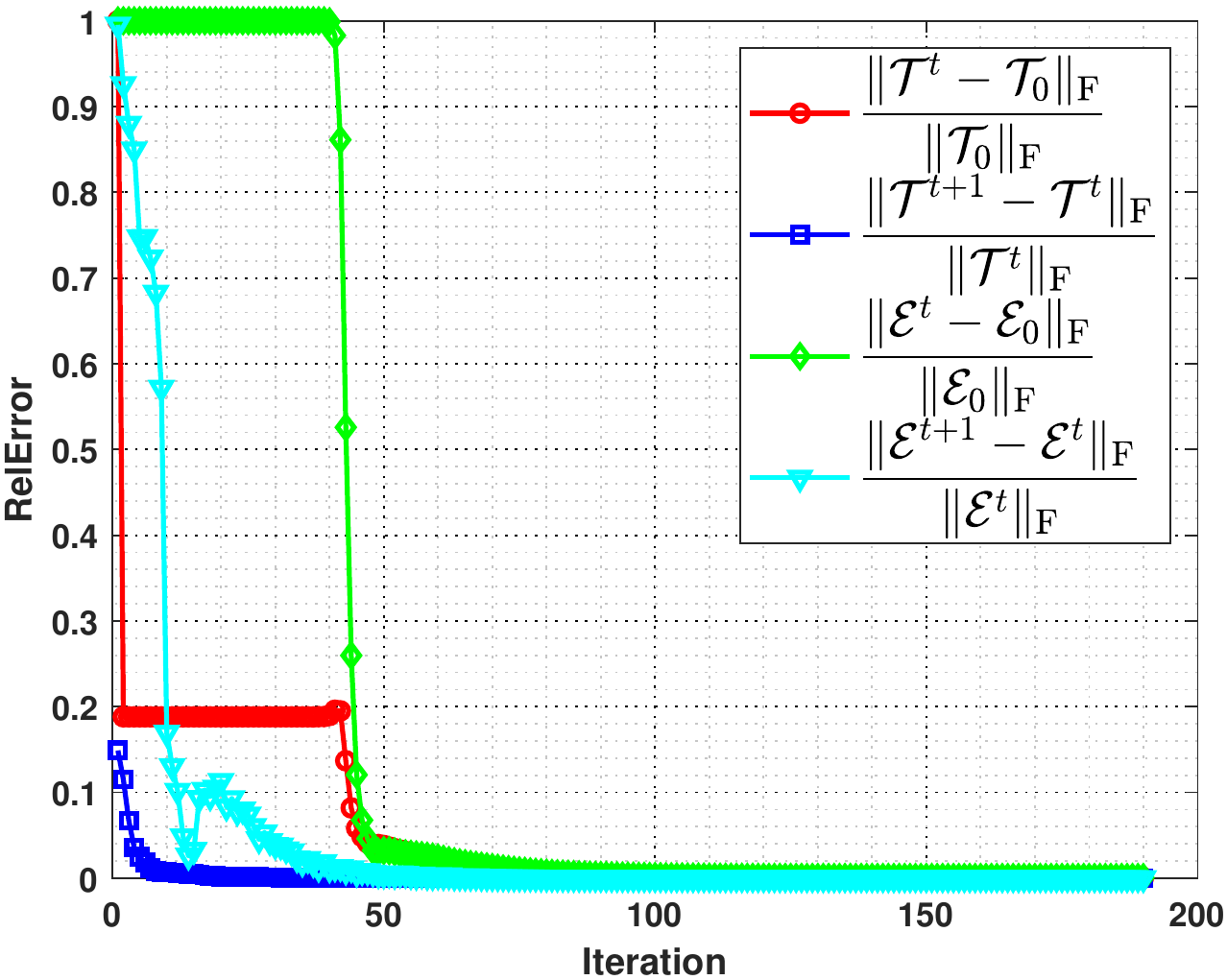}\\
\small (a) Algorithm 1 &  \small (b) Algorithm 2 \\
\end{tabular}
\vspace{-0.2cm}
\caption{Convergence curves of Algorithm 1 and Algorithm 2.}\label{fig.3}
\vspace{-0.2cm}
\end{figure}

\textbf{1) Empirical convergence}: We first verify the algorithms' convergence for solving t-CTV based TC and TRPCA models. Randomly generating a $60\times60\times60\times60$ tensor $\mathcal{T}_0$ with t-SVD rank $R = 10$, we simulate the convergence behavior of Algorithm 1 for TC problem when observing $50\%$ entries randomly. For the TRPCA model solved by Algorithm 2, we further construct a sparse tensor $\mathcal{E}_0$ whose support set is chosen uniformly at random with cardinality $m=0.05\times 60^4$, and its nonzero value are set as $\pm1$ randomly. The corresponding \textit{relative error} (RelErr\footnote{The RelErr of $\mathcal{T}$ to $\mathcal{T}_0$ is defined by ${\|\mathcal{T}-\mathcal{T}_0\|_\mathrm{F}}/{\|\mathcal{T}_0\|_\mathrm{F}}$.}) curves are plotted in Fig.\;\ref{fig.3} (a), (b), respectively, which show the errors gradually approach to zero after $50$ iterations and thus validates the convergence\footnote{We choose the DFT based t-SVD in this experiment. Afterwards, without other statements, the transform in t-SVD is set as the DFT.}.

\textbf{2) Exact recovery}: We then verify that the proposed t-CTV based TC and TRPCA model can achieve exact recovery for joint \textbf{L}+\textbf{S} tensors, as proved in our main theories.

For TC problem, we test on tensors $\mathcal{T}_0$ sized $N\times N\times 60\times60$, with varying dimension $N=100,200,400$, and set the t-SVD rank $R$ as $0.05N, 0.1N$ and $0.15N$. The sampling number $m$ is chosen by $4,3,2$ times of $d_r = R(2N-R)\times60\times60$, which can be viewed as a good quantity for reference. Table \ref{table.2} reports the recovered tensors $\hat{\mathcal{T}}$s' t-SVD ranks $\hat{R}$ and the corresponding RelErr $\|\hat{\mathcal{T}}-\mathcal{T}_0\|_\mathrm{F}/\|\mathcal{T}_0\|_\mathrm{F}$. Here, we consider three transforms: DFT, DCT, \textit{random orthogonal transform} (DOT). The results reveal that t-CTV-TC obtains correct rank estimation and accurate recovery results,i.e., RelErr$\leq10^{-5}$.

\begin{table}[tp]
\renewcommand{\arraystretch}{1.15}
\setlength\tabcolsep{3.0pt}
\footnotesize
  \caption{
  Performance of model (\ref{eq.16}) on synthetic tensors.
  }\label{table.2}
  \setlength{\abovecaptionskip}{5pt}
  \setlength{\belowcaptionskip}{5pt}
  \centering
  \vspace{-0.2cm}
\begin{tabular}{c|c |c| c|cc|cc|cc  }
     \Xhline{1pt}
      \multirow{2}{*}{$N$}  &  \multirow{2}{*}{$R$} &  \multirow{2}{*}{$\frac{m}{d_r}$ }&  \multirow{2}{*}{$p$} &
      \multicolumn{2}{c|}{DFT} &
      \multicolumn{2}{c|}{DCT} &
      \multicolumn{2}{c}{ROT} \\
       \cline{5-6}
       \cline{7-8}
       \cline{9-10}
      \qquad&\qquad&\qquad&\qquad&
     $\hat{R}$ & RelErr &$\hat{R}$ & RelErr &$\hat{R}$ & RelErr\\
     \Xhline{1pt}
     100 &5  &4& 0.39 & 5  & 4.54e-7 & 5 &6.01e-6 & 5 &4.73e-7\\
     200 &20  &3& 0.57 & 20  & 1.83e-7 & 20&  3.30e-6 & 20 &1.59e-7\\
     400 &60  &2& 0.56 & 60  &2.90e-6 & 60 & 4.07e-6 & 60 &8.92e-6\\
     \Xhline{1pt}
\end{tabular}
\vspace{-0.2cm}
\end{table}

For the latter, consider $N\times N\times60\times60$ synthetic joint \textbf{L}+\textbf{S} tensor $\mathcal{T}_0$ varying $N=100$ and $200$ with $R=0.05N$, $0.1N$, and random sparse tensor $\mathcal{E}_0$ with sparsity $m=2\times10^6,8\times10^6$ (corresponding sparsity ratio equals $5.56\%$). The tolerance to determine $\hat{\mathcal{E}}$'s sparsity is set as $0.001$. The recovered results are listed in Table \ref{table.3}, showing that the recovery is correct with accurate rank/sparsity estimation, and tiny errors. This verifies the exact recovery guarantees.

\begin{table}[tp]
\renewcommand{\arraystretch}{1.15}
\setlength\tabcolsep{3.0pt}
\footnotesize
  \caption{
  Performance of model (\ref{eq.17}) on synthetic tensors.
  }\label{table.3}
  \setlength{\abovecaptionskip}{5pt}
  \setlength{\belowcaptionskip}{5pt}
  \centering
  \vspace{-0.2cm}
\begin{tabular}{c|c |c| cccc|cccc  }
     \Xhline{1pt}
      \multirow{2}{*}{$N$} &  \multirow{2}{*}{$R$} &  \multirow{2}{*}{$m$}&
      \multicolumn{4}{c|}{DFT} &
      \multicolumn{4}{c}{DCT} \\
       \cline{4-7}
       \cline{8-11}
      \qquad&\qquad&\qquad&
     $\hat{R}$ & RelErr$\mathcal{T}$ & $\hat{m}$ & RelErr$\mathcal{E}$ &$\hat{R}$ & RelErr$\mathcal{T}$ & $\hat{m}$ & RelErr$\mathcal{E}$\\
     \Xhline{1pt}
     100 &5  &2e6& 5  & 1.85e-7 &2e6 &2.46e-7 & 5 &9.97e-6 &2e6 &1.14e-7\\
     100 &10  &2e6& 10  & 2.18e-6 &2e6 &  3.96e-6 & 10 &4.38e-7 &2e6 &7.27e-7\\
     200 &10  &8e6& 10  &9.32e-7 &8e6 & 9.84e-7 & 10 &5.03e-7  &8e6 &3.64e-6\\
     200 &20  &8e6& 20  &1.83e-6 &8e6 & 3.03e-6 & 20 &8.29e-6 &8e6 &7.10e-6\\
     \Xhline{1pt}
\end{tabular}
\vspace{-0.2cm}
\end{table}

\textbf{3) Phase transition and comparison to baselines}: To further explore the ability of t-CTV based tensor recovery methods, we examine the recovery performance with varying t-SVD ranks of $\mathcal{T}_0$ from different sampling rates $p$ for the TC problem, and that with varying t-SVD ranks of $\mathcal{T}_0$ from different sparsity of $\mathcal{E}_0$ for the TRPCA task, respectively. Moreover, two baselines are considered for comparison, including the pure \textbf{L} prior model (order-$d$ TNN minimization \cite{lu2019tensor}\cite{qin2022low}), and the \textbf{L}+\textbf{S} prior model (order-$d$ TNN plus anisotropic TV norm\cite{chen2018tensor}\cite{qiu2021robust}). We omit those only considering the pure \textbf{S} priors since such models can hardly get comparable recovery results in such cases. The objective function of these baselines are listed in Table \ref{table.4}.

\begin{figure}[tp]
\renewcommand{\arraystretch}{0.5}
\setlength\tabcolsep{0.5pt}
\centering
\vspace{-0.2cm}
\begin{tabular}{ccc}
\centering
\includegraphics[width=29.4mm, height = 29.4mm]{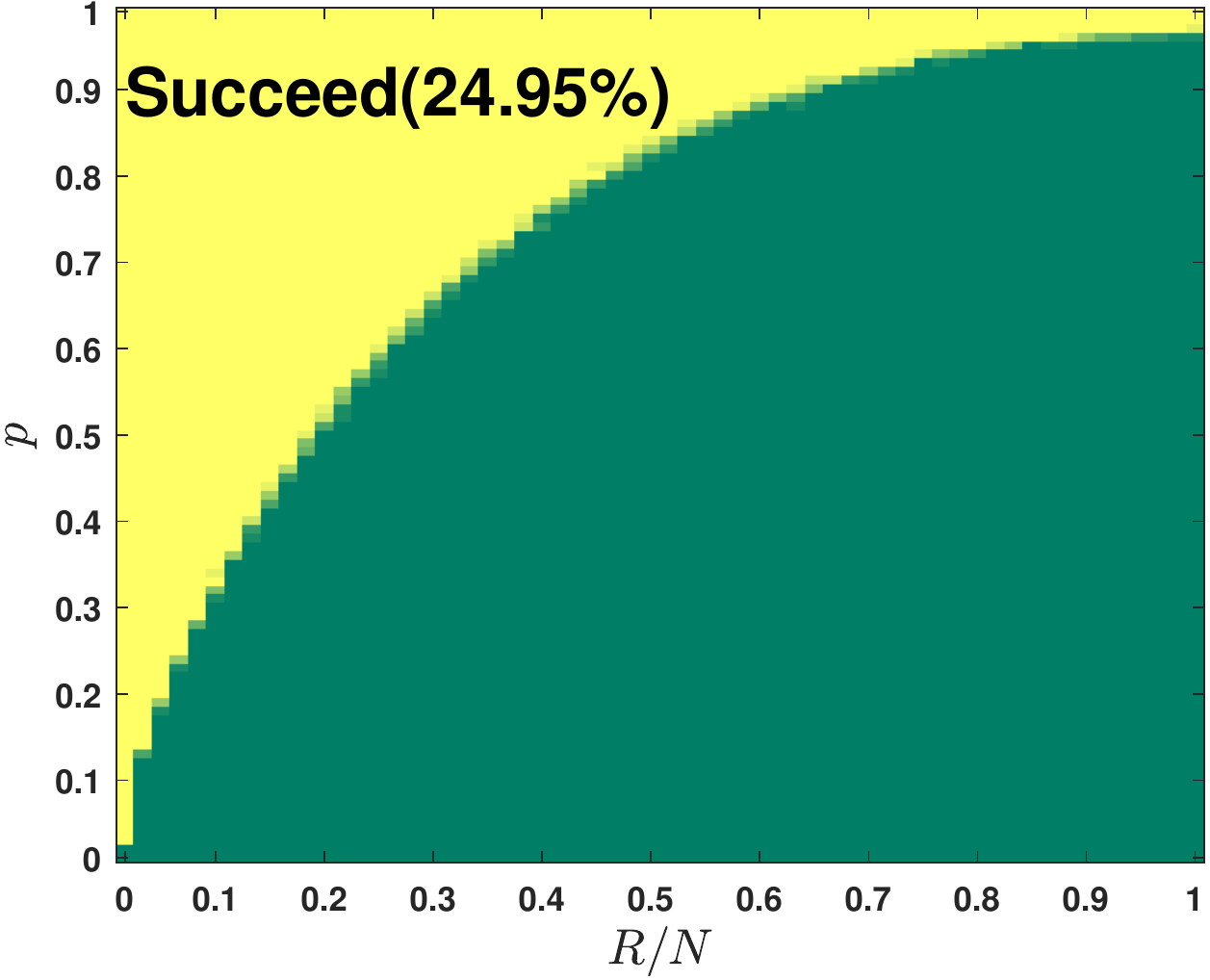}&
\includegraphics[width=29.4mm, height = 29.4mm]{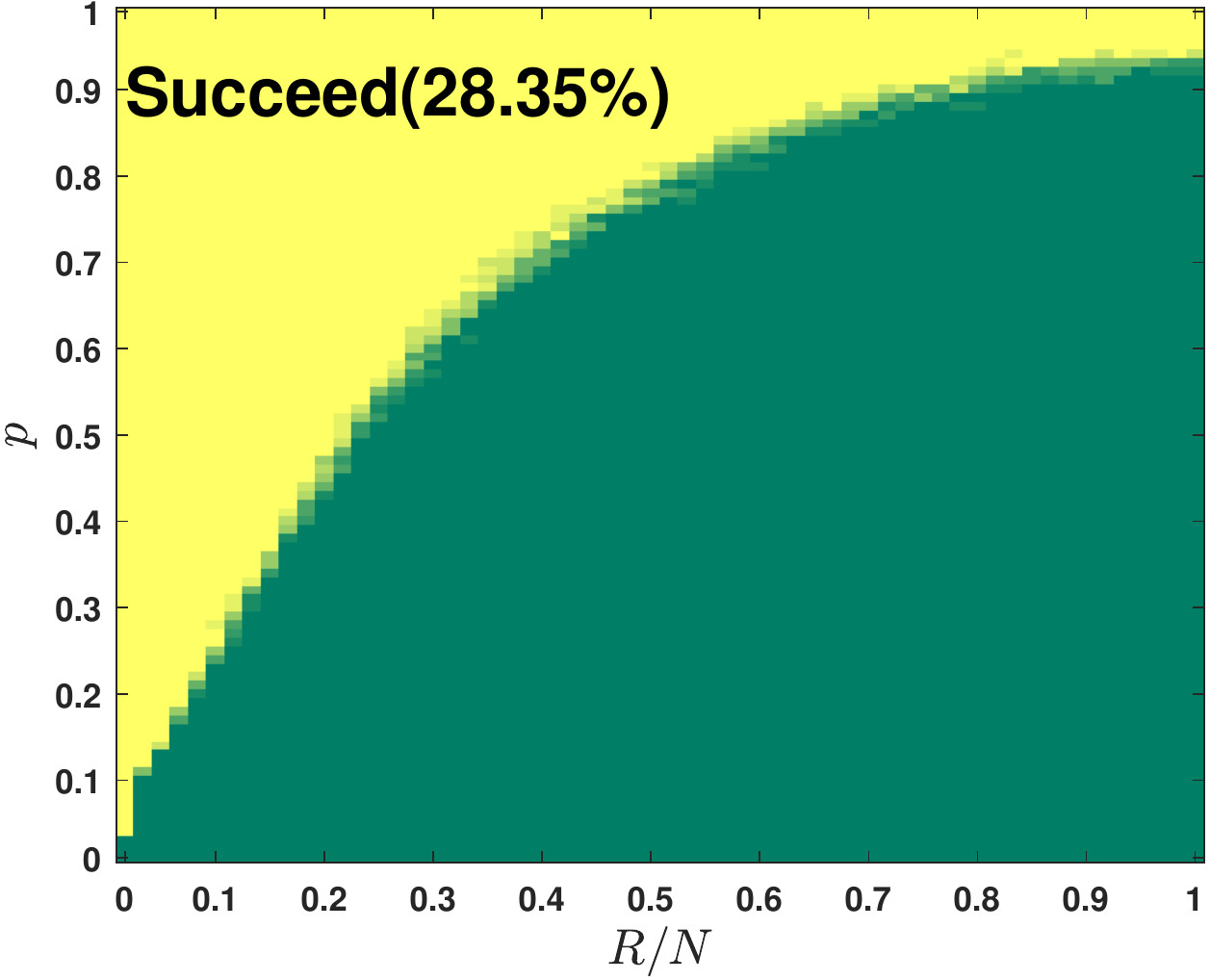}&
\includegraphics[width=29.4mm, height = 29.4mm]{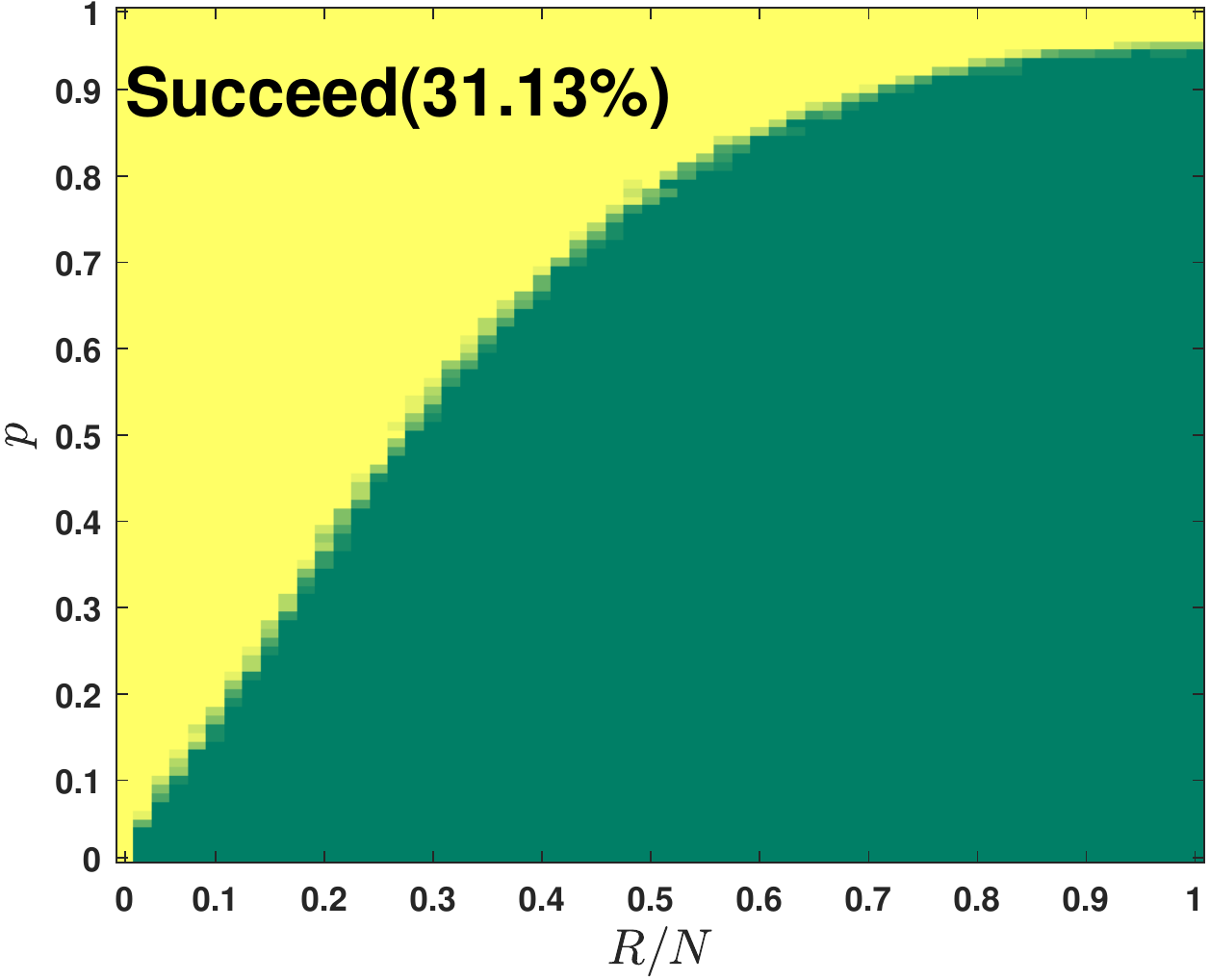}\\
\small (a) TNN &  \small (b) TNN+TV  &\small (c) t-CTV\\
\end{tabular}
\vspace{-0.2cm}
\caption{Phase transitions of t-CTV-TC model (\ref{eq.16}) with varying t-SVD ranks of $\mathcal{T}_0$ and sampling rates.}\label{fig.4}
\vspace{-0.3cm}
\end{figure}

For the TC task, we consider $\mathcal{T}_0$ sized $60\times60\times60\times60$ and set t-SVD rank $R=[1:1:60]$ and random sampling rate $p=[0.01:0.01:1]$. For each $(R/N,p)$-pair, $N=60$, we simulate 30 test instances and declare a trail to be a success if the average RelErr of $\hat{\mathcal{T}}$ is less than $0.05$. Fig.\;\ref{fig.4} plots the phase transition diagrams (yellow=$100\%$, green=$0\%$), which evidently shows that t-CTV based TC model recovers more cases than TNN and TNN+TV models. Specially, compared with TNN+TV model, requiring to carefully tune the tradeoff parameter between \textbf{L} and \textbf{S} regularizers, t-CTV-TC entirely avoids this parameter setting issue.

\begin{figure}[tp]
\renewcommand{\arraystretch}{0.5}
\setlength\tabcolsep{0.5pt}
\centering
\begin{tabular}{ccc}
\centering
\includegraphics[width=29.4mm, height = 29.4mm]{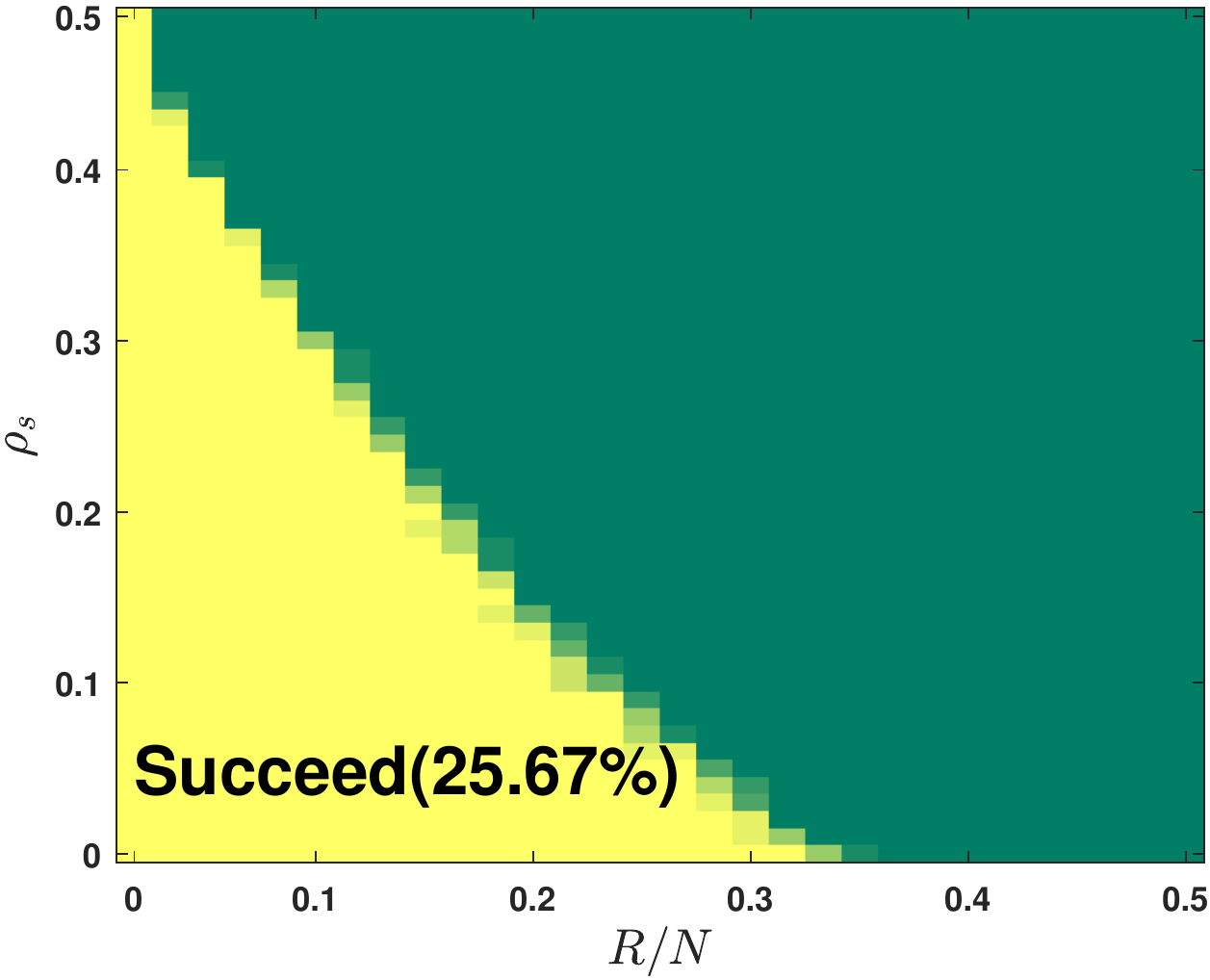}&
\includegraphics[width=29.4mm, height = 29.4mm]{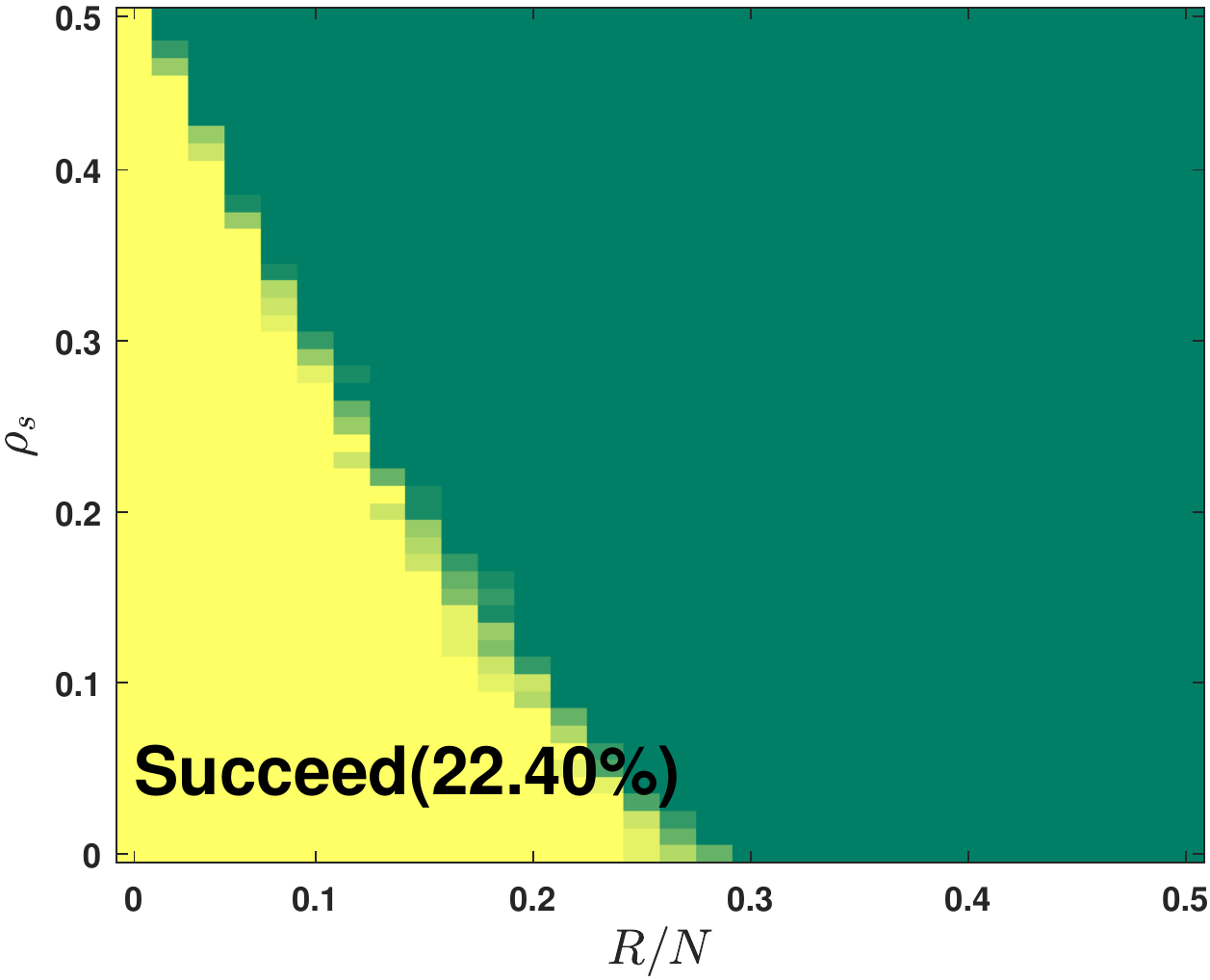}&
\includegraphics[width=29.4mm, height = 29.4mm]{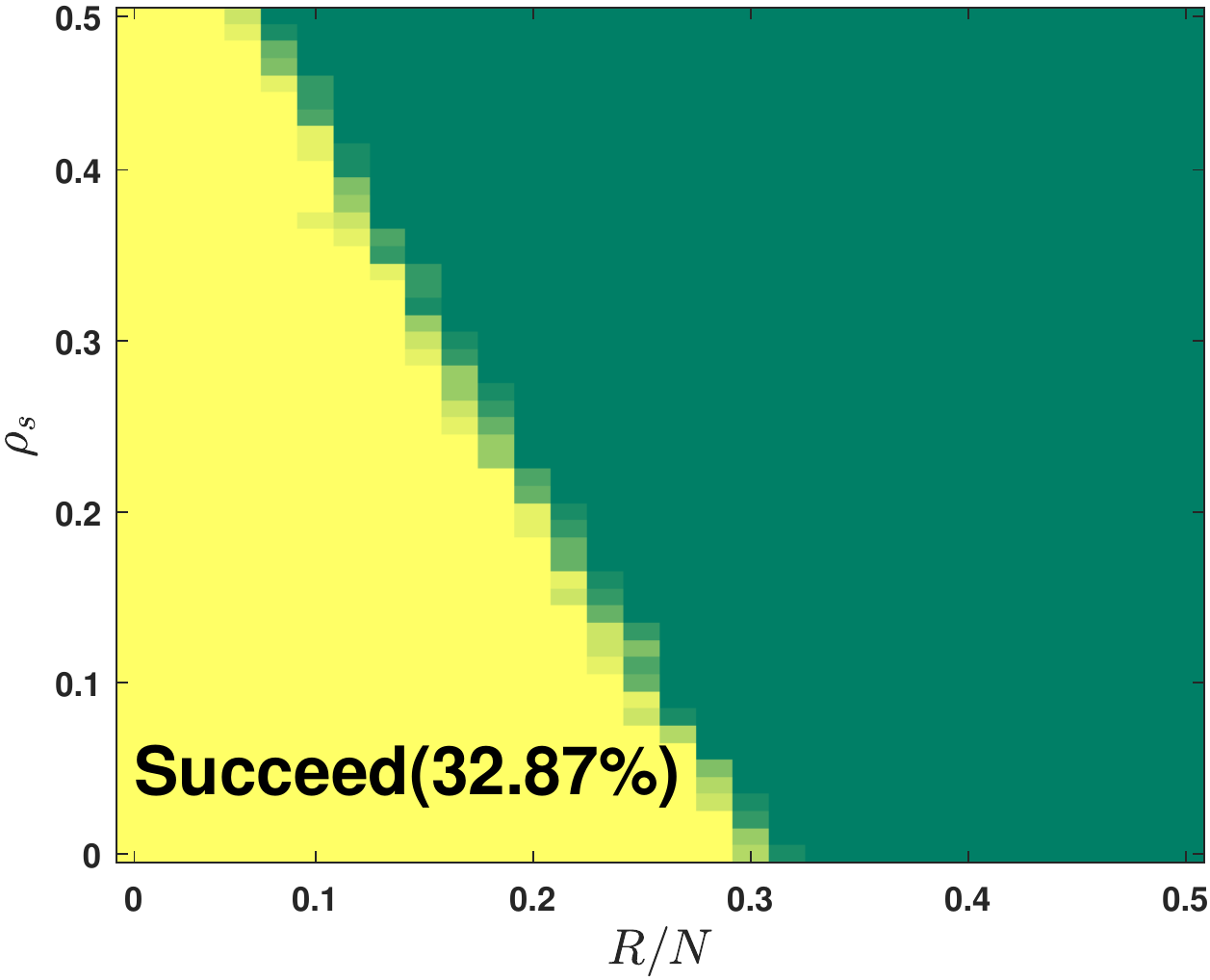}\\
\small (a) TNN &  \small (b) TNN+TV  &(c) \small t-CTV\\
\end{tabular}
\vspace{-0.2cm}
\caption{Phase transitions of t-CTV-TRPCA model (\ref{eq.17}) with varying t-SVD ranks of $\mathcal{T}_0$ and sparsity of $\mathcal{E}_0$.}\label{fig.5}
\vspace{-0.3cm}
\end{figure}

For the TRPCA task, we conduct tensor $\mathcal{T}_0$ in the same size  with varying rank $R=[1:1:30]$. As for the sparse tensor $\mathcal{E}_0$, we vary its sparsity ratio $\rho_s={m}/{60^4}$ from $0.01$ to $0.50$ with interval $0.01$. Similarly, we simulate 30 times against each $(R/N,\rho_s)$-pair for statistical stability. The corresponding fraction of correct recovery for TNN, TNN+TV and t-CTV-TRPCA are plotted in Fig.\;\ref{fig.5}. It shows that t-CTV-TRPCA attains the best recovery ability with the correct area ratio $32.87\%$, evidently much higher than those obtained by TNN and TNN+TV baselines. Note that the TNN+TV model's performance appears worse than the pure low-rank model. This can be rationally explained by that the TNN based TRPCA model has a solid recovery guarantee with determined parameter $\lambda$ \cite{lu2019tensor}, while the TNN+TV is not guaranteed in theory. Note that TNN+TV need to preset two model parameters, $\alpha$ and $\lambda$. In this phase transition experiment, we take $\lambda$ the same as that in the TNN model and carefully tune $\alpha$ from an appropriate range related to the smooth construction manner as aforementioned. By contrast, our t-CTV-TRPCA is parameter-free, and simultaneously with theoretical guarantees, making it easier and more reliable to be used in practice.

\subsection{Applications to Visual Data Inpainting}

In this subsection, we apply the proposed t-CTV-TC method to the inpainting tasks for various visual tensor data and make comparison with SOTA tensor based methods considering \textbf{L} or \textbf{L}+\textbf{S} prior models, as listed in Table \ref{table.5}. Unless otherwise stated, all parameters involved in these competing methods are optimally assigned or selected as suggested in the reference papers. The \textit{peak signal-to-noise ratio} (PSNR), \textit{structural similarity} (SSIM \cite{wang2004image}) and \textit{feature similarity} (FSIM\cite{zhang2011fsim}) are employed to evaluate the recovery performance, all of which tends better performance with larger value.

\textbf{1) Results on color images}: We randomly select 50 RGB images from the USC-SIPI\footnote{\url{https://sipi.usc.edu/database/}} and BSD\footnote{\url{https://www2.eecs.berkeley.edu/Research/Projects/CS/vision/bsds/}} databases, of sizes $256\times256\times3$ and $321\times481\times3$, respectively. First, we test on the random element-wise sampling with \textit{sampling rate} (SR) varying from $3\%$ to $80\%$. Table \ref{table.6} reports the inpainting results in terms of quantitative metrics on average. It is seen that our t-CTV method achieves the best performance in all evaluate criteria under all sampling cases. It can be seen that its advantage is more significant when SR is extremely low. As to the computational time, it can be seen that the average running time of proposed t-CTV is in the same order of magnitude as most of other methods (like SNN, KBR, TNN, SNN-TV, SPC-TV) and faster than several ones (like BCPF, IRTNN, MF-TV and TNN-TV). Fig.\;\ref{fig.6} shows the recovered images under SR=$10\%,20\%,40\%$ and $60\%$, revealing the recoveries obtained by our method are the closest to the ground truths. Considering its good recovery performance, it should be rational to say that the proposed method is efficient.

We further test the performance of all competing methods in extreme small sampling cases, when SR is set as $5\%$, $2\%$, $1\%$ and even $0.5\%$, meaning that $95\%$, $98\%$, $99\%$ and $99.5\%$ pixels are missing. Two typical cases are shown in Figs. \ref{fig.1} and \ref{fig.7}, and our method evidently makes a relatively finer recovery while all other competing methods fail to a large extent. This verifies the result in Theorem \ref{th.6} that t-CTV has a smaller lower sampling bound than conventional \textbf{L} and \textbf{L}+\textbf{S} models. As far as we know, there are no other methods capable of achieving similarly workable performance in such extreme cases for single image inpainting.

Last, we consider more challenging situations when the missing pixels are in structured masks including dead lines, wave lines, star patterns and text patterns. Some inpainting examples are depicted in Fig.\;\ref{fig.8}. One can see that our method can still perform evidently superior beyond other competing peers. Especially, it's known that the pure \textbf{L} prior models cannot estimate the missing entries when missing regions are some whole rows or columns \cite{candes2009exact}. As seen in Fig.\;\ref{fig.8}, the SNN, BCPF, KBR, IRTNN and TNN methods are all invalid completely. In comparison, our t-CTV also uses the low-rank metric TNN in form, just instead defined on the gradient tensors. Yet it can finely process such extremely ill-possed cases. It also achieves better visualization results than those obtained by \textbf{L}+\textbf{S} prior models, showing its capability of delivering the two priors in its unique regularizer.

\begin{table}[tp]
\renewcommand{\arraystretch}{1.15}
\setlength\tabcolsep{5.0pt}
\footnotesize
  \caption{
  Categories of related tensor completion methods.
  }\label{table.5}
  \setlength{\abovecaptionskip}{5pt}
  \setlength{\belowcaptionskip}{5pt}
  \centering
  \vspace{-0.2cm}
\begin{tabular}{c|c}
     \Xhline{1pt}
     Types & Methods\\
     \Xhline{1pt}
     \textbf{L}
     & SNN\cite{liu2012tensor}, BCPF \cite{zhao2015bayesian1},
     KBR\cite{xie2017kronecker}, IRTNN\cite{wang2021generalized},
     TNN\cite{qin2022low}\\
     \hline
     \textbf{L}$\&$\textbf{S}
     & MF-TV\cite{ji2016tensor},SNN-TV\cite{li2017low},SPC-TV\cite{yokota2016smooth},
     TNN-TV\cite{qiu2021robust}\\
     \Xhline{1pt}
\end{tabular}
\vspace{-0.2cm}
\end{table}

\begin{table}[tp]
\renewcommand{\arraystretch}{1.15}
\setlength\tabcolsep{3.0pt}
\footnotesize
  \caption{
  Objectives of baseline TNN and TNN+TV models.
  }\label{table.4}
  \setlength{\abovecaptionskip}{5pt}
  \setlength{\belowcaptionskip}{5pt}
  \centering
  \vspace{-0.2cm}
\begin{tabular}{c|c |c}
     \Xhline{1pt}
     Types & TC & TRPCA\\
     \Xhline{1pt}
     TNN
     &$\|\mathcal{T}\|_{\circledast,\mathfrak{L}}$
     &$\|\mathcal{T}\|_{\circledast,\mathfrak{L}}+\lambda\|\mathcal{E}\|_1$\\
     \hline
     TNN+TV
     &$\|\mathcal{T}\|_{\circledast,\mathfrak{L}}+\alpha\|\mathcal{T}\|_{\operatorname{TV}}$
     &$\|\mathcal{T}\|_{\circledast,\mathfrak{L}}+\alpha\|\mathcal{T}\|_{\operatorname{TV}}+\lambda\|\mathcal{E}\|_1$\\
     \Xhline{1pt}
\end{tabular}
\vspace{-0.2cm}
\end{table}

\begin{table*}[tp]
\renewcommand{\arraystretch}{1.15}
\setlength\tabcolsep{3.0pt}
\footnotesize
  \caption{
  Color image inpainting performances of all competing methods under different sampling rates. Each result is averaged over all data. The best and second best result are highlighted in \textbf{bold} and \underline{underline}, respectively. (/s: second).
  }\label{table.6}
  \setlength{\abovecaptionskip}{5pt}
  \setlength{\belowcaptionskip}{5pt}
  \centering
  \vspace{-0.2cm}
\begin{tabular}{l||c|c|c|c|c|c|c|c|c|c|c|c|c|c|c|c|c|c}
     \Xhline{1pt}
     SR & \multicolumn{3}{c|}{$3\%$} & \multicolumn{3}{c|}{$5\%$} & \multicolumn{3}{c|}{$10\%$} & \multicolumn{2}{c|}{$20\%$} & \multicolumn{2}{c|}{$40\%$} & \multicolumn{2}{c|}{$60\%$} & \multicolumn{2}{c|}{$80\%$} & \multirow{2}{*}{Time/s}\\
     \cline{2-18}
     Method & PSNR & SSIM & FSIM & PSNR & SSIM & FSIM & PSNR & SSIM & FSIM & PSNR & SSIM & PSNR & SSIM & PSNR & SSIM & PSNR & SSIM & \quad \\
     \Xhline{1pt}
     SNN & 15.99 & 0.292 & 0.654 & 17.21 & 0.365 & \underline{0.690} & 19.62 & 0.498 & 0.752 & 22.71 & 0.674 & 27.28 & 0.859 & 31.48 & 0.944 & 36.59 & 0.983 & 10.49\\
     BCPF & 15.98 & 0.258 & 0.640 & 17.50 & 0.328 & 0.677 & 20.10 & 0.452 & 0.744 & 23.45 & 0.637 & 27.53 & 0.815 & 30.27 & 0.889 & 32.17 & 0.924 & 84.69\\
     KBR & 15.20 & 0.219 & 0.614 & 16.21 & 0.272 & 0.643 & 18.67 & 0.386 & 0.708 & 23.36 & 0.631 & \underline{28.96} & 0.859 & \underline{33.77} & 0.948 & \underline{40.16} & \underline{0.989} & 27.15\\
     IRTNN & 11.82 & 0.123 & 0.563 & 14.63 & 0.210 & 0.614 & 18.46 & 0.375 & 0.710 & 22.43 & 0.597 & 27.80 & 0.829 & 32.50 & 0.932 & 38.32 & 0.979 & 45.95\\
     TNN & 15.33 & 0.254 & 0.640 & 16.67 & 0.321 & 0.675 & 19.16 & 0.454 & 0.740 & 22.49 & 0.642 & 27.40 & 0.845 & 32.02 & 0.940 & 37.87 & 0.983 & 5.96\\
     MF-TV & 6.46 & 0.048 & 0.520 & 7.21 & 0.064 & 0.509 & 9.06 & 0.108 & 0.504 & 15.13 & 0.327 & 26.37 & 0.809 & 31.84 & 0.939 & 36.17 & 0.981 & 200.3\\
     SNN-TV & 16.73 & 0.432 & 0.600 & 18.31 & 0.498 & 0.658 & \underline{21.79} & \underline{0.657} & \underline{0.753} & 24.48 & 0.789 & 28.47 & \underline{0.909} & 32.08 & \underline{0.960} & 36.59 & 0.987 & 14.28\\
     SPC-TV & 15.66 & 0.322 & \underline{0.661} & 17.20 & 0.377 & \underline{0.690} & 19.76 & 0.504 & \underline{0.753} & 22.87 & 0.672 & 26.71 & 0.841 & 29.67 & 0.923 & 33.14 & 0.970 & 17.96\\
     TNN-TV & \underline{17.66} & \underline{0.460} & 0.613 & \underline{19.32} & \underline{0.532} & 0.668 & 21.16 & 0634 & 0.717 & \underline{24.71} & \underline{0.792} & 28.39 & 0.908 & 31.80 & 0.959 & 36.52 & 0.987 & 40.14\\
     \textbf{t-CTV} & \textbf{21.32} & \textbf{0.622} & \textbf{0.782} & \textbf{22.67} & \textbf{0.682} & \textbf{0.820} & \textbf{24.59} & \textbf{0.763} & \textbf{0.868} & \textbf{27.18} & \textbf{0.849} & \textbf{31.31} & \textbf{0.931} & \textbf{35.26} & \textbf{0.970} & \textbf{40.55} & \textbf{0.993} & 17.13\\
     \Xhline{1pt}
\end{tabular}
\vspace{-0.2cm}
\end{table*}

\begin{figure*}[tp]
\renewcommand{\arraystretch}{0.5}
\setlength\tabcolsep{0.5pt}
\centering
\vspace{-0.1cm}
\begin{tabular}{cccccccccccc}
\centering
\scriptsize{\tiny 6.21/0.026} & \scriptsize{\tiny 24.84/0.790} & \scriptsize{\tiny \underline{25.39}/\underline{0.792}} & \scriptsize{\tiny 24.05/0.752} & \scriptsize{\tiny 24.63/0.777} & \scriptsize{\tiny 24.77/\underline{0.792}} & \scriptsize{\tiny 11.04/0.175} & \scriptsize{\tiny 21.58/0.539} & \scriptsize{\tiny 24.47/0.775} & \scriptsize{\tiny 20.64/0.479} & \scriptsize{\tiny \textbf{27.16}/\textbf{0.853}} & \scriptsize{\tiny PSNR/SSIM}\\
\includegraphics[width=14.8mm, height = 21mm]{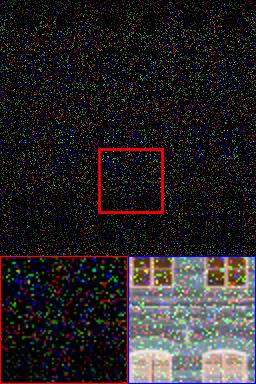}&
\includegraphics[width=14.8mm, height = 21mm]{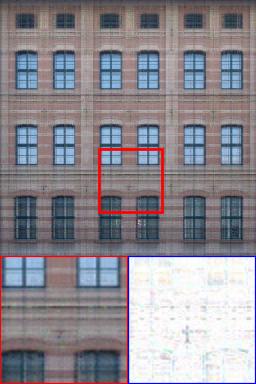}&
\includegraphics[width=14.8mm, height = 21mm]{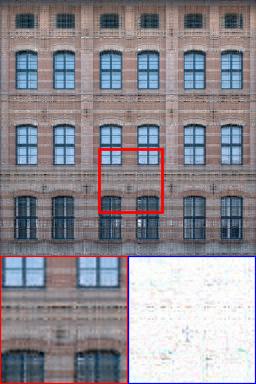}&
\includegraphics[width=14.8mm, height = 21mm]{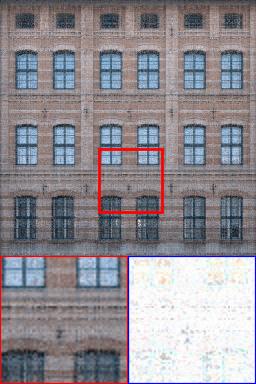}&
\includegraphics[width=14.8mm, height = 21mm]{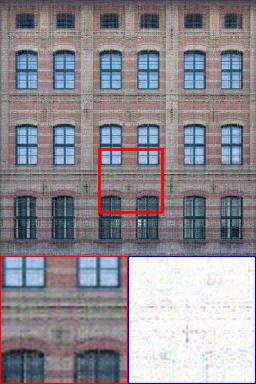}&
\includegraphics[width=14.8mm, height = 21mm]{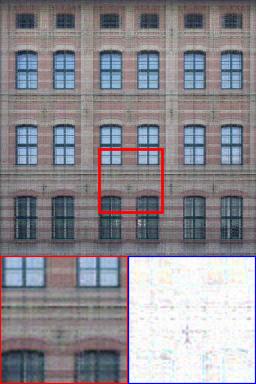}&
\includegraphics[width=14.8mm, height = 21mm]{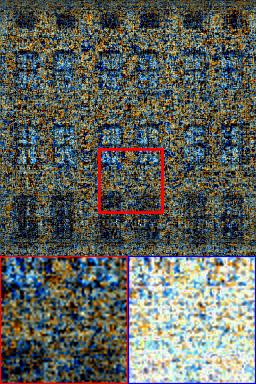}&
\includegraphics[width=14.8mm, height = 21mm]{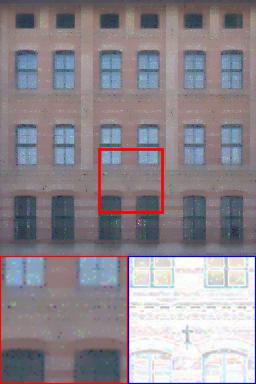}&
\includegraphics[width=14.8mm, height = 21mm]{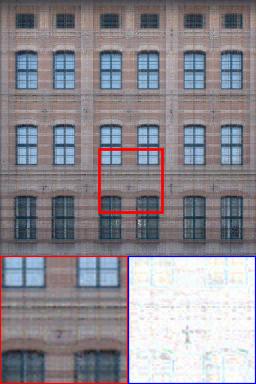}&
\includegraphics[width=14.8mm, height = 21mm]{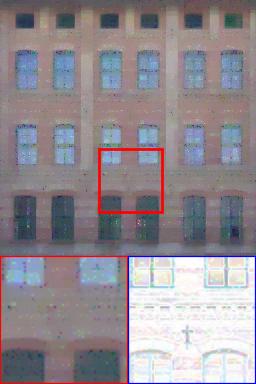}&
\includegraphics[width=14.8mm, height = 21mm]{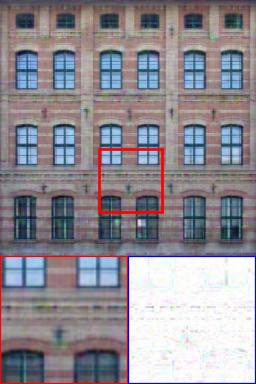}&
\includegraphics[width=14.8mm, height = 21mm]{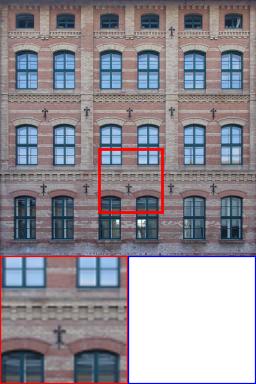}\\
\scriptsize{\tiny 5.27/0.036} & \scriptsize{\tiny 24.28/0.745} & \scriptsize{\tiny 25.31/0.704} & \scriptsize{\tiny 25.69/0.726} & \scriptsize{\tiny 25.11/0.674} & \scriptsize{\tiny 24.33/0.697} & \scriptsize{\tiny 20.75/0.505} & \scriptsize{\tiny 26.25/0.858} & \scriptsize{\tiny 24.01/0.735} & \scriptsize{\tiny \underline{27.77}/\underline{0.868}} & \scriptsize{\tiny \textbf{29.71}/\textbf{0.888}} & \scriptsize{\tiny PSNR/SSIM}\\
\includegraphics[width=14.8mm, height = 21mm]{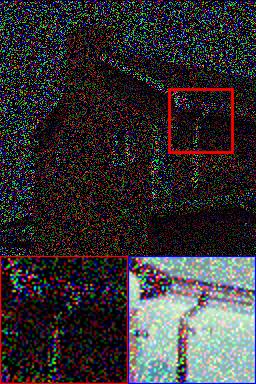}&
\includegraphics[width=14.8mm, height = 21mm]{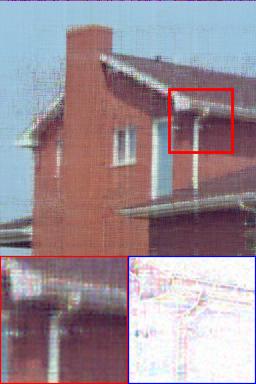}&
\includegraphics[width=14.8mm, height = 21mm]{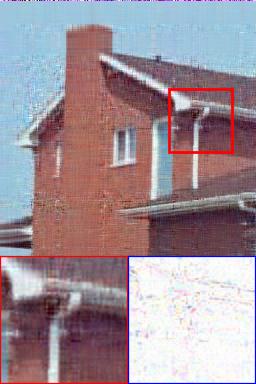}&
\includegraphics[width=14.8mm, height = 21mm]{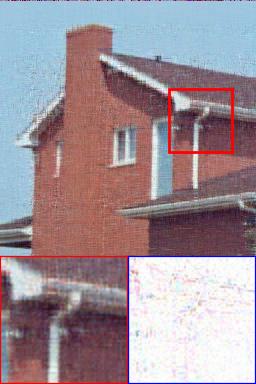}&
\includegraphics[width=14.8mm, height = 21mm]{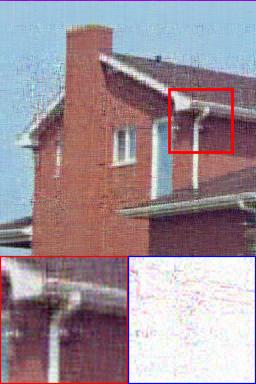}&
\includegraphics[width=14.8mm, height = 21mm]{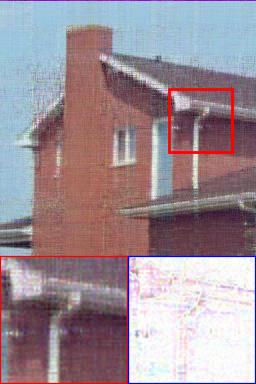}&
\includegraphics[width=14.8mm, height = 21mm]{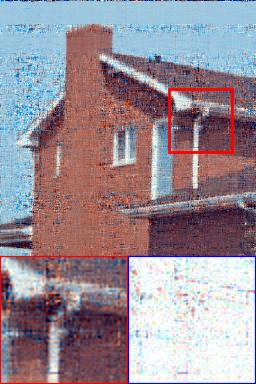}&
\includegraphics[width=14.8mm, height = 21mm]{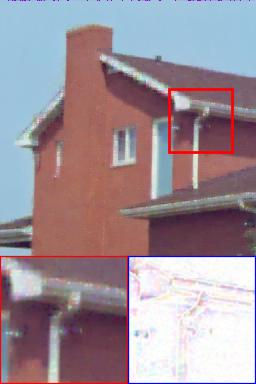}&
\includegraphics[width=14.8mm, height = 21mm]{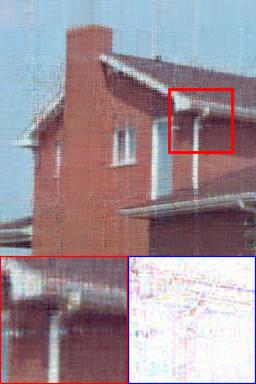}&
\includegraphics[width=14.8mm, height = 21mm]{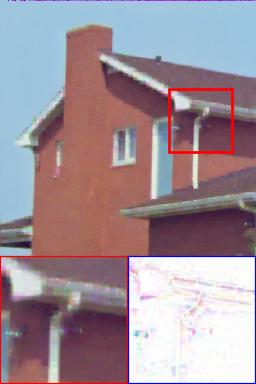}&
\includegraphics[width=14.8mm, height = 21mm]{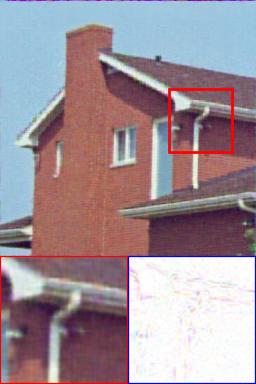}&
\includegraphics[width=14.8mm, height = 21mm]{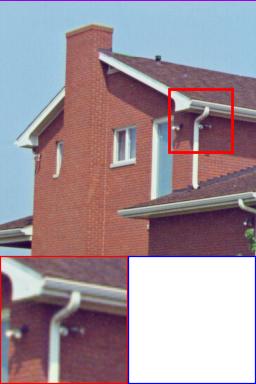}\\
\scriptsize{\tiny 8.44/0.125} & \scriptsize{\tiny 27.52/0.869} & \scriptsize{\tiny 28.00/0.835} & \scriptsize{\tiny \underline{29.82}/0.881} & \scriptsize{\tiny 29.12/0.868} & \scriptsize{\tiny 28.05/0.868} & \scriptsize{\tiny 25.67/0.826} & \scriptsize{\tiny 29.03/0.912} & \scriptsize{\tiny 27.54/0.858} & \scriptsize{\tiny 29.05/\underline{0.914}} & \scriptsize{\tiny \textbf{32.54}/\textbf{0.941}} & \scriptsize{\tiny PSNR/SSIM}\\
\includegraphics[width=14.8mm, height = 21mm]{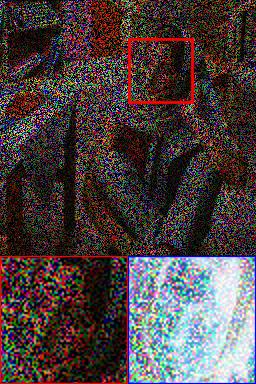}&
\includegraphics[width=14.8mm, height = 21mm]{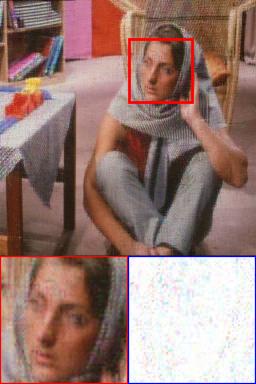}&
\includegraphics[width=14.8mm, height = 21mm]{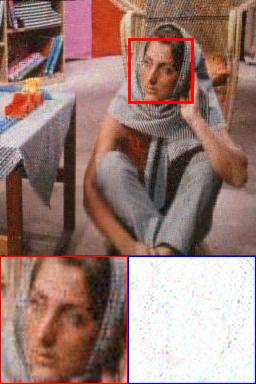}&
\includegraphics[width=14.8mm, height = 21mm]{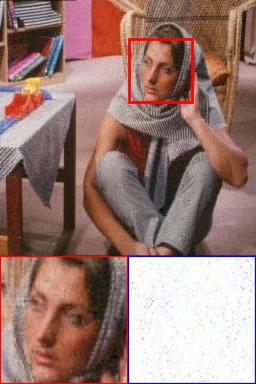}&
\includegraphics[width=14.8mm, height = 21mm]{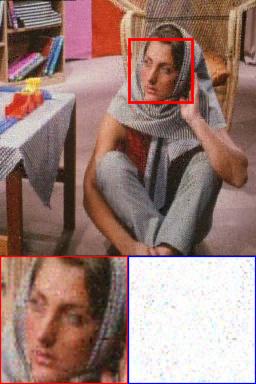}&
\includegraphics[width=14.8mm, height = 21mm]{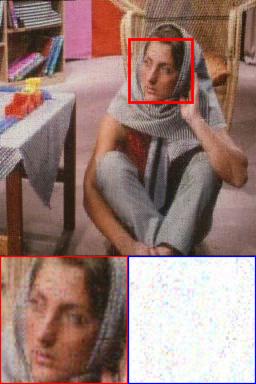}&
\includegraphics[width=14.8mm, height = 21mm]{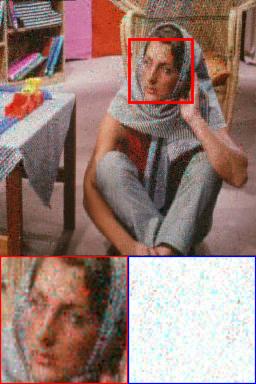}&
\includegraphics[width=14.8mm, height = 21mm]{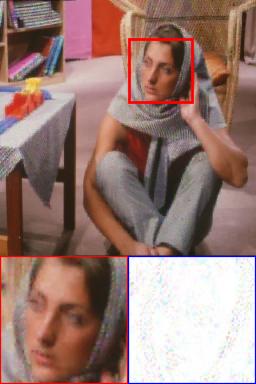}&
\includegraphics[width=14.8mm, height = 21mm]{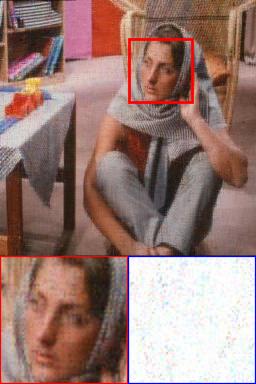}&
\includegraphics[width=14.8mm, height = 21mm]{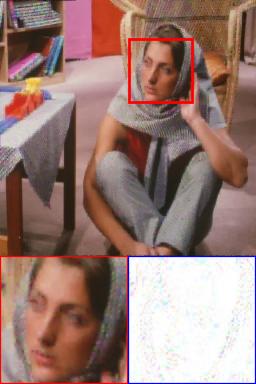}&
\includegraphics[width=14.8mm, height = 21mm]{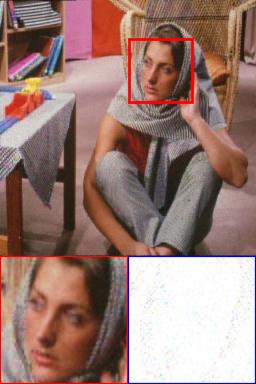}&
\includegraphics[width=14.8mm, height = 21mm]{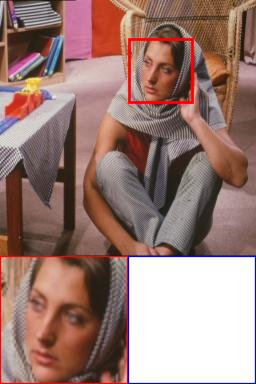}\\
\scriptsize \textbf{Observed} & \scriptsize SNN & \scriptsize BCPF & \scriptsize KBR & \scriptsize IRTNN & \scriptsize TNN & \scriptsize MF-TV & \scriptsize SNN-TV & \scriptsize SPC-TV &\scriptsize TNN-TV & \scriptsize \textbf{t-CTV} & \scriptsize \textbf{Ground truth}\\
\end{tabular}
\vspace{-0.2cm}
\caption{Color image inpainting results by all competing methods. From top to bottom: SR equals $10\%$, $20\%$, and $40\%$, respectively. For better viewing, we display the magnified map of a patch and corresponding error map (difference from the ground truth) under each image. Error maps with less color information indicate better restoration performance.}\label{fig.6}
\vspace{-0.5cm}
\end{figure*}

\begin{figure}[!tbp]
\renewcommand{\arraystretch}{0.5}
\setlength\tabcolsep{0.5pt}
\centering
\begin{tabular}{cccccccc}
\centering
\includegraphics[width=10.8mm, height = 10.8mm]{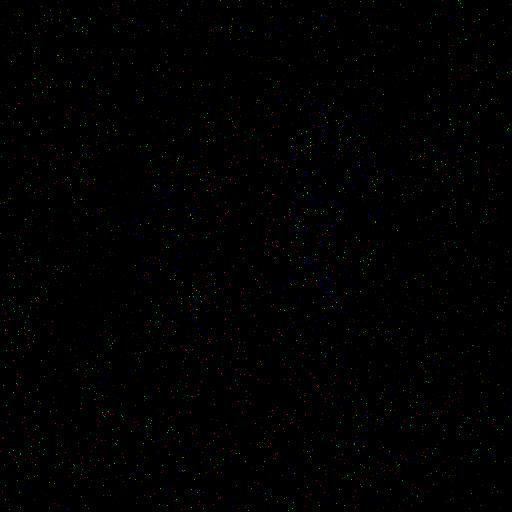}&
\includegraphics[width=10.8mm, height = 10.8mm]{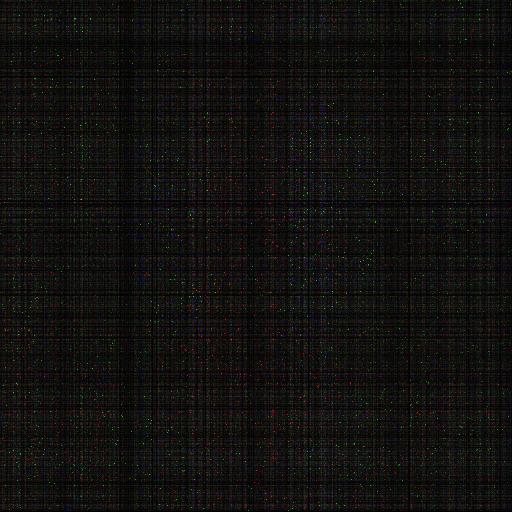}&
\includegraphics[width=10.8mm, height = 10.8mm]{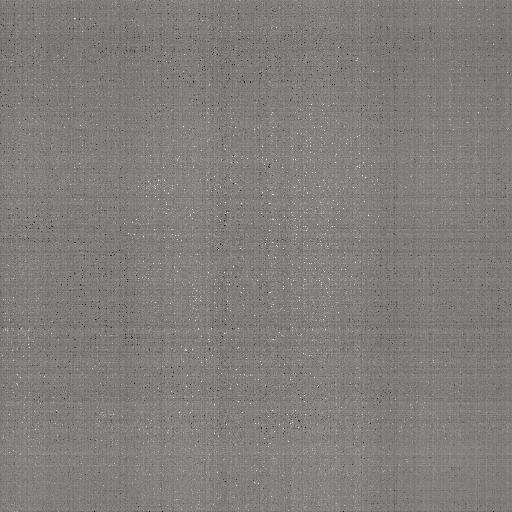}&
\includegraphics[width=10.8mm, height = 10.8mm]{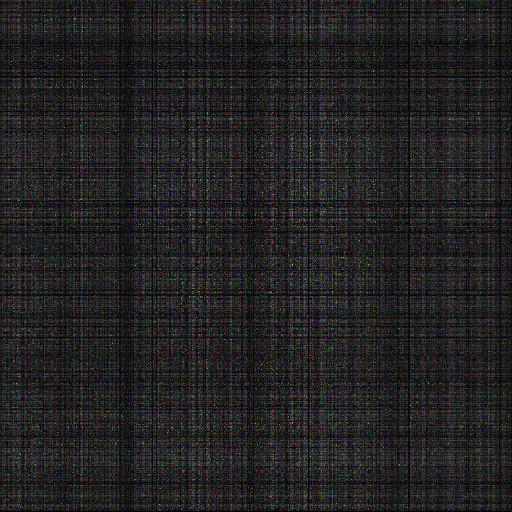}&
\includegraphics[width=10.8mm, height = 10.8mm]{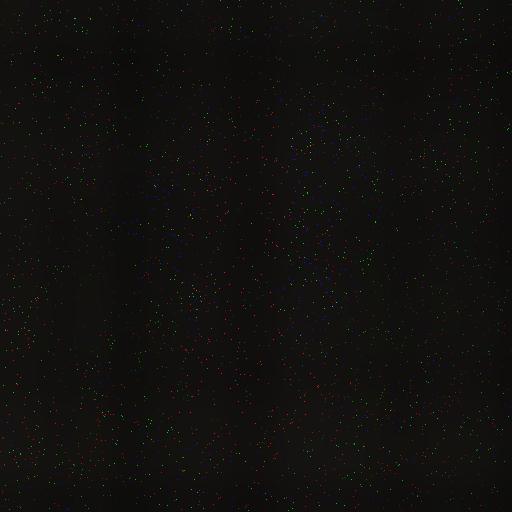}&
\includegraphics[width=10.8mm, height = 10.8mm]{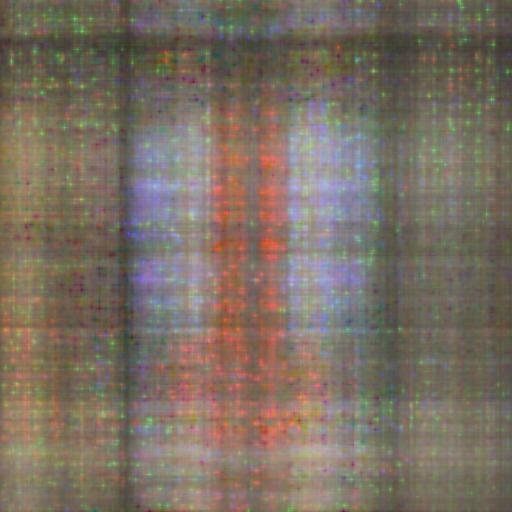}&
\includegraphics[width=10.8mm, height = 10.8mm]{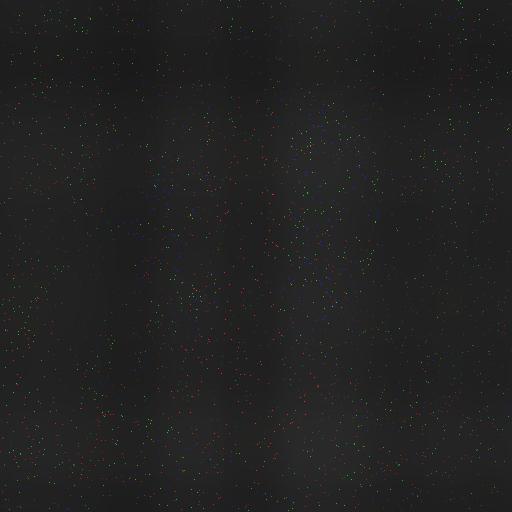}&
\includegraphics[width=10.8mm, height = 10.8mm]{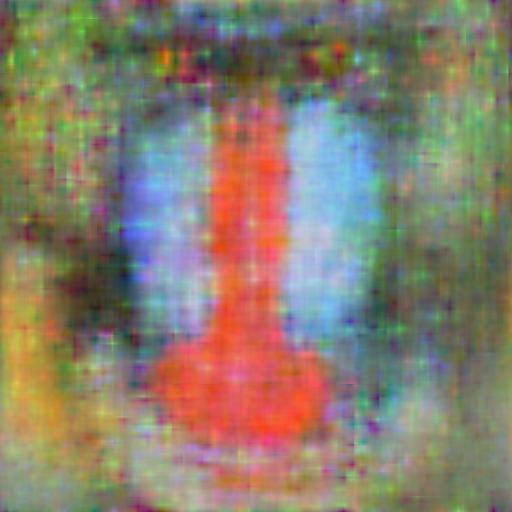}\\
\includegraphics[width=10.8mm, height = 10.8mm]{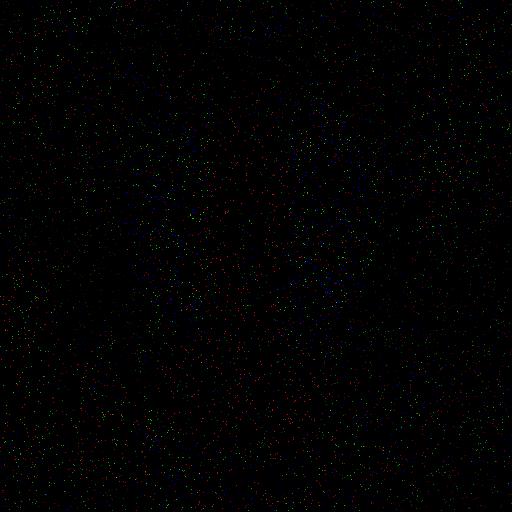}&
\includegraphics[width=10.8mm, height = 10.8mm]{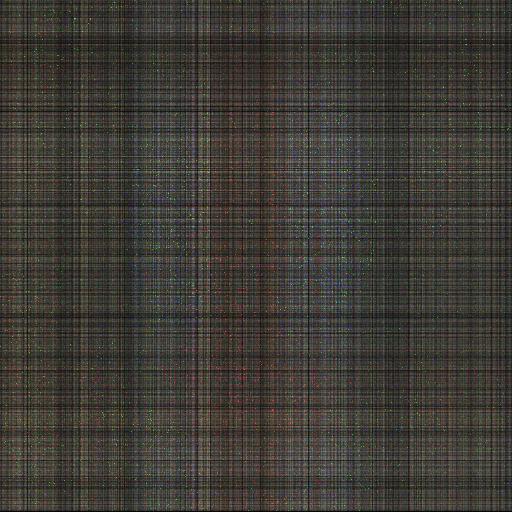}&
\includegraphics[width=10.8mm, height = 10.8mm]{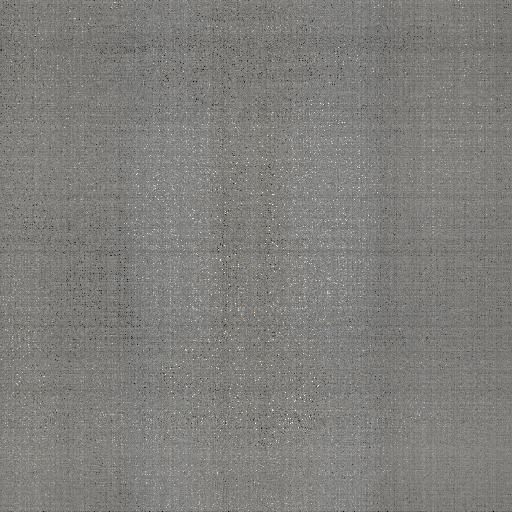}&
\includegraphics[width=10.8mm, height = 10.8mm]{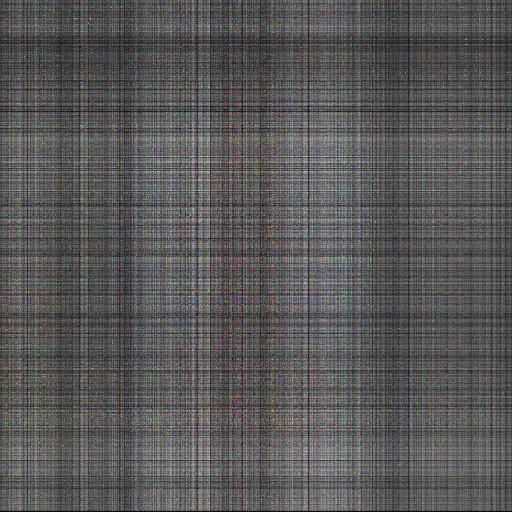}&
\includegraphics[width=10.8mm, height = 10.8mm]{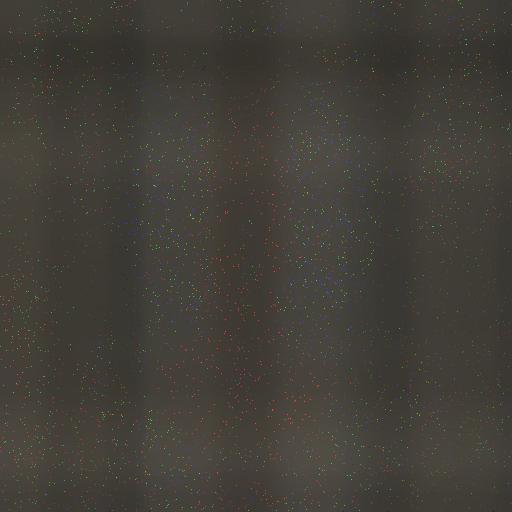}&
\includegraphics[width=10.8mm, height = 10.8mm]{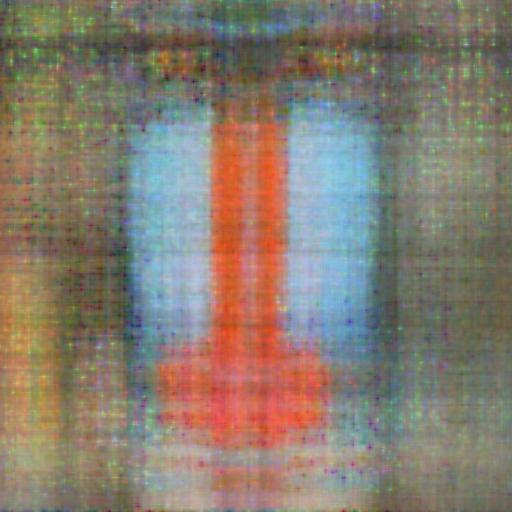}&
\includegraphics[width=10.8mm, height = 10.8mm]{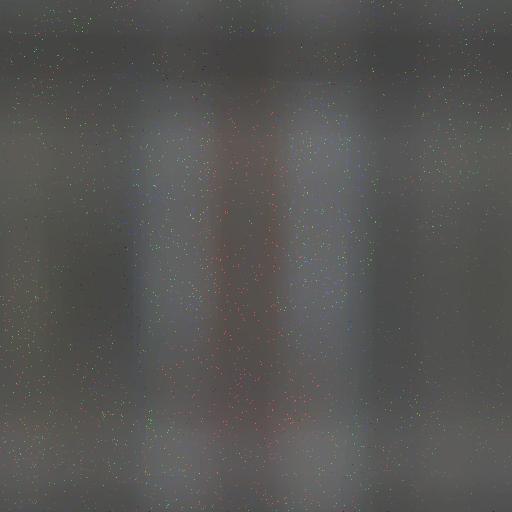}&
\includegraphics[width=10.8mm, height = 10.8mm]{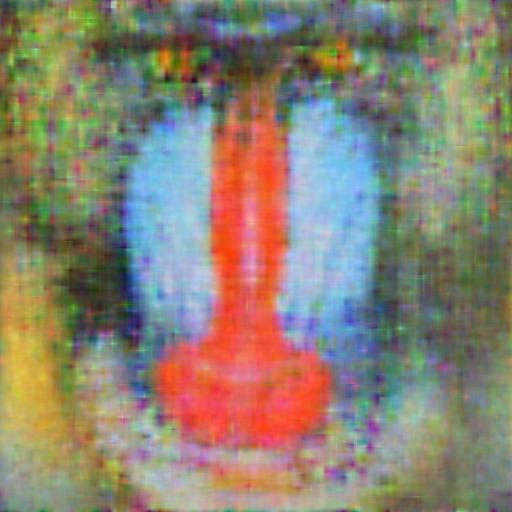}\\
\includegraphics[width=10.8mm, height = 10.8mm]{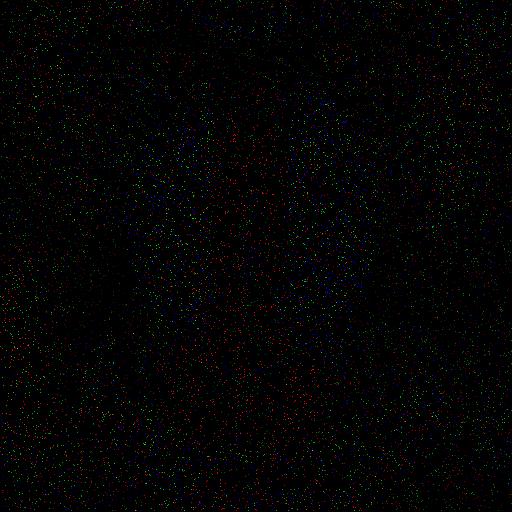}&
\includegraphics[width=10.8mm, height = 10.8mm]{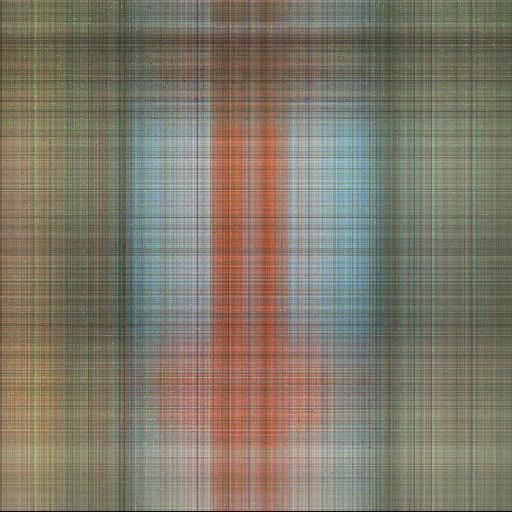}&
\includegraphics[width=10.8mm, height = 10.8mm]{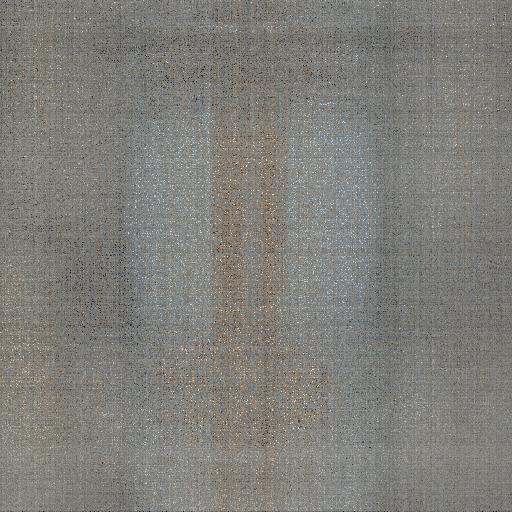}&
\includegraphics[width=10.8mm, height = 10.8mm]{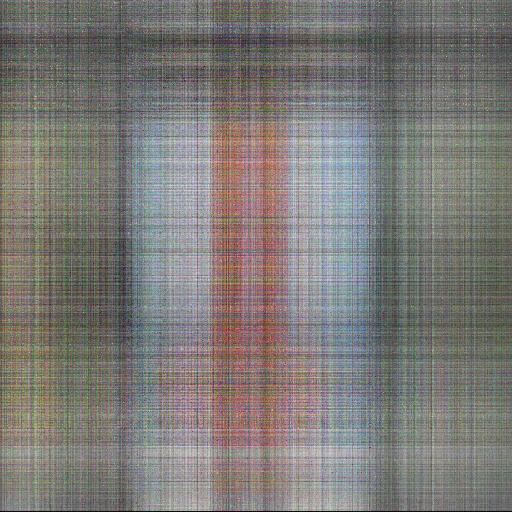}&
\includegraphics[width=10.8mm, height = 10.8mm]{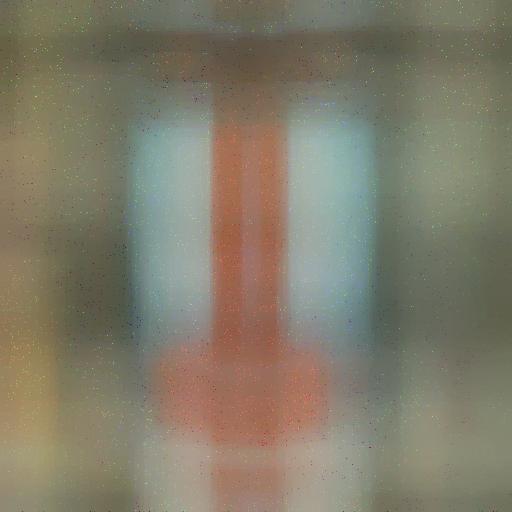}&
\includegraphics[width=10.8mm, height = 10.8mm]{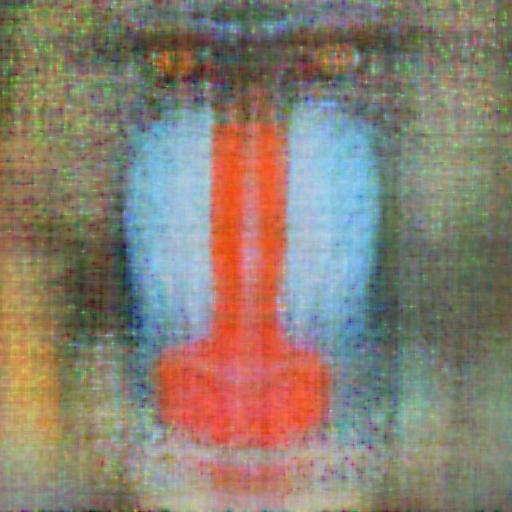}&
\includegraphics[width=10.8mm, height = 10.8mm]{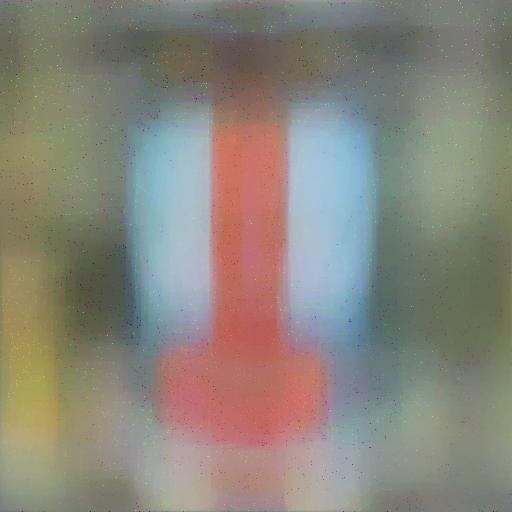}&
\includegraphics[width=10.8mm, height = 10.8mm]{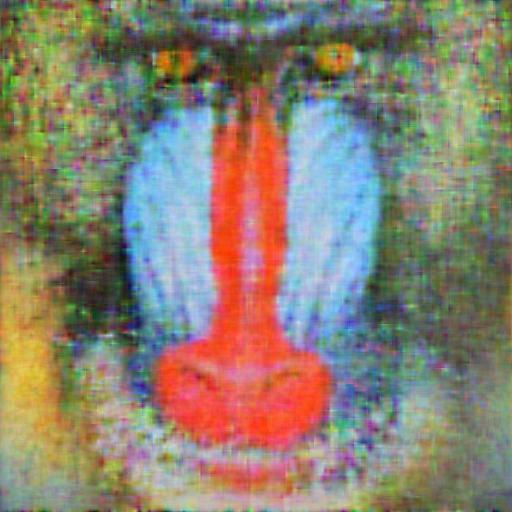}\\
\includegraphics[width=10.8mm, height = 10.8mm]{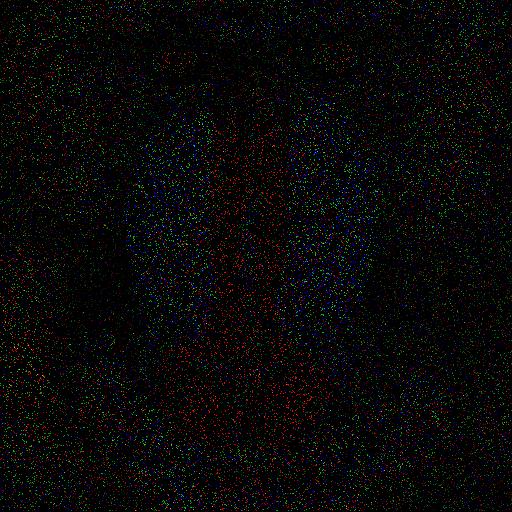}&
\includegraphics[width=10.8mm, height = 10.8mm]{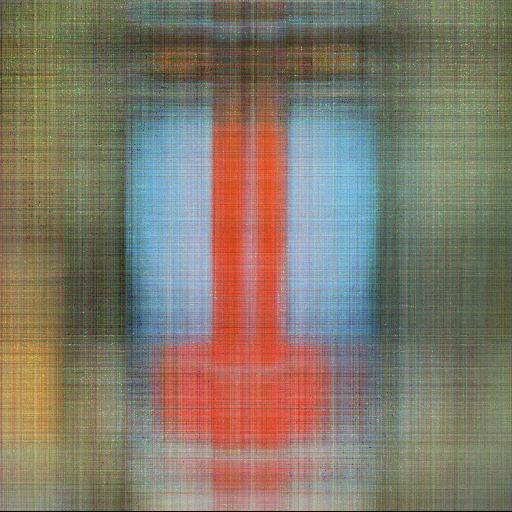}&
\includegraphics[width=10.8mm, height = 10.8mm]{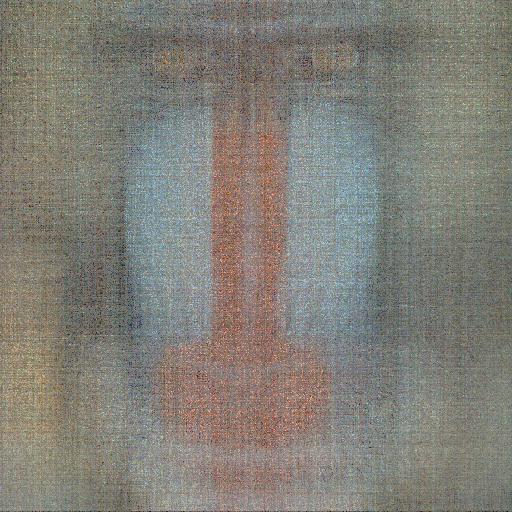}&
\includegraphics[width=10.8mm, height = 10.8mm]{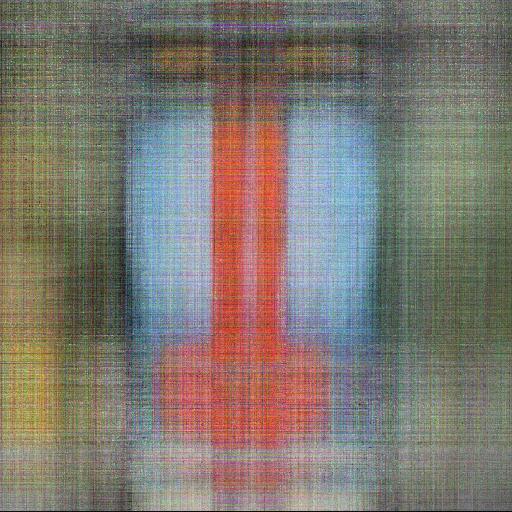}&
\includegraphics[width=10.8mm, height = 10.8mm]{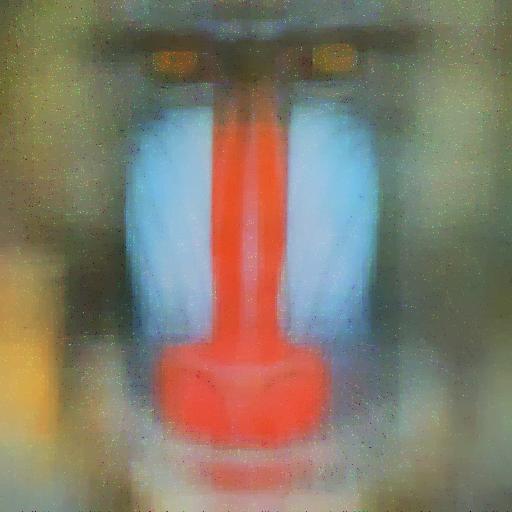}&
\includegraphics[width=10.8mm, height = 10.8mm]{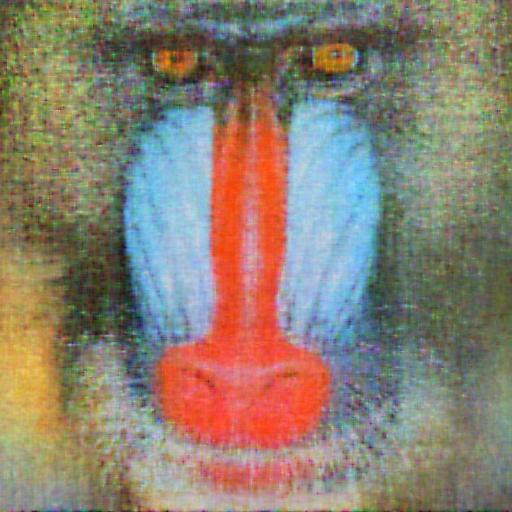}&
\includegraphics[width=10.8mm, height = 10.8mm]{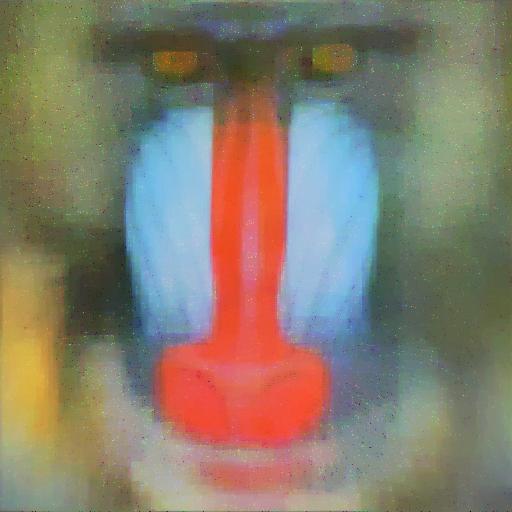}&
\includegraphics[width=10.8mm, height = 10.8mm]{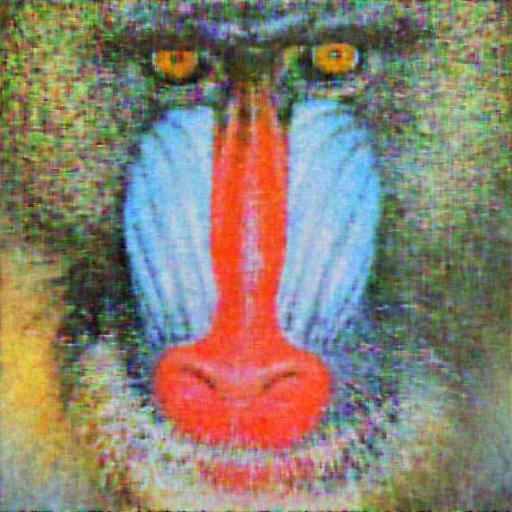}\\
\scriptsize \textbf{Observed} & \scriptsize SNN &  \scriptsize KBR & \scriptsize TNN  & \scriptsize SNN-TV & \scriptsize SPC-TV &\scriptsize TNN-TV & \scriptsize \textbf{t-CTV} \\
\end{tabular}
\vspace{-0.2cm}
\caption{Color image inpainting results obtained by all competing methods. From top to bottom: SR $=0.5\%, 1\%, 2\%,5\%$.}\label{fig.7}
\vspace{-0.3cm}
\end{figure}

\begin{figure*}[tp]
\renewcommand{\arraystretch}{0.5}
\setlength\tabcolsep{0.5pt}
\centering
\begin{tabular}{cccccccccccc}
\centering
\scriptsize{ \tiny 12.08/0.336} & \scriptsize{ \tiny 12.14/0.332} & \scriptsize{ \tiny 12.11/0.336} & \scriptsize{ \tiny 12.08/0.336} & \scriptsize{ \tiny 12.14/0.338} & \scriptsize{ \tiny 14.59/0.383} & \scriptsize{ \tiny 15.63/0.408} & \scriptsize{ \tiny 35.52/0.937} & \scriptsize{ \tiny 12.57/0.345} & \scriptsize{ \tiny \underline{35.61}/\underline{0.940}} & \scriptsize{ \tiny \textbf{37.51}/\textbf{0.956}} & \scriptsize{ \tiny PSNR/SSIM}\\
\includegraphics[width=14.8mm, height = 14.8mm]{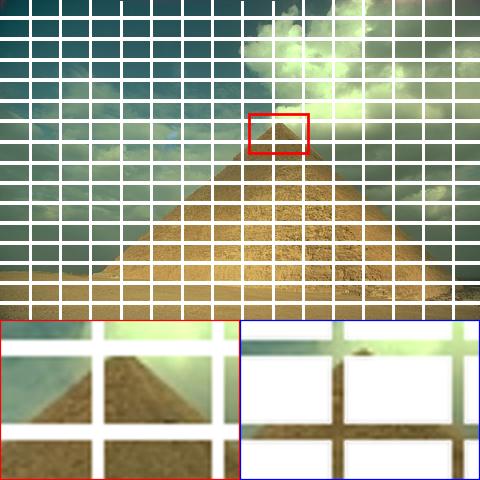}&
\includegraphics[width=14.8mm, height = 14.8mm]{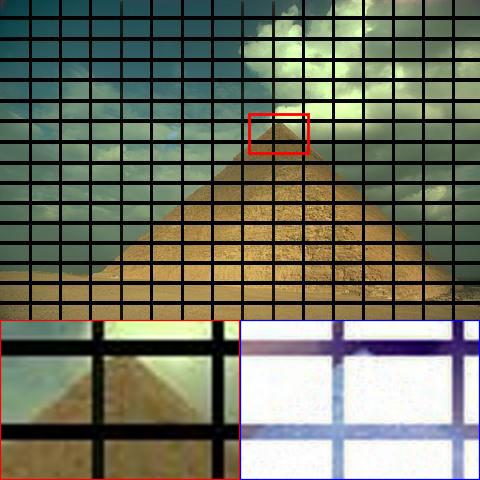}&
\includegraphics[width=14.8mm, height = 14.8mm]{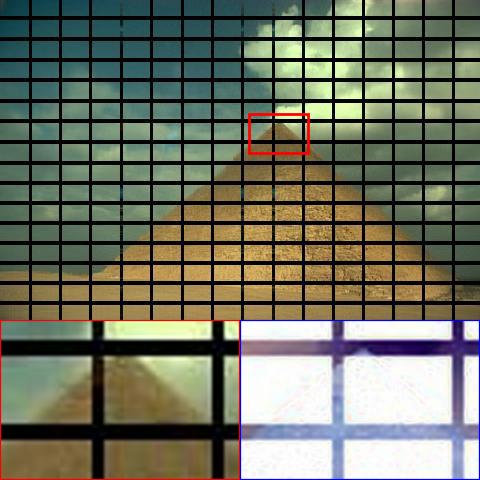}&
\includegraphics[width=14.8mm, height = 14.8mm]{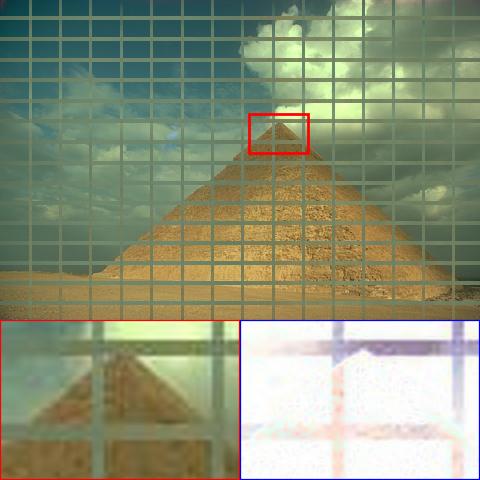}&
\includegraphics[width=14.8mm, height = 14.8mm]{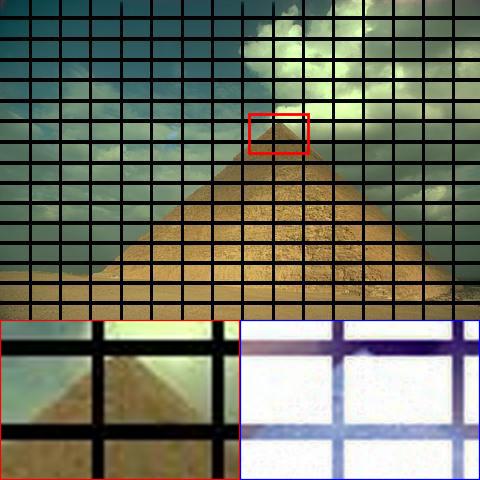}&
\includegraphics[width=14.8mm, height = 14.8mm]{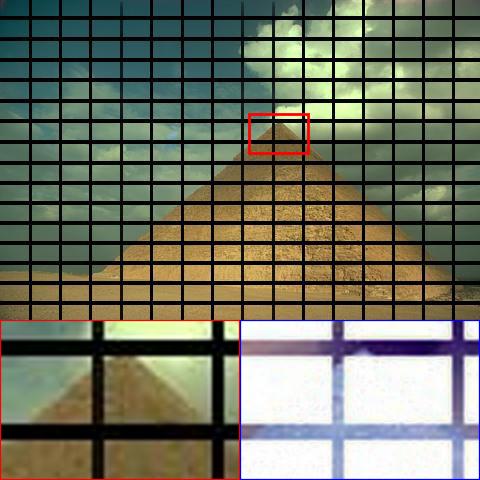}&
\includegraphics[width=14.8mm, height = 14.8mm]{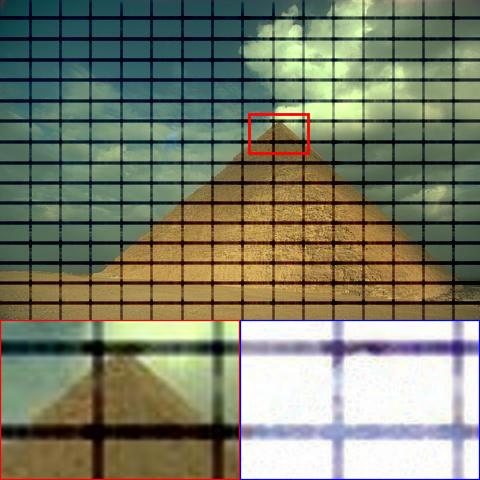}&
\includegraphics[width=14.8mm, height = 14.8mm]{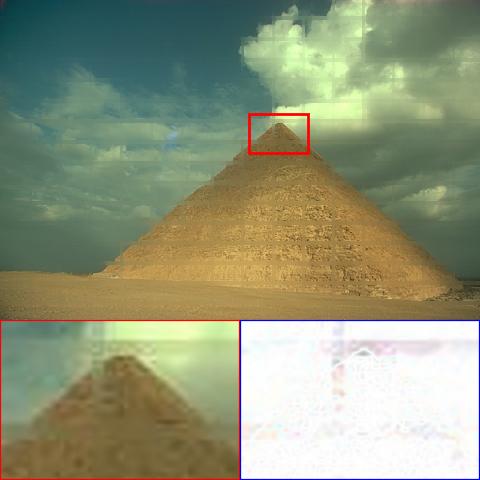}&
\includegraphics[width=14.8mm, height = 14.8mm]{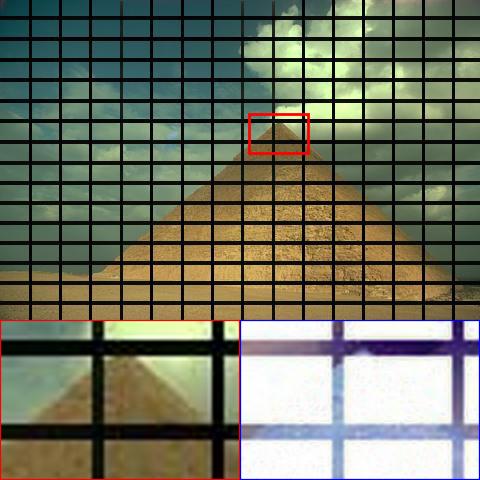}&
\includegraphics[width=14.8mm, height = 14.8mm]{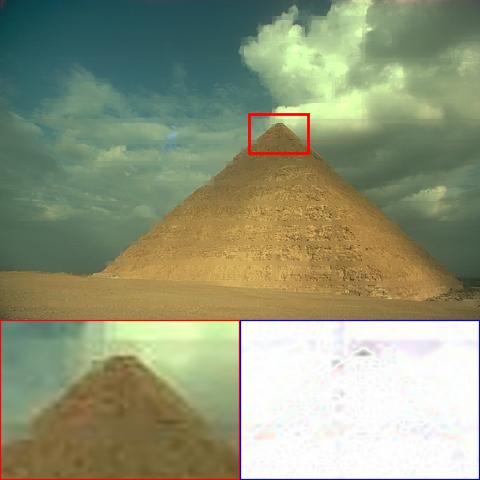}&
\includegraphics[width=14.8mm, height = 14.8mm]{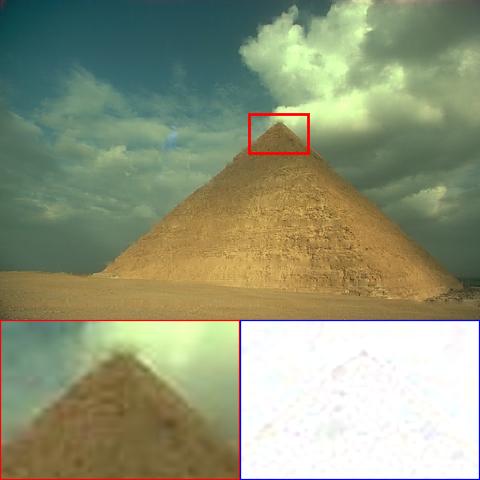}&
\includegraphics[width=14.8mm, height = 14.8mm]{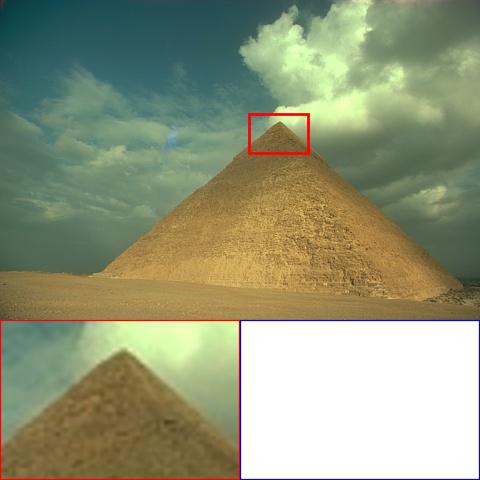}\\
\scriptsize{ \tiny 16.90/0.794} & \scriptsize{ \tiny 24.55/0.856} & \scriptsize{ \tiny 21.48/0.778} & \scriptsize{ \tiny 23.30/0.836} & \scriptsize{ \tiny 22.55/0.842} & \scriptsize{ \tiny 24.48/0.854} & \scriptsize{ \tiny 19.10/0.821} & \scriptsize{ \tiny 26.01/0.884} & \scriptsize{ \tiny \underline{26.24}/\underline{0.886}} & \scriptsize{ \tiny 25.71/0.866} & \scriptsize{ \tiny \textbf{27.36}/\textbf{0.896}} & \scriptsize{ \tiny PSNR/SSIM}\\
\includegraphics[width=14.8mm, height = 14.8mm]{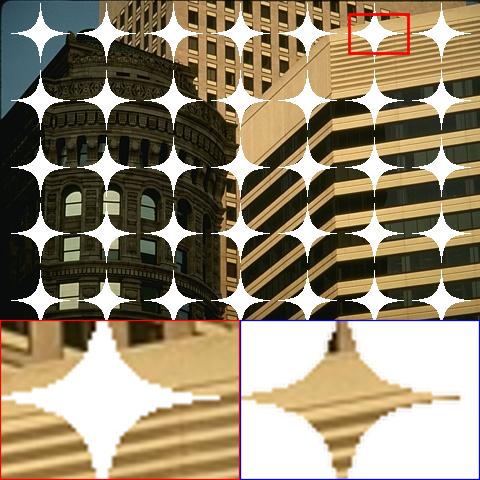}&
\includegraphics[width=14.8mm, height = 14.8mm]{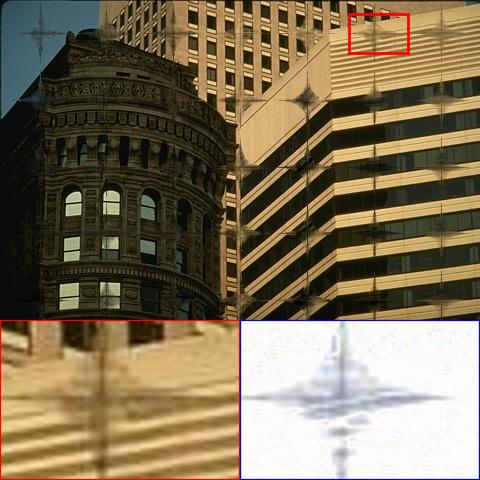}&
\includegraphics[width=14.8mm, height = 14.8mm]{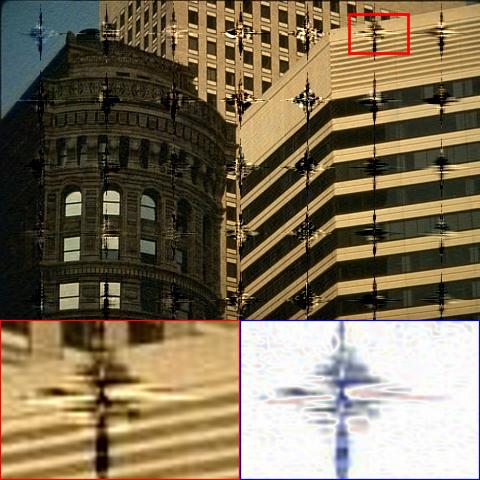}&
\includegraphics[width=14.8mm, height = 14.8mm]{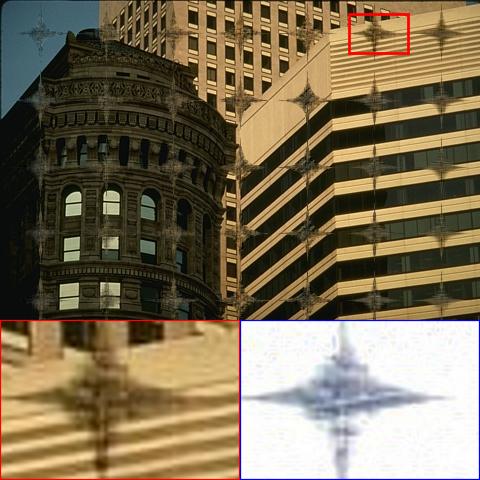}&
\includegraphics[width=14.8mm, height = 14.8mm]{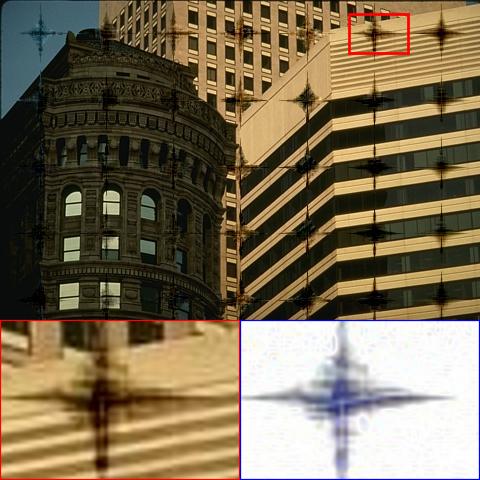}&
\includegraphics[width=14.8mm, height = 14.8mm]{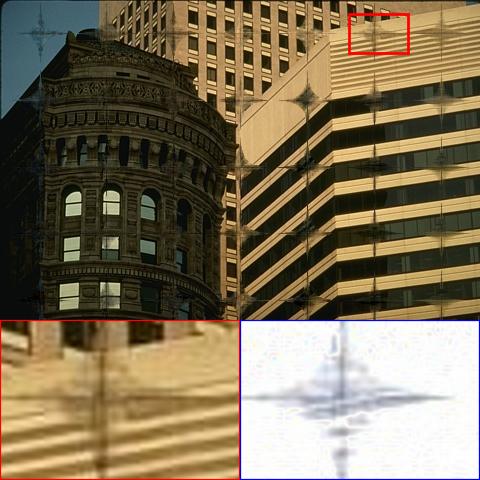}&
\includegraphics[width=14.8mm, height = 14.8mm]{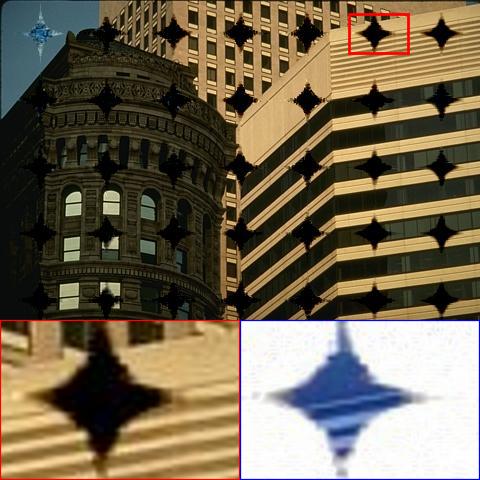}&
\includegraphics[width=14.8mm, height = 14.8mm]{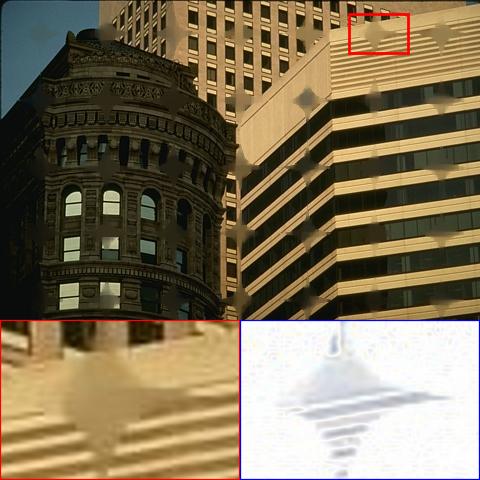}&
\includegraphics[width=14.8mm, height = 14.8mm]{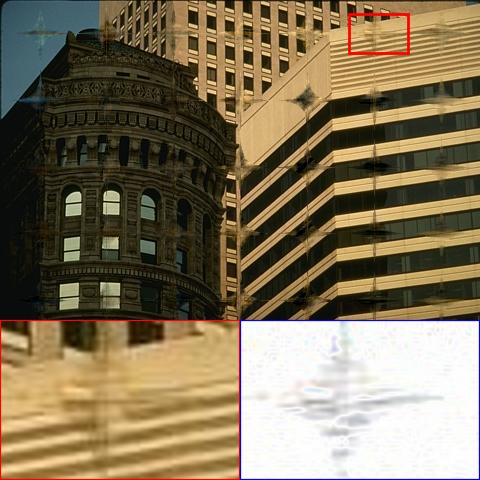}&
\includegraphics[width=14.8mm, height = 14.8mm]{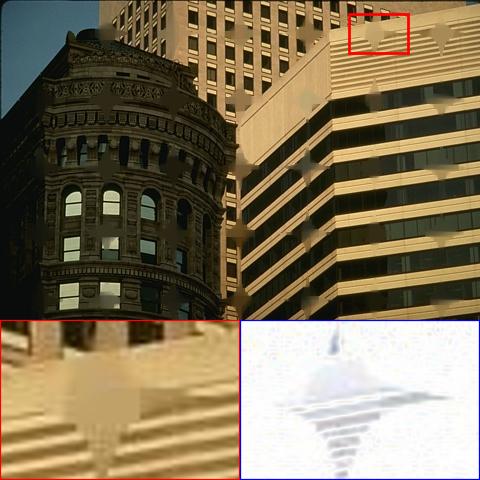}&
\includegraphics[width=14.8mm, height = 14.8mm]{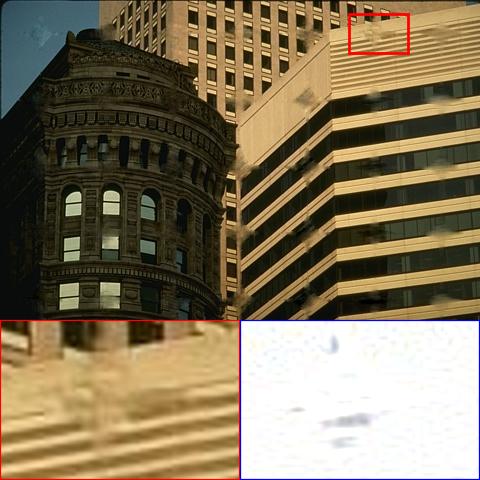}&
\includegraphics[width=14.8mm, height = 14.8mm]{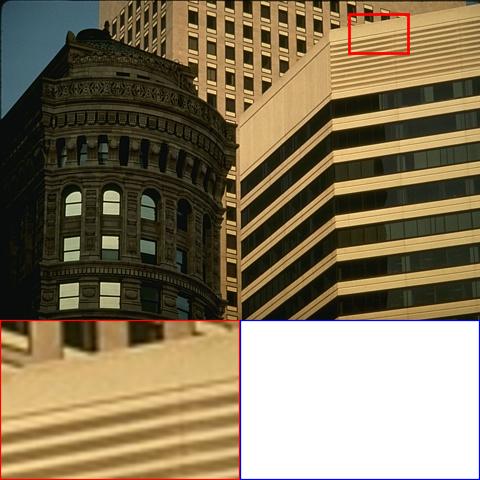}\\
\scriptsize{ \tiny 14.49/0.469} & \scriptsize{ \tiny 22.85/0.742} & \scriptsize{ \tiny 21.75/0.657} & \scriptsize{ \tiny 22.17/0.680} & \scriptsize{ \tiny 22.58/0.697} & \scriptsize{ \tiny 22.70/0.725} & \scriptsize{ \tiny 23.09/0.695} & \scriptsize{ \tiny 24.67/\underline{0.855}} & \scriptsize{ \tiny 23.20/0.754} & \scriptsize{ \tiny \underline{24.93}/0.813} & \scriptsize{ \tiny \textbf{26.18}/\textbf{0.867}} & \scriptsize{ \tiny PSNR/SSIM}\\
\includegraphics[width=14.8mm, height = 14.8mm]{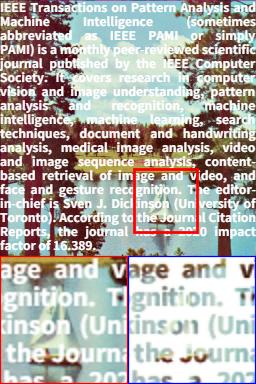}&
\includegraphics[width=14.8mm, height = 14.8mm]{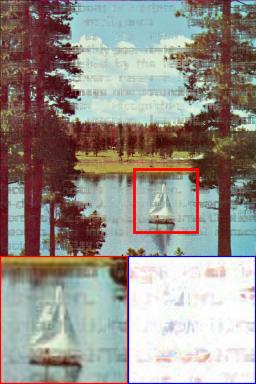}&
\includegraphics[width=14.8mm, height = 14.8mm]{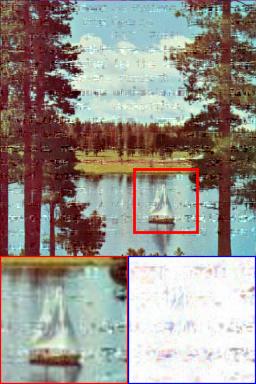}&
\includegraphics[width=14.8mm, height = 14.8mm]{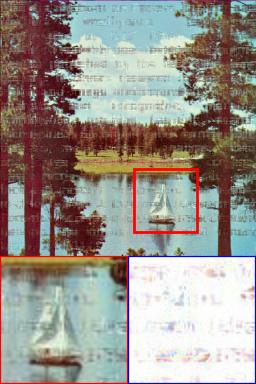}&
\includegraphics[width=14.8mm, height = 14.8mm]{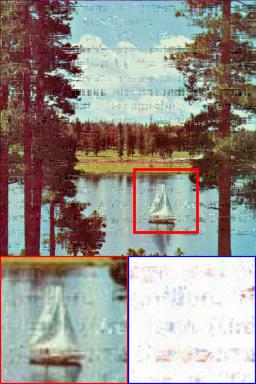}&
\includegraphics[width=14.8mm, height = 14.8mm]{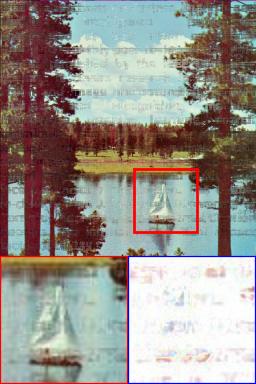}& \includegraphics[width=14.8mm, height = 14.8mm]{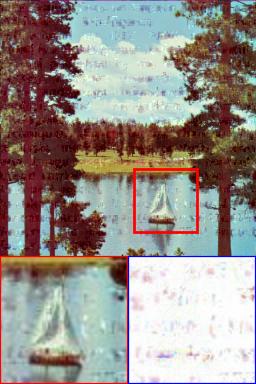}&
\includegraphics[width=14.8mm, height = 14.8mm]{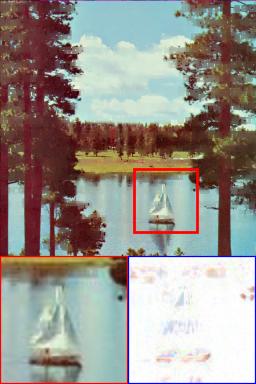}&
\includegraphics[width=14.8mm, height = 14.8mm]{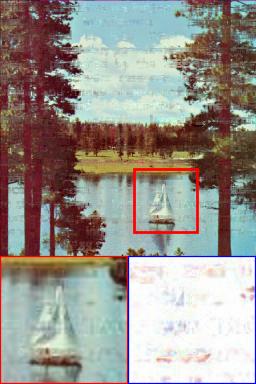}&
\includegraphics[width=14.8mm, height = 14.8mm]{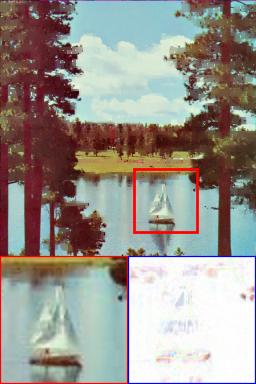}&
\includegraphics[width=14.8mm, height = 14.8mm]{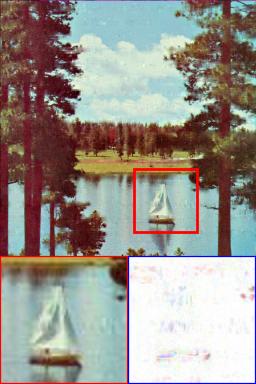}&
\includegraphics[width=14.8mm, height = 14.8mm]{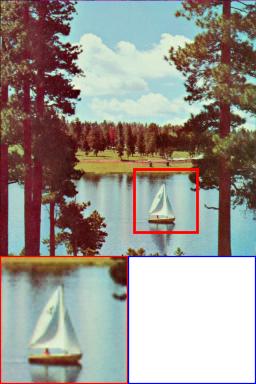}\\
\scriptsize \textbf{Observed} & \scriptsize SNN & \scriptsize BCPF & \scriptsize KBR & \scriptsize IRTNN & \scriptsize TNN & \scriptsize MF-TV & \scriptsize SNN-TV & \scriptsize SPC-TV &\scriptsize TNN-TV & \scriptsize \textbf{t-CTV} & \scriptsize \textbf{Ground truth}\\
\end{tabular}
\vspace{-0.2cm}
\caption{Color image inpainting results by all competing methods on three types of structured masked images.}\label{fig.8}
\vspace{-0.5cm}
\end{figure*}

\begin{figure}[tp]
\centering
\begin{tabular}{@{}c@{}@{}c@{}}
\centering
\subfloat{\includegraphics[width=45mm, height = 30mm]{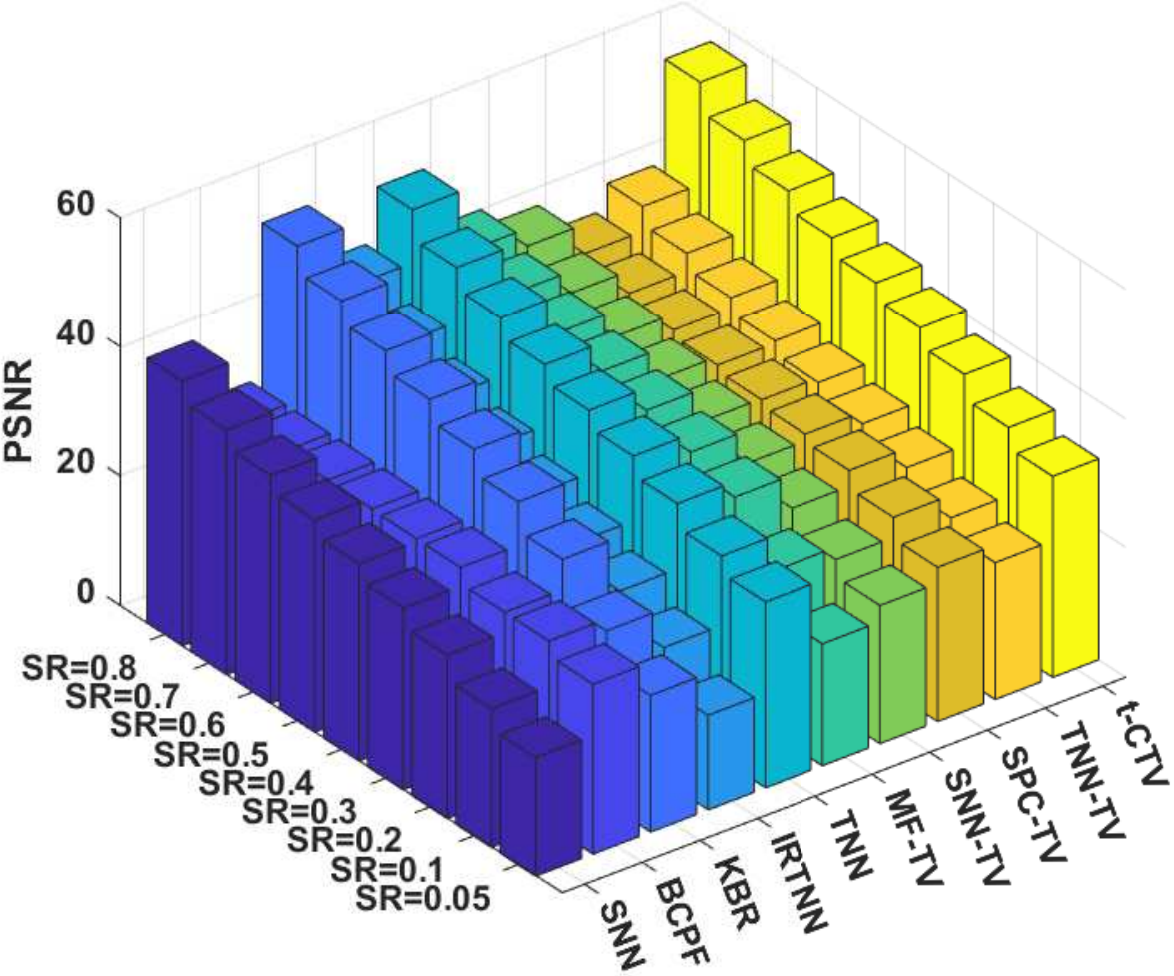}}&
\subfloat{\includegraphics[width=45mm, height = 30mm]{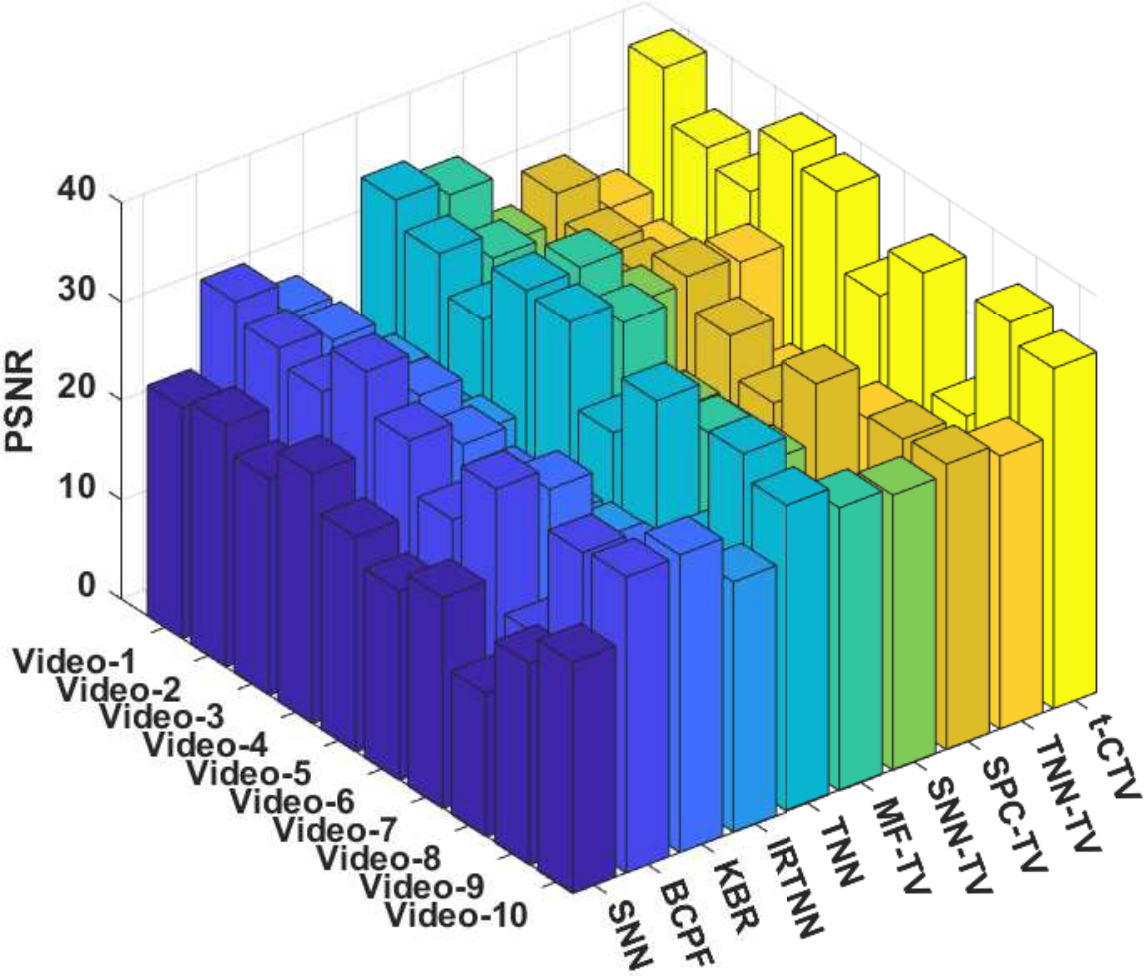}}\\
\specialrule{0em}{-6pt}{4pt}
\small (a) All SR cases &  \small (b) SR $=10\%$ \\
\specialrule{0em}{-6pt}{4pt}
\end{tabular}
\vspace{-0.2cm}
\caption{Performance comparison in terms of PSNR of recovered color videos obtained by all competing methods.}\label{fig.9}
\vspace{-0.2cm}
\end{figure}

\begin{table}[!htbp]
\renewcommand{\arraystretch}{1.15}
\setlength\tabcolsep{3.0pt}
\footnotesize
  \caption{
  ERGAS comparison of all competing methods on HSIs inpainting under SR $=5\%$.
  }\label{table.7}
  \setlength{\abovecaptionskip}{5pt}
  \setlength{\belowcaptionskip}{5pt}
  \centering
  \vspace{-0.2cm}
\begin{tabular}{l||c|c|c|c|c|c}
     \Xhline{1pt}
      \multirow{2}{*}{Method} & \multicolumn{5}{c|}{HSIs Data} & \multirow{2}{*}{Average}\\
     \cline{2-6}
     \qquad & Cuprite & DCMall & KSC & Pavia & PaviaU & \qquad\\
     \Xhline{1pt}
     SNN & 153.6 & 451.4 & 240.0 & 411.0 & 374.9 & 326.2\\
     BCPF & 77.62 & 208.9 & 132.9 & 148.4 & 128.2 & 139.2\\
     KBR & 44.19 & \underline{168.6} & \underline{94.78} & \underline{120.3} & \underline{101.8} & \underline{105.9}\\
     IRTNN & 928.1 & 841.8 & 535.2 & 361.1 & 404.2 & 614.1\\
     TNN & 53.64 & 245.9 & 176.0 & 128.4 & 124.5 & 145.7\\
     MF-TV & \underline{40.25} & 564.5 & 433.3 & 444.2 & 413.0 & 379.0\\
     SNN-TV & 135.7 & 453.0 & 211.0 & 361.6 & 321.8 & 296.6\\
     SPC-TV & 84.33 & 262.5 & 156.3 & 206.0 & 182.6 & 178.3\\
     TNN-TV & 127.4 & 422.9 & 205.0 & 305.8 & 286.6 & 269.6\\
     \textbf{t-CTV} & \textbf{25.16} & \textbf{119.9} & \textbf{75.73} & \textbf{70.01} & \textbf{59.91} & \textbf{70.15}\\
     \Xhline{1pt}
\end{tabular}
\vspace{-0.2cm}
\end{table}

\begin{figure}[!htbp]
\renewcommand{\arraystretch}{0.5}
\setlength\tabcolsep{0.5pt}
\centering
\vspace{-0.2cm}
\begin{tabular}{cccc}
\centering
\tiny 9.47/0.017/1014 & \tiny 16.85/0.332/451.4 & \tiny 23.59/0.759/208.9 & \tiny \underline{25.56}/\underline{0.841}/\underline{168.6}\\
\includegraphics[width=21.8mm, height = 21.8mm]{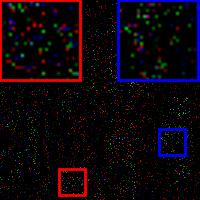}&
\includegraphics[width=21.8mm, height = 21.8mm]{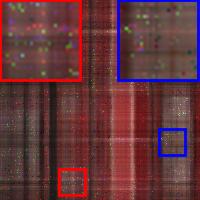}&
\includegraphics[width=21.8mm, height = 21.8mm]{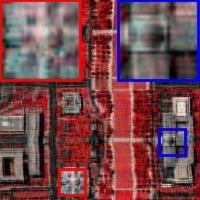}&
\includegraphics[width=21.8mm, height = 21.8mm]{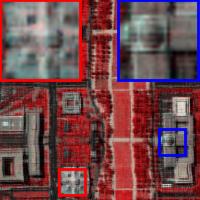}\\
\scriptsize \textbf{Observed} & \scriptsize SNN & \scriptsize BCPF & \scriptsize KBR\\
\tiny 11.45/0.095/841.8 & \tiny 22.49/0.717/245.9 & \tiny 18.18/0.491/564.5 & \tiny 16.85/0.244/453.0\\
\includegraphics[width=21.8mm, height = 21.8mm]{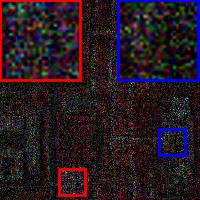}&
\includegraphics[width=21.8mm, height = 21.8mm]{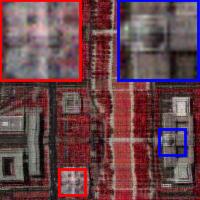}&
\includegraphics[width=21.8mm, height = 21.8mm]{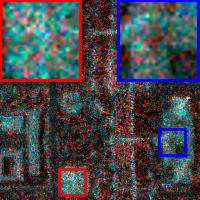}&
\includegraphics[width=21.8mm, height = 21.8mm]{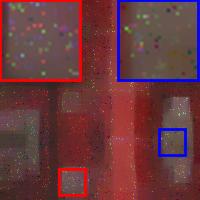}\\
\scriptsize IRTNN & \scriptsize TNN & \scriptsize MF-TV & \scriptsize SNN-TV\\
\tiny 21.64/0.675/262.5 & \tiny 17.44/0.291/422.9 & \tiny \textbf{29.04}/\textbf{0.929}/\textbf{119.9} & \tiny PSNR/SSIM/ERGAS\\
\includegraphics[width=21.8mm, height = 21.8mm]{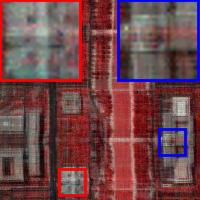}&
\includegraphics[width=21.8mm, height = 21.8mm]{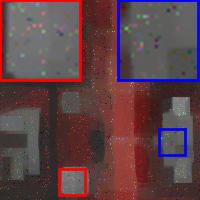}&
\includegraphics[width=21.8mm, height = 21.8mm]{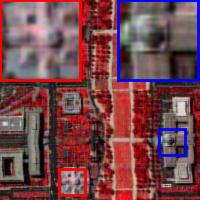}&
\includegraphics[width=21.8mm, height = 21.8mm]{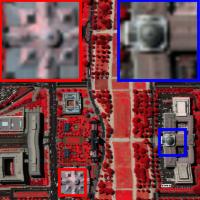}\\
\scriptsize SPC-TV &\scriptsize TNN-TV & \scriptsize \textbf{t-CTV} & \scriptsize \textbf{Ground truth}\\
\end{tabular}
\vspace{-0.2cm}
\caption{HSI inpainting results under SR $=5\%$. The displayed pseudo-color image uses bands 49-27-7 as R-G-B.}\label{fig.11}
\end{figure}

\textbf{2) Results on color videos}: In this experiment, we test all competing methods on 10 color video sequences from the YUV\footnote{\url{http://trace.eas.asu.edu/yuv/}} database. All videos are of size $176\times144\times3\times200$. Note that the MF-TV and IRTNN methods cannot be directly used for order-4 tensors. One can reshape a video as $176\times144\times600$ or processing it frame-by-frame as an order-3 tensor. Since the former is very time-consuming and the performance of most methods on it is poor, we choose the latter in experiments. Fig.\;\ref{fig.9}(a) plots the average PSNR values of different methods under different levels of SR, and Fig.\;\ref{fig.9}(b) plots the detailed PSNR values of each recovered video using different methods when SR $=10\%$. Both bar graphs evidently show that our t-CTV method obtains the leading performance over other competing methods distinctly.

\textbf{3) Results on hyperspectral images (HSIs)}: We choose five widely used HSIs, including Cuprite, DCMall, KSC, Pavia and Pavia university (PaviaU), in this experiment. All HSIs are preprocessed with size $200\times200\times50$. Similarly, we test the inpainting task under different levels of SR. Table \ref{table.7} lists the detailed recovery in terms of \textit{erreur relative globale adimensionnelle de synthese} (ERGAS \cite{wald2002data}) when SR $=5\%$. The smaller value of ERGAS indicates better restoration performance. Fig.\;\ref{fig.11} shows the pseudo-color image of the recovered HSIs via different methods. It is seen that our method performs better than all other competitors from the enlarged local area.

\textbf{4) Results on more visual data in high-order tensor format}: We then implement experiments on more visual data in tensor format, including order-$3$ CT and MRI medical images sized $256\times256\times127$\footnote{\url{https://www.cancerimagingarchive.net/}}, order-$4$ hyperspectral videos sized $188\times120\times33\times31$\footnote{\url{http://openremotesensing.net/knowledgebase/hyperspectral-video}} and order-$5$ light field images sized $216\times324\times3\times17\times17$\footnote{\url{http://lightfield.stanford.edu/lfs.html}}. Table \ref{table.8} lists the mean PSNR (MPSNR), mean SSIM (MSSIM) along with the third or more dimensions, and ERGAS values of the recovered data under SR $=10\%,30\%$ and $50\%$. These results consistently substantiate the advantages and potentials of our method for various high-order visual tensor data recovery. Results on some typical examples are shown in Fig.\;\ref{fig.13}, Fig.\;\ref{fig.14} and Fig.\;\ref{fig.15} for better visualization. It should be indicated that all current competing methods take a relatively long running time in these tasks. Thus, exploring faster algorithms in handling high-order high-dimensional tensor should be a very meaningful topic in future research, e.g., parallelizing design \cite{yao2018large}, accelerating algorithm \cite{li2020accelerated}, and so on.

\begin{table}[!htbp]
\renewcommand{\arraystretch}{1.15}
\setlength\tabcolsep{3.0pt}
\footnotesize
  \caption{
   More data inpainting performances of all competing methods under different sampling rates. (/m: minute).
  }\label{table.8}
  \setlength{\abovecaptionskip}{5pt}
  \setlength{\belowcaptionskip}{5pt}
  \centering
  \vspace{-0.2cm}
\begin{tabular}{@{}c@{}|@{}c@{}||c|c|c|@{}c@{}|@{}c@{}|@{}c@{}|@{}c@{}}
     \Xhline{1pt}
     SR & Metric & SNN  & KBR  & TNN & SNN-TV & SPC-TV & TNN-TV & \textbf{t-CTV}\\
     \Xhline{1pt}
     \multicolumn{9}{c}{Order-$3$ CT$\&$MRI Medical Images}\\
     \hline
     \hline
     \multirow{3}{*}{$10\%$} & MPSNR & 19.90 & \underline{25.86} & 22.44 & 23.23 & 21.79 & 22.89 & \textbf{26.48} \\
     \qquad & MSSIM & 0.533 & 0.710 & 0.458 & \underline{0.716} & 0.478 & 0.675 & \textbf{0.839}\\
     \qquad & ERGAS & 339.8 & \underline{184.9} & 266.8 & 236.1 & 277.5 & 252.5 & \textbf{169.4}\\
     \hline
     \multirow{3}{*}{$30\%$} & MPSNR  & 25.63 & \textbf{32.23} & 27.07 & 28.66 & 25.83 & 28.60 & \underline{30.70}\\
     \qquad & MSSIM & 0.766 & \textbf{0.896} & 0.699 & 0.870 & 0.632 & 0.875 & \underline{0.879}\\
     \qquad & ERGAS & 180.2 & \textbf{89.80} & 160.2 & 128.7 & 177.9 & 132.4 & \underline{106.4}\\
     \hline
     \multirow{3}{*}{$50\%$} & MPSNR & 30.18 & \textbf{36.07} & 30.94 & 32.38 & 28.09 &32.59 & \underline{34.14}\\
     \qquad & MSSIM & 0.892 & \textbf{0.952} & 0.839 & 0.934 & 0.726 & 0.937 & \underline{0.940}\\
     \qquad & ERGAS & 108.9 & \textbf{58.09} & 104.4 & 84.68 & 138.2 & 84.37 & \underline{72.84}\\
     \hline
     \multicolumn{2}{c||}{Time/m} & 7.41 & 17.08 & 4.11 & 11.25 & 53.29 & 17.24 & 17.96 \\
     \hline
     \hline
     \multicolumn{9}{c}{Order-$4$ Hyperspectral Videos}\\
     \hline
     \hline
     \multirow{3}{*}{$10\%$} & MPSNR & 25.47 & 33.34 & \underline{41.13} & 28.90 & 35.69 & 28.26 & \textbf{44.70}\\
     \qquad & MSSIM & 0.746 & 0.936 & \underline{0.977} & 0.816 & 0.943 & 0.825 & \textbf{0.988}\\
     \qquad & ERGAS & 148.5 & 60.01 & \underline{24.56} & 100.1 & 45.86 & 107.6 & \textbf{16.23}\\
     \hline
     \multirow{3}{*}{$30\%$} & MPSNR & 33.79 & \underline{49.07} & 48.41 & 35.96 & 37.62 & 36.87 & \textbf{50.89}\\
     \qquad & MSSIM & 0.927 & \underline{0.995} & 0.994 & 0.953 & 0.956 & 0.960 & \textbf{0.996}\\
     \qquad & ERGAS & 56.97 & \underline{9.824} & 10.59 & 44.41 & 36.71 & 39.98 & \textbf{7.956}\\
     \hline
     \multirow{3}{*}{$50\%$} & MPSNR & 38.99 & \textbf{54.44} & 52.89 & 41.15 & 39.31 & 42.10 & \underline{54.18}\\
     \qquad & MSSIM & 0.973 & \textbf{0.998} & \textbf{0.998} & 0.982 & 0.967 & \underline{0.985} & \textbf{0.998}\\
     \qquad & ERGAS & 31.33 & \textbf{5.293} & 6.369 & 24.44 & 30.25 & 21.92 & \underline{5.448}\\
     \hline
     \multicolumn{2}{c||}{Time/m} & 26.31 & 63.91 & 9.24 & 45.92 & 84.75 & 39.21 & 56.75 \\
     \hline
     \hline
     \multicolumn{9}{c}{Order-$5$ Light Field Images}\\
     \hline
     \hline
     \multirow{3}{*}{$10\%$} & MPSNR & 18.06 & 16.50 & \underline{28.58} & 20.64 & 21.40 & 20.00 & \textbf{30.32}\\
     \qquad & MSSIM & 0.554 & 0.379 & \underline{0.800} & 0.748 & 0.627 & 0.742 & \textbf{0.871}\\
     \qquad & ERGAS & 450.2 & 544.4 & \underline{137.4} & 335.5 & 308.9 & 361.9 & \textbf{111.9}\\
     \hline
     \multirow{3}{*}{$30\%$} & MPSNR & 23.60 & 24.82 & \underline{36.53} & 26.13 & 26.06 & 25.63 & \textbf{38.19}\\
     \qquad & MSSIM & 0.761 & 0.780 & \underline{0.945} & 0.892 & 0.747 & 0.890 & \textbf{0.965}\\
     \qquad & ERGAS & 238.9 & 210.8 & \underline{55.93} & 179.3 & 179.9 & 191.5 & \textbf{45.83}\\
     \hline
     \multirow{3}{*}{$50\%$} & MPSNR & 27.86 & 39.40 & \underline{42.44} & 30.43 & 28.02 & 30.07 & \textbf{43.73}\\
     \qquad & MSSIM & 0.884 & \underline{0.982} & 0.981 & 0.948 & 0.803 & 0.947 & \textbf{0.987}\\
     \qquad & ERGAS & 146.4 & 39.94 & \underline{28.25} & 109.4 & 143.2 & 115.7 & \textbf{24.38}\\
     \hline
     \multicolumn{2}{c||}{Time/m} & 77.21 & 202.8 & 27.70 & 640.24 & 417.5 & 99.61 & 128.5 \\
     \hline
     \hline
     \Xhline{1pt}
\end{tabular}
\end{table}

\begin{figure}[!htbp]
\renewcommand{\arraystretch}{0.5}
\setlength\tabcolsep{0.5pt}
\centering
\begin{tabular}{ccccccc}
\centering
\includegraphics[width=12.3mm, height = 12.3mm]{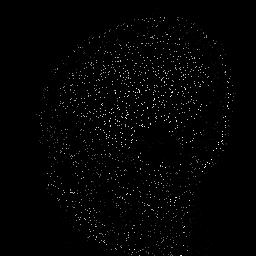}&
\includegraphics[width=12.3mm, height = 12.3mm]{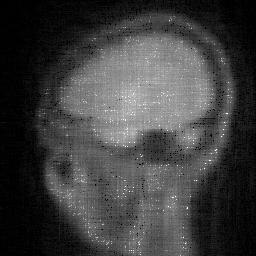}&
\includegraphics[width=12.3mm, height = 12.3mm]{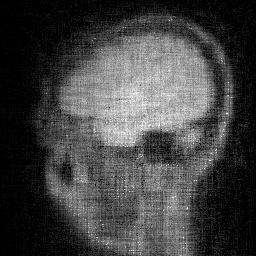}&
\includegraphics[width=12.3mm, height = 12.3mm]{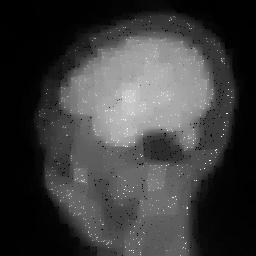}&
\includegraphics[width=12.3mm, height = 12.3mm]{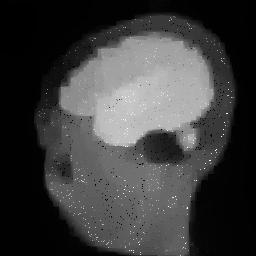}&
\includegraphics[width=12.3mm, height = 12.3mm]{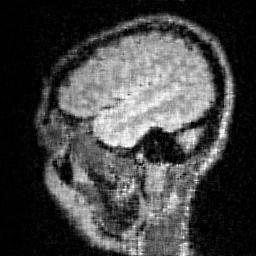}&
\includegraphics[width=12.3mm, height = 12.3mm]{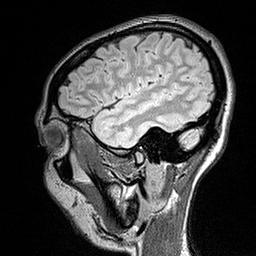}\\
\includegraphics[width=12.3mm, height = 12.3mm]{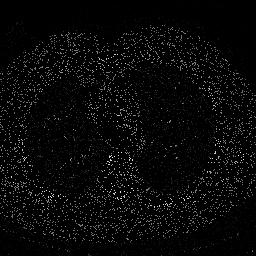}&
\includegraphics[width=12.3mm, height = 12.3mm]{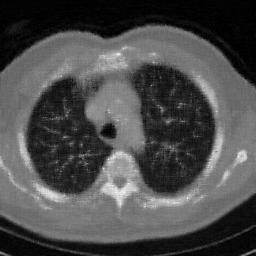}&
\includegraphics[width=12.3mm, height = 12.3mm]{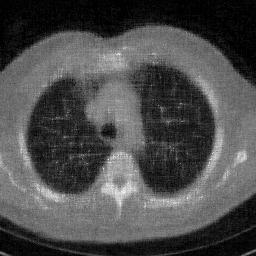}&
\includegraphics[width=12.3mm, height = 12.3mm]{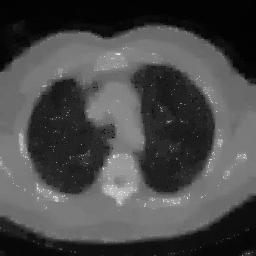}&
\includegraphics[width=12.3mm, height = 12.3mm]{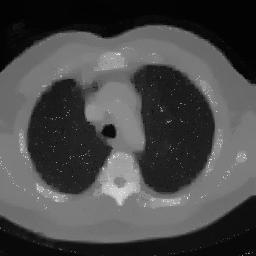}&
\includegraphics[width=12.3mm, height = 12.3mm]{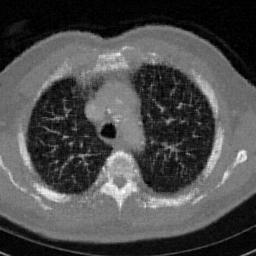}&
\includegraphics[width=12.3mm, height = 12.3mm]{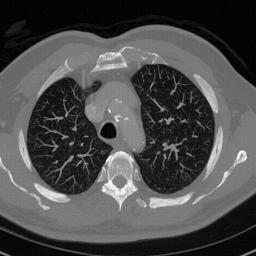}\\
\includegraphics[width=12.3mm, height = 12.3mm]{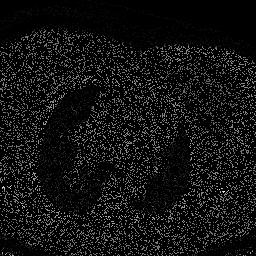}&
\includegraphics[width=12.3mm, height = 12.3mm]{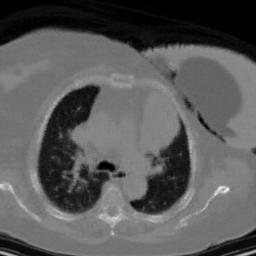}&
\includegraphics[width=12.3mm, height = 12.3mm]{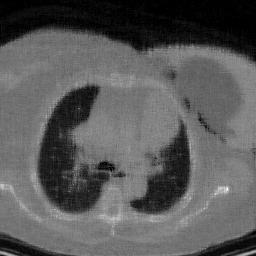}&
\includegraphics[width=12.3mm, height = 12.3mm]{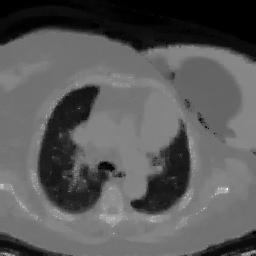}&
\includegraphics[width=12.3mm, height = 12.3mm]{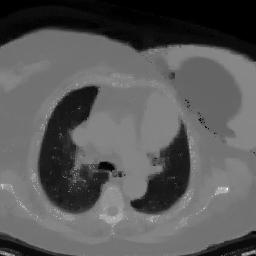}&
\includegraphics[width=12.3mm, height = 12.3mm]{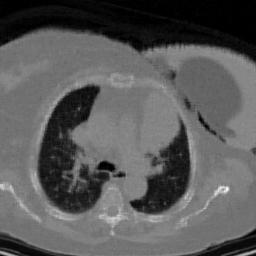}&
\includegraphics[width=12.3mm, height = 12.3mm]{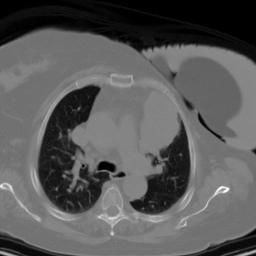}\\
\scriptsize \textbf{Observed} & \scriptsize KBR  & \scriptsize TNN & \scriptsize SNN-TV & \scriptsize TNN-TV & \scriptsize \textbf{t-CTV} & \scriptsize \textbf{Original}\\
\end{tabular}
\vspace{-0.2cm}
\caption{MRI and CT image inpainting results. From top to bottom: SR $=5\%,10\%$ and $20\%$.}\label{fig.13}
\vspace{-0.5cm}
\end{figure}

\begin{figure}[!htbp]
\renewcommand{\arraystretch}{0.5}
\setlength\tabcolsep{0.5pt}
\centering
\begin{tabular}{cccc}
\centering
\tiny 25.47/0.746/148.5 & \tiny 33.34/0.936/60.01 & \tiny \underline{41.13}/\underline{0.977}/\underline{24.56} & \tiny 28.90/0.816/100.1 \\
\includegraphics[width=21.5mm, height = 14mm]{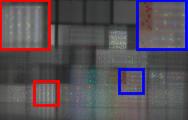}&
\includegraphics[width=21.5mm, height = 14mm]{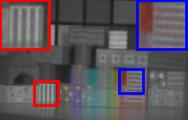}&
\includegraphics[width=21.5mm, height = 14mm]{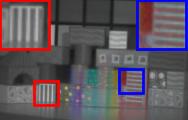}&
\includegraphics[width=21.5mm, height = 14mm]{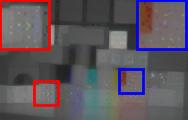}\\
\scriptsize SNN & \scriptsize KBR  & \scriptsize TNN & \scriptsize SNN-TV \\
\tiny 35.69/0.943/45.86 & \tiny 28.26/0.825/107.6 & \tiny \textbf{44.70}/\textbf{0.988}/\textbf{16.23} & \tiny PSNR/SSIM/ERGAS\\
\includegraphics[width=21.5mm, height = 14mm]{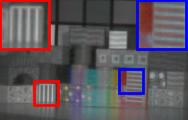}&
\includegraphics[width=21.5mm, height = 14mm]{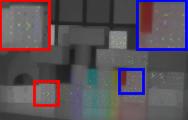}&
\includegraphics[width=21.5mm, height = 14mm]{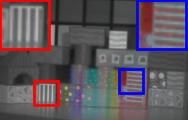}&
\includegraphics[width=21.5mm, height = 14mm]{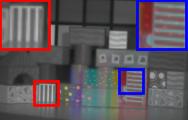}\\
\scriptsize SPC-TV & \scriptsize TNN-TV  & \scriptsize \textbf{t-CTV} & \scriptsize \textbf{Ground truth} \\
\end{tabular}
\vspace{-0.2cm}
\caption{Hyperspectral video inpainting results under SR $=10\%$. The displayed pseudo-color image is the $7$-th frame of the recovered video with band 7-21-28 as R-G-B.}\label{fig.14}
\vspace{-0.5cm}
\end{figure}

\begin{figure*}[!htbp]
\renewcommand{\arraystretch}{0.5}
\setlength\tabcolsep{0.5pt}
\centering
\begin{tabular}{ccccccccc}
\centering
\tiny 5.22/0.037/0.051 & \tiny 15.44/0.329/0.626 & \tiny 14.82/0.323/0.627 & \tiny \underline{26.11}/\underline{0.743}/\underline{0.845} & \tiny 17.76/0.558/0.707 & \tiny 19.68/0.519/0.710 & \tiny 16.99/0.571/0.708 & \tiny \textbf{27.82}/\textbf{0.824}/\textbf{0.875} & \tiny PSNR/SSIM/FSIM\\
\includegraphics[width=19.8mm, height = 19.8mm]{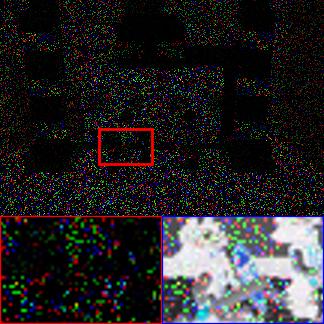}&
\includegraphics[width=19.8mm, height = 19.8mm]{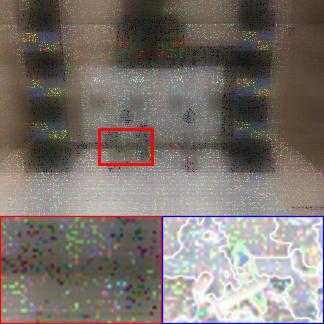}&
\includegraphics[width=19.8mm, height = 19.8mm]{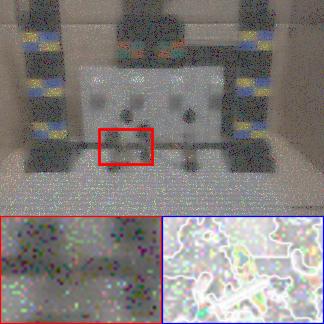}&
\includegraphics[width=19.8mm, height = 19.8mm]{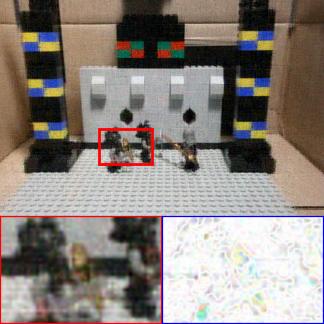}&
\includegraphics[width=19.8mm, height = 19.8mm]{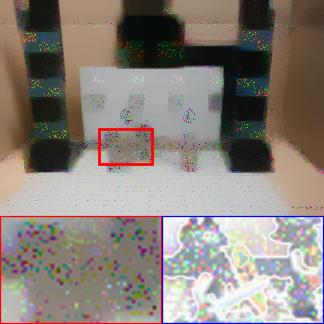}&
\includegraphics[width=19.8mm, height = 19.8mm]{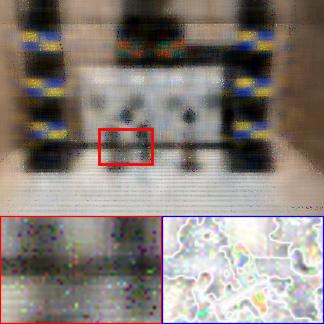}&
\includegraphics[width=19.8mm, height = 19.8mm]{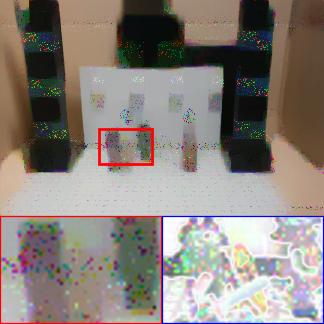}&
\includegraphics[width=19.8mm, height = 19.8mm]{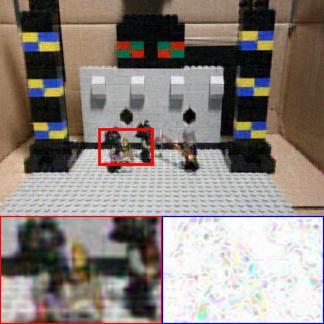}&
\includegraphics[width=19.8mm, height = 19.8mm]{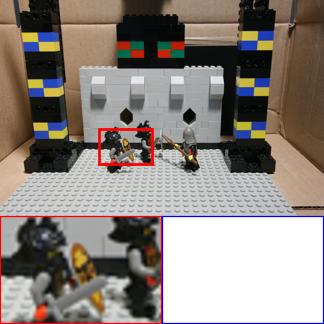}\\
\tiny 6.93/0.128/0.134 & \tiny 22.81/0.785/0.838 & \tiny 24.27/0.755/0.844 & \tiny \underline{34.70}/\underline{0.935}/\underline{0.957} & \tiny 25.41/0.910/0.925 & \tiny 25.37/0.755/0.837 & \tiny 24.83/0.904/0.923 & \tiny \textbf{36.23}/\textbf{0.960}/\textbf{0.971} & \tiny PSNR/SSIM/FSIM\\
\includegraphics[width=19.8mm, height = 19.8mm]{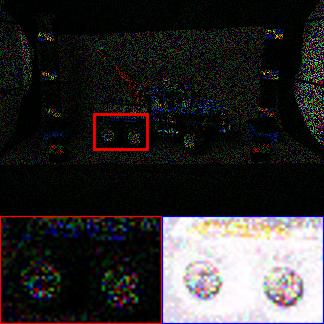}&
\includegraphics[width=19.8mm, height = 19.8mm]{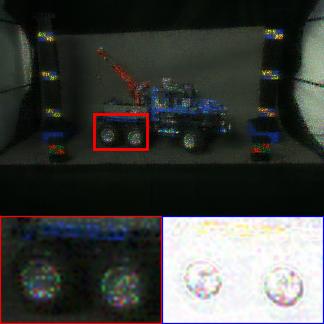}&
\includegraphics[width=19.8mm, height = 19.8mm]{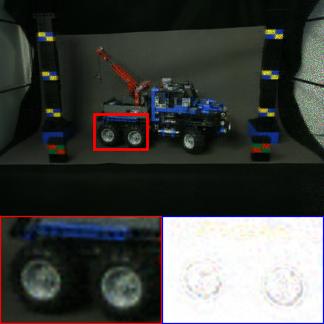}&
\includegraphics[width=19.8mm, height = 19.8mm]{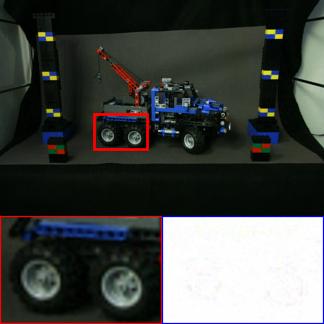}&
\includegraphics[width=19.8mm, height = 19.8mm]{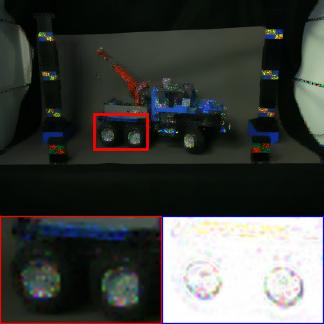}&
\includegraphics[width=19.8mm, height = 19.8mm]{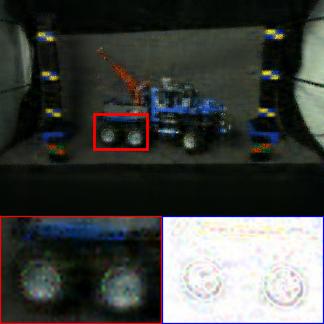}&
\includegraphics[width=19.8mm, height = 19.8mm]{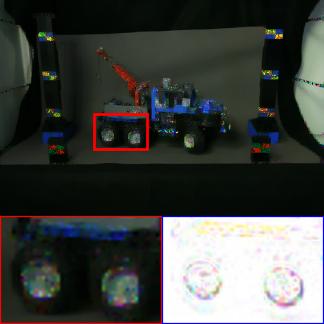}&
\includegraphics[width=19.8mm, height = 19.8mm]{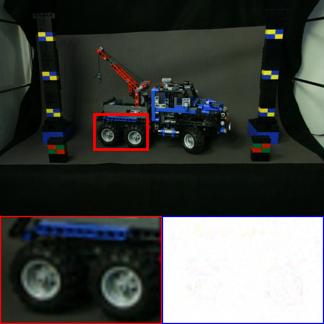}&
\includegraphics[width=19.8mm, height = 19.8mm]{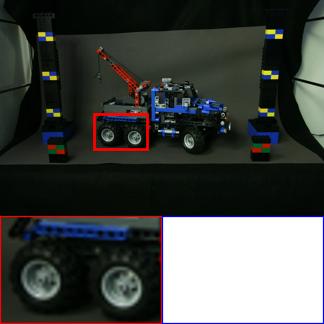}\\
\scriptsize \textbf{Observed} & \scriptsize SNN & \scriptsize KBR & \scriptsize TNN & \scriptsize SNN-TV & \scriptsize SPC-TV & \scriptsize TNN-TV  & \scriptsize \textbf{t-CTV} & \scriptsize \textbf{Ground truth} \\
\end{tabular}
\vspace{-0.2cm}
\caption{Light field image inpainting results. From top to bottom: SR $=10\%$ and $30\%$. The displayed images are the $7$-th band and the $20$-th frame of the recovered LFIs. }\label{fig.15}
\vspace{-0.2cm}
\end{figure*}

\subsection{Applications to Visual Data Denoising}
We then apply the proposed t-CTV-TRPCA method to the denoising task. Similarly, we compare our results with related \textbf{L} and \textbf{L}+\textbf{S} modeling methods, summarized in Table \ref{table.9}. Note that there exist multiple parameters in these competing methods. We choose the suggested parameter setting in corresponding release codes. In comparison, our t-CTV-RPCA method doesn't need to tune parameters where the sole parameter $\lambda$ is set as $1/\sqrt{n_{(1)}\ell}$ given by Theorem \ref{th.7}.

\begin{table}[tp]
\renewcommand{\arraystretch}{1.15}
\setlength\tabcolsep{3.0pt}
\footnotesize
  \caption{
  Categories of related tensor RPCA methods.
  }\label{table.9}
  \setlength{\abovecaptionskip}{5pt}
  \setlength{\belowcaptionskip}{5pt}
  \centering
\begin{tabular}{c|c}
     \Xhline{1pt}
     Types & Methods\\
     \Xhline{1pt}
     \textbf{L}
     & SNN\cite{huang2015provable}, TNN\cite{lu2019tensor},
     KBR\cite{xie2017kronecker}\\
     \hline
     \textbf{L}$\&$\textbf{S}
     & LRTV\cite{he2015total}, LRTDTV\cite{wang2017hyperspectral}, TLR-HTV/TLR-SSTV\cite{chen2018tensor}\\
     \Xhline{1pt}
\end{tabular}
\vspace{-0.2cm}
\end{table}

\begin{table*}[tp]
\renewcommand{\arraystretch}{1.15}
\setlength\tabcolsep{3.0pt}
\footnotesize
  \caption{
   Visual data denoising performances of all competing methods under different levels of noise.
  }\label{table.10}
  \setlength{\abovecaptionskip}{5pt}
  \setlength{\belowcaptionskip}{5pt}
  \centering
\begin{tabular}{l||@{}c|c|c@{}|@{}c|c|c@{}|@{}c|c|c@{}|@{}c|c|c@{}|@{}c|c|c@{}|@{}c|c|c@{}|c}
     \Xhline{1pt}
      Noise & \multicolumn{3}{c|}{S = 0.1} & \multicolumn{3}{c|}{S = 0.2} & \multicolumn{3}{c|}{S = 0.3} & \multicolumn{3}{c|}{S = 0.4} & \multicolumn{3}{c|}{S = 0.1, G = 0.05} & \multicolumn{3}{c|}{S = 0.2, G = 0.05}& \multirow{2}{*}{Time/s}\\
     \cline{2-19}
     Method & PSNR & SSIM & FSIM & PSNR & SSIM & FSIM & PSNR & SSIM & FSIM & PSNR & SSIM & FSIM & PSNR & SSIM & FSIM & PSNR & SSIM & FSIM  & \quad \\
     \Xhline{1pt}
     \multicolumn{19}{c}{Order-$3$ Color Images}\\
     \hline
     \hline
     SNN & 25.65 & 0.826 & 0.878 & 24.68 & 0.789 & 0.858 & 23.69 & 0.746 & 0.837 & \underline{22.64} & \underline{0.691} & \underline{0.813} & 24.39 & 0.708 & 0.840 & 23.60 & \underline{0.669} & \underline{0.823} & 16.60\\
     TNN & 29.85 & \underline{0.923} & 0.944 & \underline{28.16} & \underline{0.871} & \underline{0.922} & \underline{25.92} & \underline{0.748} & \underline{0.877} & 21.97 & 0.479 & 0.767 & 25.92 & 0.672 & 0.857 & \underline{24.30} & 0.588 &0.820 & 9.18\\
     KBR & 25.57 & 0.819 & 0.883 & 24.31 & 0.763 & 0.856 & 22.86 & 0.677 & 0.822 & 21.28 & 0.548 & 0.770 & 24.13 & 0.690 & 0.841 & 23.00 & 0.634 &0.816 & 29.81\\
     LRTV & 28.35 & 0.807 & 0.927 & 24.33 & 0.624 & 0.849 & 21.25 & 0.467 & 0.762 & 18.77 & 0.346 & 0.677 & 26.07 & 0.681 & 0.863 & 23.15 & 0.545 &0.796 & 11.24\\
     LRTDTV & 29.49 & 0.845 & 0.929 & 25.47 & 0.698 & 0.859 & 22.30 & 0.539 &0.775 & 19.65 & 0.398 & 0.692 & \underline{26.78} & \underline{0.729} & \underline{0.875} & 23.97 & 0.607 &0.814 & 11.21\\
     TLR-HTV & \underline{31.64} & 0.893 & \underline{0.949} & 25.79 & 0.660 & 0.861 & 17.82 & 0.309 & 0.659 & 11.70 & 0.111 & 0.469 & 26.14 & 0.659 & 0.854 & 21.85 & 0.472 & 0.758 & 12.62\\
     \textbf{t-CTV} & \textbf{31.65} & \textbf{0.933} & \textbf{0.960} & \textbf{30.56} & \textbf{0.913} & \textbf{0.949} & \textbf{29.36} & \textbf{0.884} & \textbf{0.935} & \textbf{27.97} & \textbf{0.836} & \textbf{0.914} & \textbf{28.20} & \textbf{0.774} & \textbf{0.891} & \textbf{27.25} & \textbf{0.733} & \textbf{0.872} & 18.20\\
     \hline
     \hline
     \multicolumn{19}{c}{Order-$3$ Hyperspectral Images}\\
     \hline
     \hline
     SNN & 42.78 & 0.976 & 0.978 & 40.60 & 0.974 & 0.985 & 37.86 & 0.970 & 0.974 & 34.49 & 0.948 & 0.964 & 31.36 & 0.920 & 0.952 & 30.26 & 0.907 & 0.946 & 80.80\\
     TNN & \underline{44.57} & \underline{0.991} & \underline{0.993} & \underline{42.73} & \underline{0.988} & \underline{0.992} & \underline{40.29} & \underline{0.982} & \underline{0.988} & \underline{36.30} & \underline{0.955} & \underline{0.976} & 29.55 & 0.747 & 0.904 & 28.19 & 0.690 & 0.881 & 77.87\\
     KBR & 36.39 & 0.971 & 0.979 & 35.38 & 0.966 & 0.975 & 33.94 & 0.957 & 0.968 & 32.32 & 0.924 & 0.949 & 32.90 & \underline{0.925} & \underline{0.956} & 32.01 & \underline{0.913} & \underline{0.950} & 161.7\\
     LRTV & 38.01 & 0.978 & 0.988 & 36.06 & 0.962 & 0.979 & 34.10 & 0.933 & 0.964 & 32.15 & 0.890 & 0.941 & 32.22 & 0.898 & 0.940 & 32.20 & 0.879 & 0.929 & 12.89\\
     LRTDTV & 39.63 & 0.978 & 0.990 & 38.75 & 0.975 & 0.988 & 37.45 & 0.968 & 0.984 & 35.66 & 0.953 & 0.975 & \textbf{35.20} & \textbf{0.930} & \textbf{0.963} & \textbf{34.30} & \textbf{0.918} & \textbf{0.956} & 53.22\\
     TLR-SSTV & 37.68 & 0.973 & 0.983 & 36.09 & 0.962 & 0.976 & 34.42 & 0.946 & 0.966 & 32.75 & 0.923 & 0.951 & 31.72 & 0.855 & 0.928 & 31.04 & 0.841 & 0.917 & 23.94\\
     \textbf{t-CTV} & \textbf{46.32} & \textbf{0.993} & \textbf{0.996} & \textbf{44.93} & \textbf{0.992} & \textbf{0.995} & \textbf{43.27} & \textbf{0.989} & \textbf{0.993} & \textbf{41.37} & \textbf{0.985} & \textbf{0.991} & \underline{33.01} & 0.871 & 0.940 & \underline{32.28} &0.843 & 0.927 & 118.0\\
     \hline
     \hline
     \multicolumn{19}{c}{Order-$4$ Color Videos}\\
     \hline
     \hline
     SNN & 21.50 & 0.696 & 0.797 & 20.70 & 0.666 & 0.779 & 19.82 & 0.634 & 0.762 & 18.82 & 0.600 & 0.744 & 20.60 & 0.642 & 0.772 & 19.91 & 0.618 & 0.759 & 278.3\\
     TNN & \underline{37.31} & \underline{0.982} & \underline{0.986} & \underline{35.49} & \underline{0.971} & \underline{0.980} & \underline{33.29} & \underline{0.939} & \underline{0.965} & \underline{28.94} & 0.789 & \underline{0.907} & 28.00 & 0.694 & 0.868 & 26.35 & 0.624 & 0.834 & 188.2\\
     KBR & 29.52 & 0.906 & 0.938 & 28.72 & 0.896 & 0.931 & 27.71 & 0.878 & 0.920 & 25.99 & \underline{0.836} & 0.894 & 28.35 & \textbf{0.844} & \textbf{0.914} & 27.45 & \textbf{0.825} & \textbf{0.904} & 267.5\\
     LRTV & 31.03 & 0.851 & 0.943 & 23.64 & 0.543 & 0.804 & 19.69 & 0.362 & 0.681 & 16.94 & 0.252 & 0.582 & 26.48 & 0.649 & 0.835 & 22.23 & 0.462 & 0.735 & 64.30\\
     LRTDTV & 32.91 & 0.905 & 0.958 & 28.80 & 0.806 & 0.909 & 25.32 & 0.673 & 0.839 & 22.24 & 0.510 & 0.742 & 29.37 & 0.806 & 0.908 & 26.55 & 0.702 & 0.857 & 174.1\\
     TLR-SSTV & 36.23 & 0.969 & 0.981 & 34.25 & 0.949 & 0.971 & 31.45 & 0.894 & 0.948 & 27.54 & 0.758 & 0.889 & \underline{29.27} & 0.772 & 0.900 & \underline{27.75} & 0.715 & 0.873 & 177.3\\
     \textbf{t-CTV} & \textbf{39.73} & \textbf{0.987} & \textbf{0.990} & \textbf{38.17} & \textbf{0.982} & \textbf{0.987} & \textbf{36.62} & \textbf{0.974} & \textbf{0.983} & \textbf{34.85} & \textbf{0.959} & \textbf{0.974} & \textbf{30.90} & \underline{0.817} & \underline{0.910} & \textbf{29.79} & \underline{0.779} & \underline{0.891} & 260.4\\
     \hline
     \hline
     \Xhline{1pt}
\end{tabular}
\vspace{-0.3cm}
\end{table*}

\begin{figure*}[tp]
\renewcommand{\arraystretch}{0.5}
\setlength\tabcolsep{0.5pt}
\centering
\vspace{-0.2cm}
\begin{tabular}{ccccccccc}
\centering
\tiny 12.13/0.144/0.583 & \tiny 24.51/0.844/0.855/ & \tiny 27.54/\underline{0.891}/\underline{0.926} & \tiny 24.73/0.844/0.873 & \tiny 25.57/0.773/0.870 & \tiny 27.25/0.785/0.902 & \tiny \underline{28.14}/0.813/0.923 & \tiny \textbf{29.38}/\textbf{0.944}/\textbf{0.951} & \tiny PSNR/SSIM/FSIM\\
\includegraphics[width=19.8mm, height = 19.8mm]{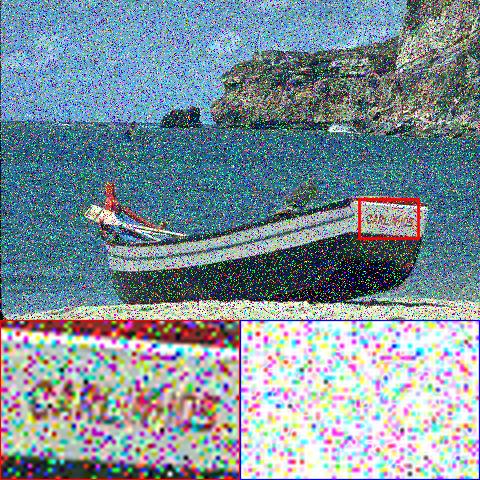}&
\includegraphics[width=19.8mm, height = 19.8mm]{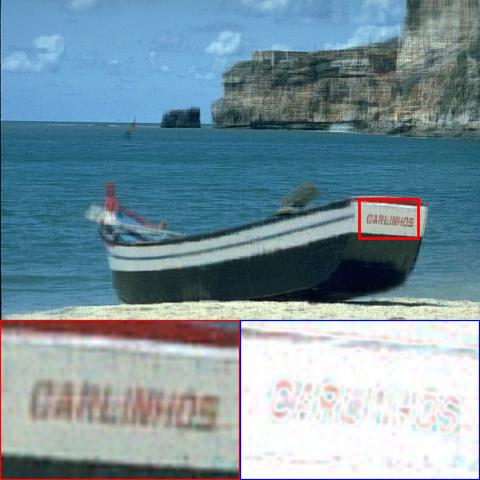}&
\includegraphics[width=19.8mm, height = 19.8mm]{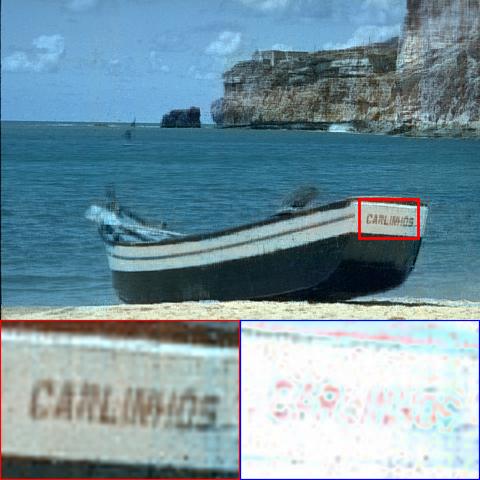}&
\includegraphics[width=19.8mm, height = 19.8mm]{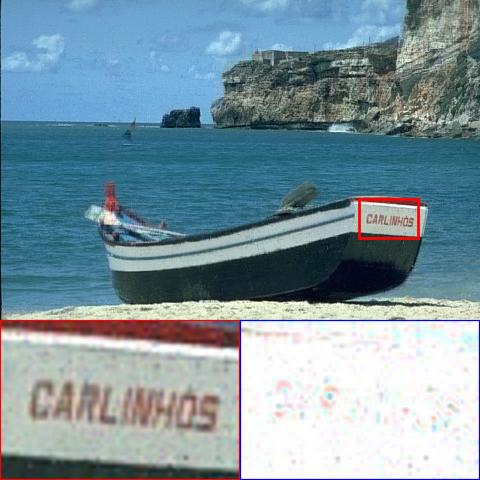}&
\includegraphics[width=19.8mm, height = 19.8mm]{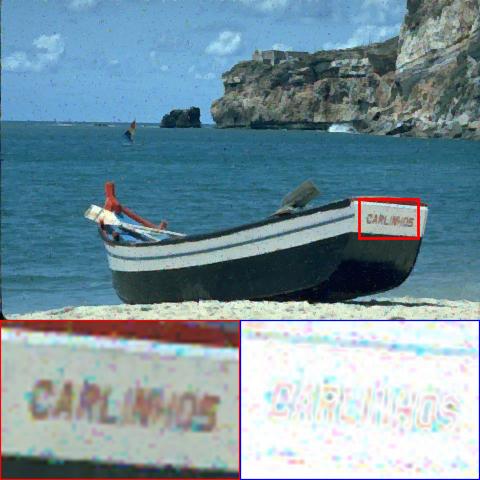}&
\includegraphics[width=19.8mm, height = 19.8mm]{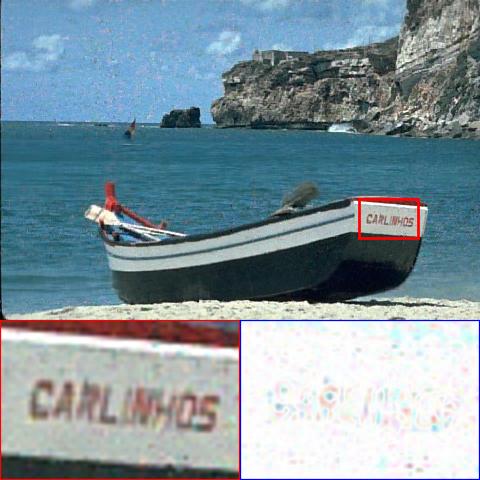}&
\includegraphics[width=19.8mm, height = 19.8mm]{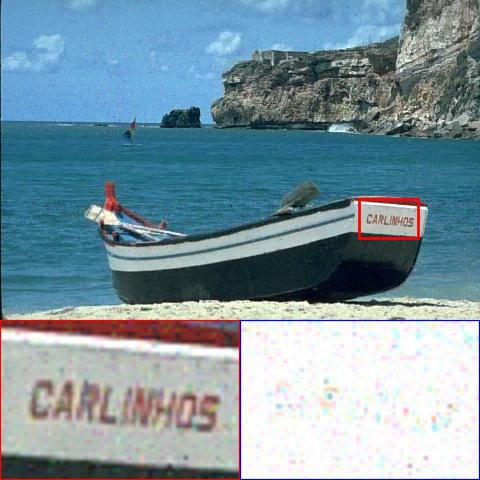}&
\includegraphics[width=19.8mm, height = 19.8mm]{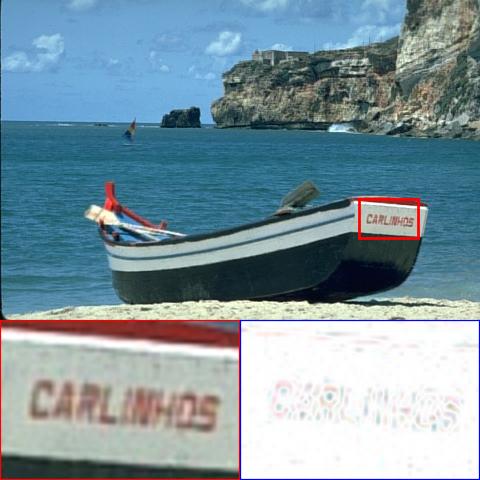}&
\includegraphics[width=19.8mm, height = 19.8mm]{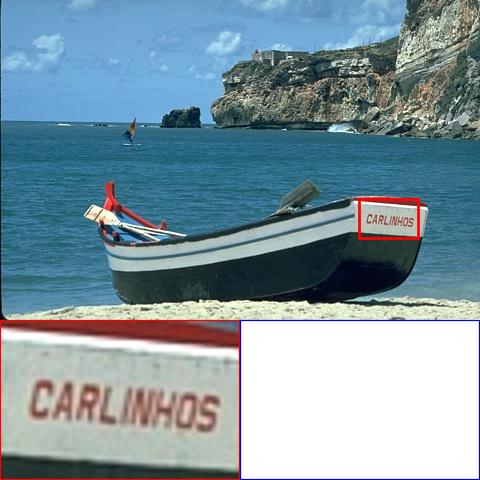}\\
\tiny 8.57/0.087/0.434 & \tiny 21.10/0.601/0.792 & \tiny 18.23/0.348/0.687 & \tiny 18.99/0.457/0.714 & \tiny 20.62/0.554/0.761 & \tiny \underline{23.90}/\underline{0.734}/\underline{0.844} & \tiny 21.20/0.512/0.776 & \tiny \textbf{26.92}/\textbf{0.878}/\textbf{0.922} & \tiny PSNR/SSIM/FSIM\\
\includegraphics[width=19.8mm, height = 19.8mm]{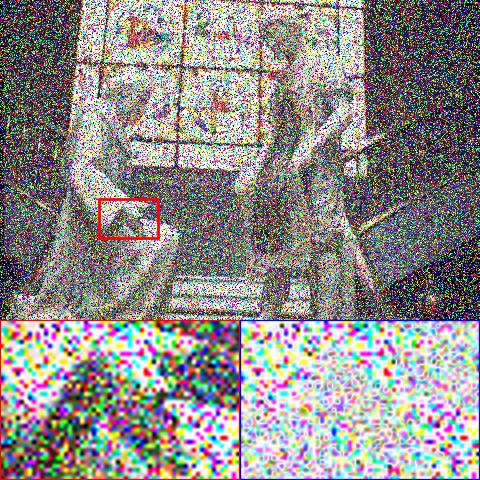}&
\includegraphics[width=19.8mm, height = 19.8mm]{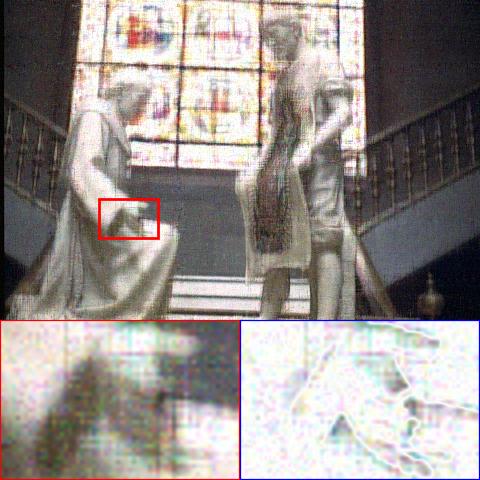}&
\includegraphics[width=19.8mm, height = 19.8mm]{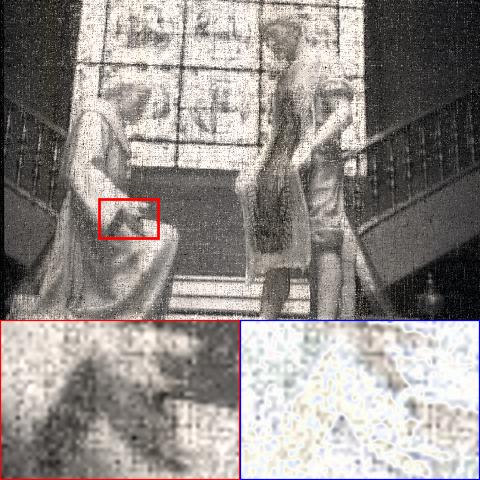}&
\includegraphics[width=19.8mm, height = 19.8mm]{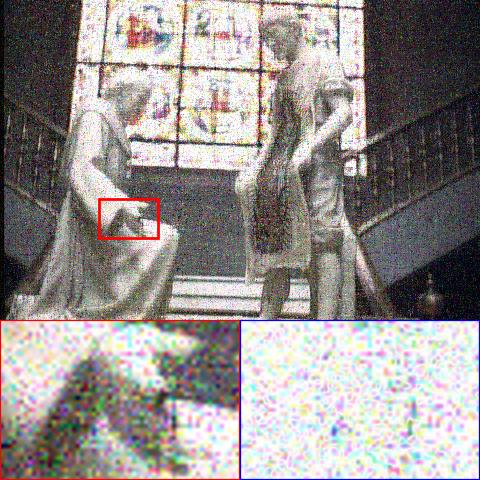}&
\includegraphics[width=19.8mm, height = 19.8mm]{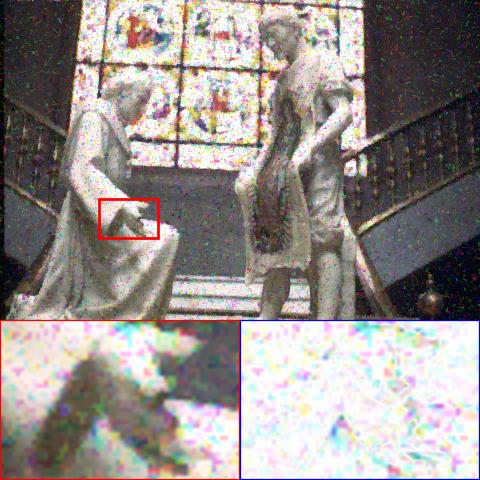}&
\includegraphics[width=19.8mm, height = 19.8mm]{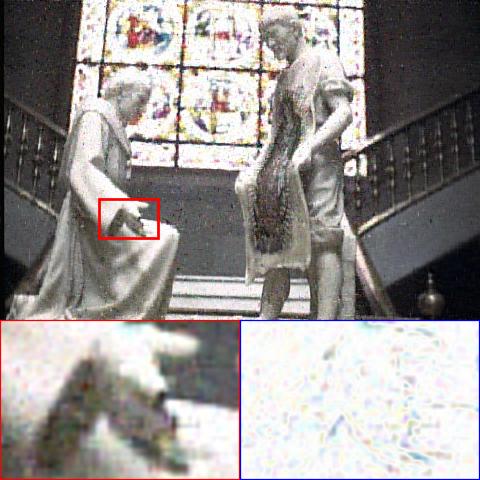}&
\includegraphics[width=19.8mm, height = 19.8mm]{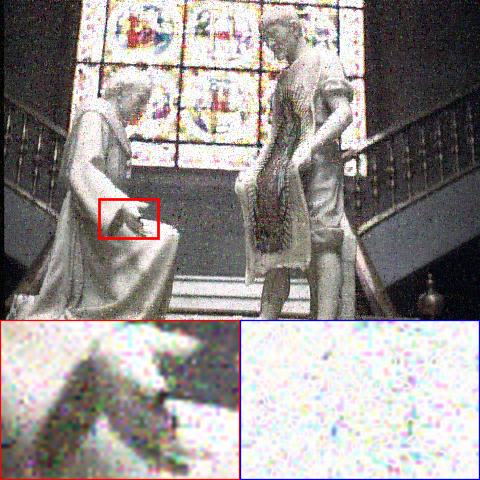}&
\includegraphics[width=19.8mm, height = 19.8mm]{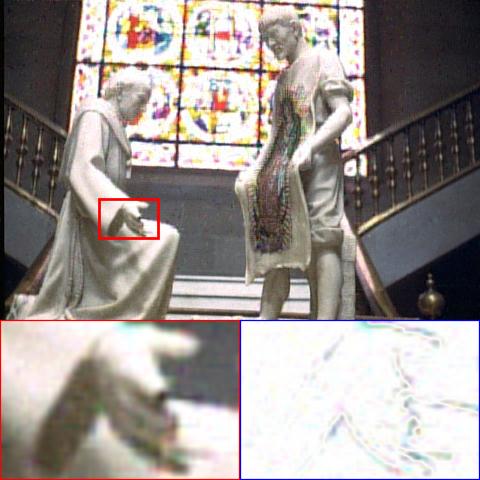}&
\includegraphics[width=19.8mm, height = 19.8mm]{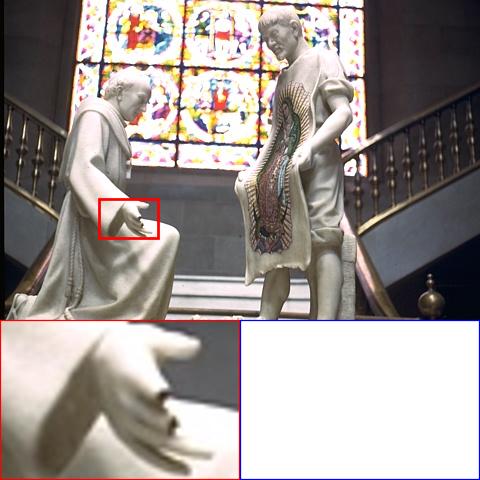}\\
\tiny 11.56/0.171/0.555 & \tiny 22.97/0.615/0.819 & \tiny 22.89/0.582/0.814 & \tiny 21.38/0.642/0.823 & \tiny \underline{24.10}/0.697/0.854 & \tiny 23.39/\underline{0.721}/\underline{0.858} & \tiny 23.27/0.604/0.823 & \tiny \textbf{26.02}/\textbf{0.739}/\textbf{0.870} & \tiny PSNR/SSIM/FSIM\\
\includegraphics[width=19.8mm, height = 19.8mm]{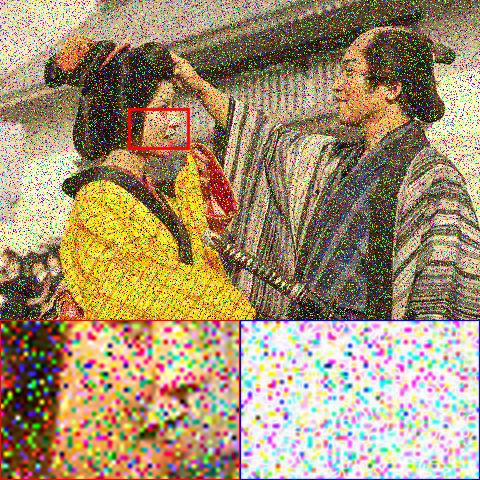}&
\includegraphics[width=19.8mm, height = 19.8mm]{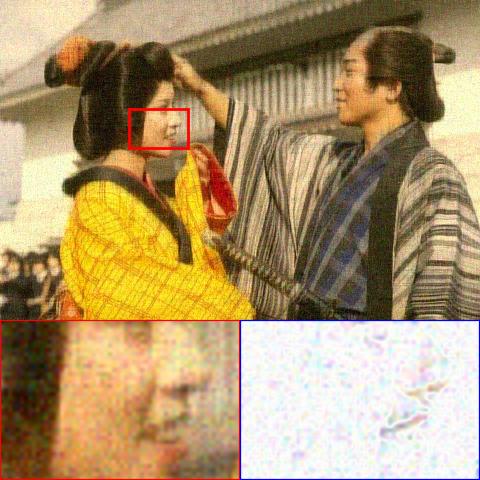}&
\includegraphics[width=19.8mm, height = 19.8mm]{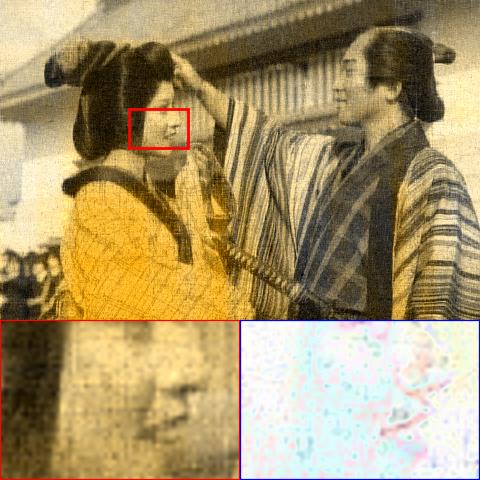}&
\includegraphics[width=19.8mm, height = 19.8mm]{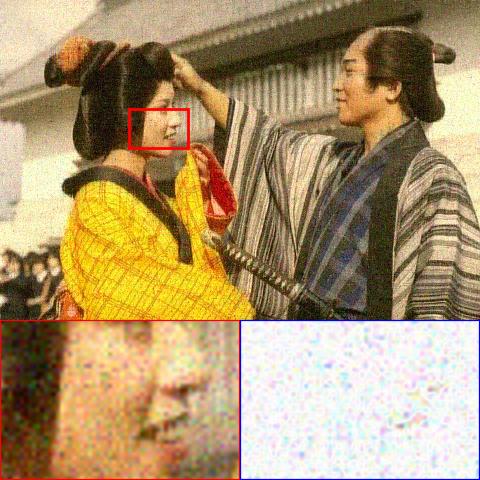}&
\includegraphics[width=19.8mm, height = 19.8mm]{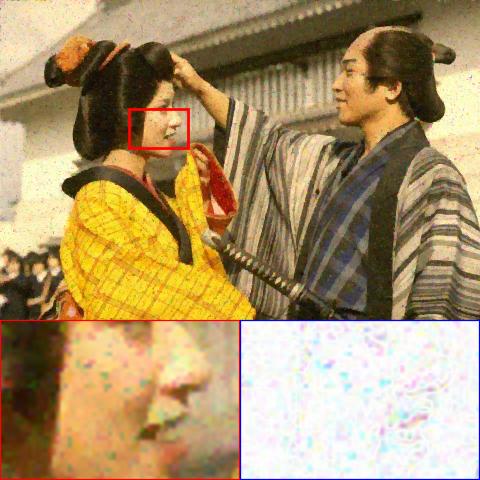}&
\includegraphics[width=19.8mm, height = 19.8mm]{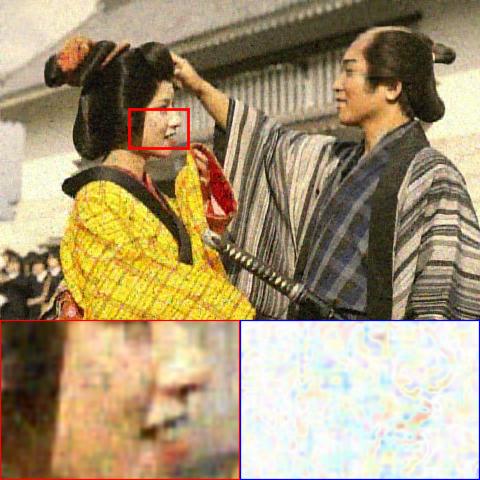}&
\includegraphics[width=19.8mm, height = 19.8mm]{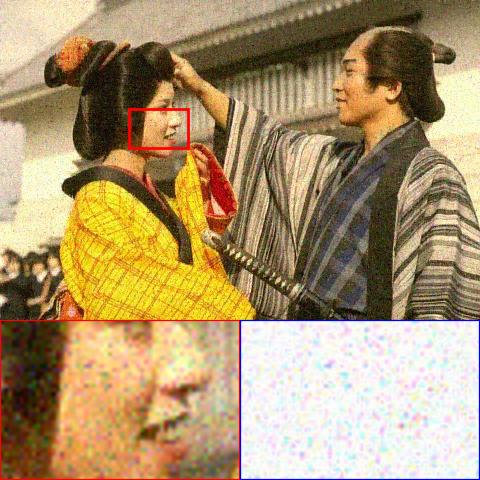}&
\includegraphics[width=19.8mm, height = 19.8mm]{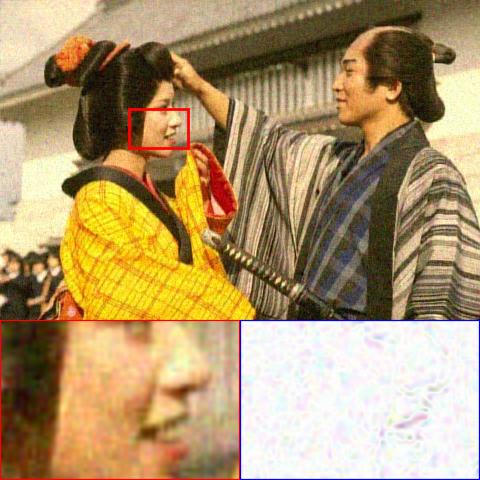}&
\includegraphics[width=19.8mm, height = 19.8mm]{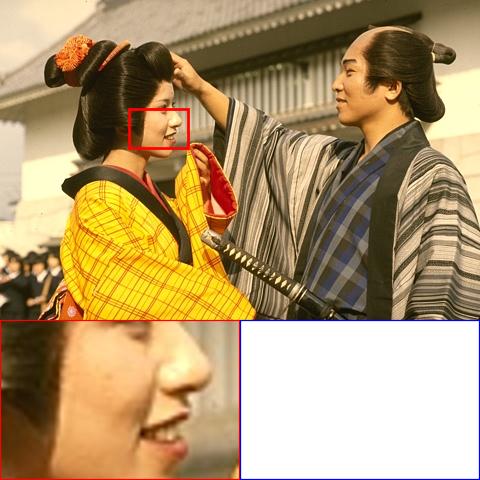}\\
\scriptsize \textbf{Noisy} & \scriptsize SNN & \scriptsize KBR & \scriptsize TNN & \scriptsize LRTV & \scriptsize LRTDTV & \scriptsize TLR-HTV  & \scriptsize \textbf{t-CTV} & \scriptsize \textbf{Ground truth} \\
\end{tabular}
\vspace{-0.2cm}
\caption{Color image denoising results by all competing methods. From top to bottom: cases containing sparse noise of noise percentages $0.2$ and $0.4$, images containing mixed noise with sparse noise with noise percentage $0.2$ and Gaussian noise with zero-mean and standard deviation $0.05$.}\label{fig.16}
\end{figure*}

\begin{figure}[tp]
\renewcommand{\arraystretch}{0.5}
\setlength\tabcolsep{0.5pt}
\centering
\vspace{-0.2cm}
\begin{tabular}{ccc}
\centering
\tiny 8.10/0.017/1872 & \tiny 35.87/0.933/89.24 & \tiny 36.31/0.931/77.79 \\
\includegraphics[width=29mm, height = 20mm]{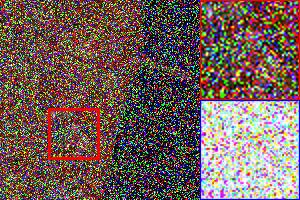}&
\includegraphics[width=29mm, height = 20mm]{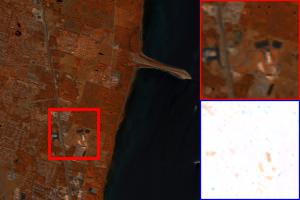}&
\includegraphics[width=29mm, height = 20mm]{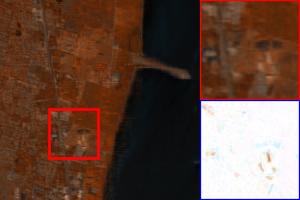}\\
\scriptsize \textbf{Noisy} & \scriptsize SNN & \scriptsize KBR \\
\tiny 33.10/0.913/102.9 & \tiny 32.93/0.846/109.5 & \tiny \underline{36.87}/\underline{0.943}/\underline{64.49}\\
\includegraphics[width=29mm, height = 20mm]{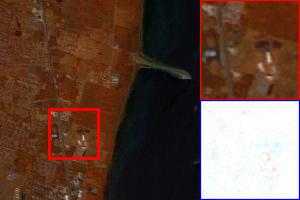}&
\includegraphics[width=29mm, height = 20mm]{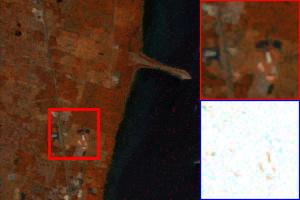}&
\includegraphics[width=29mm, height = 20mm]{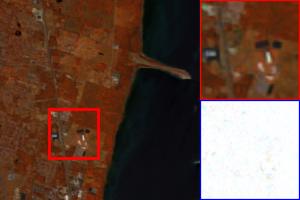}\\\
\scriptsize TNN & \scriptsize LRTV & \scriptsize LRTDTV \\
\tiny 33.87/0.909/92.33 & \tiny \textbf{42.38}/\textbf{0.980}/\textbf{44.94} & \tiny PSNR/SSIM/ERGAS\\
\includegraphics[width=29mm, height = 20mm]{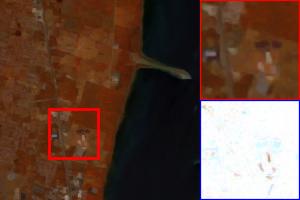}&
\includegraphics[width=29mm, height = 20mm]{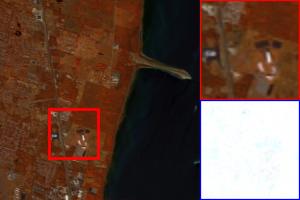}&
\includegraphics[width=29mm, height = 20mm]{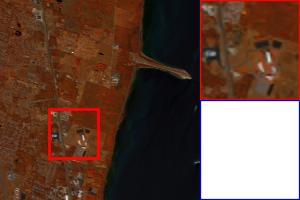}\\
\scriptsize TLR-SSTV  & \scriptsize \textbf{t-CTV}  & \scriptsize \textbf{Ground truth}\\
\end{tabular}
\vspace{-0.2cm}
\caption{HSI ``KSC" denoising results on a typical HSI containing sparse noise with noise percentage $0.4$.}\label{fig.18}
\vspace{-0.3cm}
\end{figure}

\begin{figure*}[tp]
\renewcommand{\arraystretch}{0.4}
\setlength\tabcolsep{0.5pt}
\centering
\begin{tabular}{ccccccccc}
\centering
\tiny 10.49/0.098/0.455 & \tiny 21.86/0.770/0.825 & \tiny 32.96/0.952/0.967 & \tiny \underline{38.14}/\underline{0.964}/\underline{0.987} & \tiny 29.85/0.809/0.912 & \tiny 31.03/0.869/0.947 & \tiny 32.66/0.936/0.964 & \tiny \textbf{41.76}/\textbf{0.986}/\textbf{0.993} & \tiny PSNR/SSIM/FSIM\\
\includegraphics[width=19.8mm, height = 16mm]{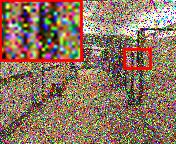}&
\includegraphics[width=19.8mm, height = 16mm]{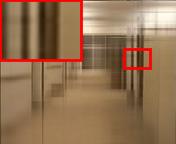}&
\includegraphics[width=19.8mm, height = 16mm]{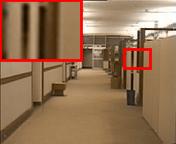}&
\includegraphics[width=19.8mm, height = 16mm]{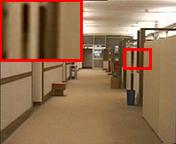}&
\includegraphics[width=19.8mm, height = 16mm]{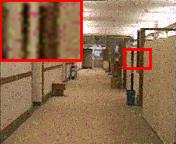}&
\includegraphics[width=19.8mm, height = 16mm]{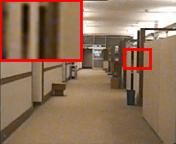}&
\includegraphics[width=19.8mm, height = 16mm]{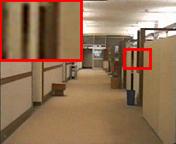}&
\includegraphics[width=19.8mm, height = 16mm]{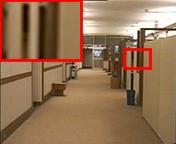}&
\includegraphics[width=19.8mm, height = 16mm]{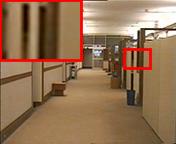}\\
\tiny 8.67/0.078/0.399 & \tiny 18.59/0.683/0.771 & \tiny 26.49/0.902/0.922 & \tiny \underline{32.70}/\underline{0.924}/\underline{0.965} & \tiny 24.75/0.735/0.850 & \tiny 26.92/0.882/0.919 & \tiny 28.94/0.914/0.943 & \tiny \textbf{36.92}/\textbf{0.987}/\textbf{0.990} & \tiny PSNR/SSIM/FSIM\\
\includegraphics[width=19.8mm, height = 16mm]{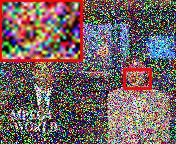}&
\includegraphics[width=19.8mm, height = 16mm]{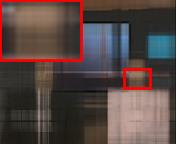}&
\includegraphics[width=19.8mm, height = 16mm]{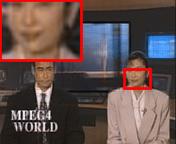}&
\includegraphics[width=19.8mm, height = 16mm]{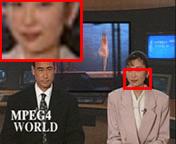}&
\includegraphics[width=19.8mm, height = 16mm]{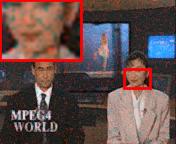}&
\includegraphics[width=19.8mm, height = 16mm]{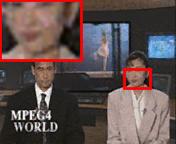}&
\includegraphics[width=19.8mm, height = 16mm]{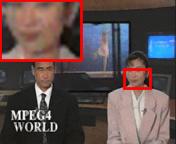}&
\includegraphics[width=19.8mm, height = 16mm]{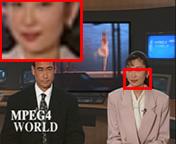}&
\includegraphics[width=19.8mm, height = 16mm]{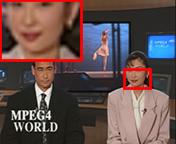}\\
\scriptsize \textbf{Noisy} & \scriptsize SNN & \scriptsize KBR & \scriptsize TNN & \scriptsize LRTV & \scriptsize LRTDTV & \scriptsize TLR-SSTV  & \scriptsize \textbf{t-CTV}  & \scriptsize \textbf{Ground truth}\\
\end{tabular}
\vspace{-0.2cm}
\caption{Color video denoising results by all competing methods. From top to bottom: cases containing sparse noise with noise percentage $0.3, 0.4$.
}\label{fig.19}
\vspace{-0.3cm}
\end{figure*}

We use the aforementioned databases of color images, HSIs and color videos for test. In Table \ref{table.10}, we list the denoising results on \textit{sparse salt and pepper noise} (S) with different percentages, and their mixed noise with a bit \textit{Gaussian noise} (G) with zero-mean and standard deviation $0.05$. From this table, we can find easily that our t-CTV based method achieves the best performance in all sparse noise cases. Especially, when the percentage of sparse noise is getting increased, our leading advantages are clearer. For example, our method is almost 5dB in PSNR higher than the second-best method when $40\%$ percentage pixels are corrupted by sparse noise.

As for mixed noise cases, our method also shows competitive performance. It should be indicated that our t-CTV-TRPCA model is only a generic model imposing $L_1$ norm on noise elements succeeding from conventional RPCA models, specifically suitable for sparse noises in nature. The purpose of considering some Gaussian noise is to test our model's stability. For these related methods including LRTV, LRTDTV and TLR-HTV/SSTV, their models include extra $L_2$-norm element, and thus is formulated appropriate for mixed Gaussian noises. From the results of Table \ref{table.10}, it can be observed that our method can overmatch other competitors even without such explicit Gaussian noise item, revealing the powerfulness of such regularization term on expressing natural visual tensors with \textbf{L}+\textbf{S} priors.

We further display several typical recovered examples for visualization comparison in Fig.\;\ref{fig.16}, Fig.\;\ref{fig.18} and Fig.\;\ref{fig.19}. From these figures, it can be easily seen that our t-CTV based method achieves better noise removal performance. Specifically, the denoising results by our method maintain clearer outlines and more faithful local details compared with other competitors, validating its fine capability in extracting the natural images with intrinsic \textbf{L}+\textbf{S} structures.

\section{Conclusion}\label{sec.8}

In this study, we have proposed a new regularization term, called  t-CTV, to encode both \textbf{L}+\textbf{S} priors, commonly possessed by natural visual tensor data, into a unique concise form. By formulating the regularization in two typical tensor recovery tasks, including TC and TRPCA, we can prove their exact recovery guarantee theoretically for both models. This should be the first theoretical exact recovery result along this \textbf{L}+\textbf{S} prior modeling research, revealing its general reliability and potential usefulness of the proposed regularizer. We also prove that such \textbf{L}+\textbf{S} modeling manner is with a lower sampling complexity bound beyond conventional \textbf{L} and/or \textbf{S} prior models, which has been comprehensively validated by our experiments in different types of practical visual tensor data. Especially, even under the extremely low sampling rage $0.5\%$, our method can still get an acceptable recovery effect, when all other competing methods totally fail in such cases. We'll try to testify more tensor recovery tasks, like outlier detection and background extraction, by combining this regularizer into their corresponding models to further validate its effect, and attempt to explore more targeted theory for further revealing its working insights in our future investigations. Furthermore, it is also meaningful to extend the theoretical and algorithmic explorations of this work to other ones with low-rank regularizers under different matrix/tensor transformations in our future research.


%



\ifCLASSOPTIONcaptionsoff
  \newpage
\fi



%
%
%

\bibliographystyle{IEEEtran}
\bibliography{IEEEabrv,mybibfile}

%

\begin{IEEEbiography}[{\includegraphics[width=1in,height=1.25in,clip,keepaspectratio]{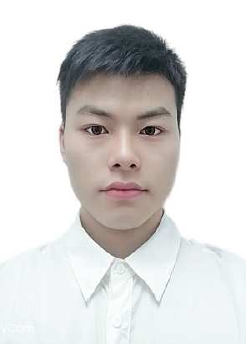}}]{Hailin Wang}
is currently pursuing a Ph.D. degree with the School of Mathematics and Statistics, Xi'an Jiaotong University, Xi'an, China. He received the B.Sc and M.Sc degree in statistic from Southwest University, Chongqing, China, in 2019, 2021, respectively. His research interests include compressed sensing and low-rank matrix/tensor analysis.
\end{IEEEbiography}

\begin{IEEEbiography}[{\includegraphics[width=1in,height=1.25in,clip,keepaspectratio]{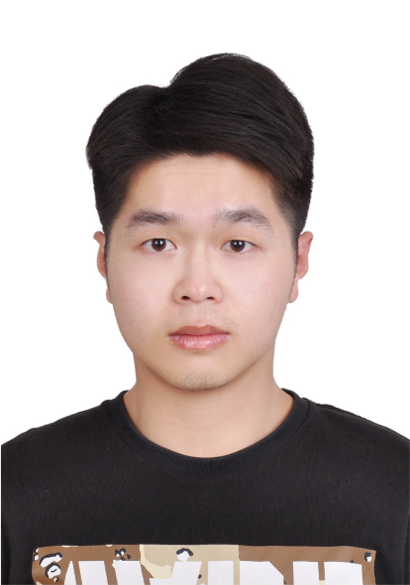}}]{Jiangjun Peng}
received the B.Sc degree in mathematical from Northwest University, Xi'an, China, in 2015 and the M.Sc degree in mathematics from Xi'an Jiaotong University, Xi'an, China, in 2018. He was a researcher in Tencent from 2018 to 2020. He is currently pursuing a Ph.D. degree with the School of Mathematics and Statistics, Xi'an Jiaotong University, Xi'an. His current research interests include low-rank matrix/tensor analysis, theoretical machine learning and model-driven deep learning.
\end{IEEEbiography}

\begin{IEEEbiography}[{\includegraphics[width=1in,height=1.25in,clip,keepaspectratio]{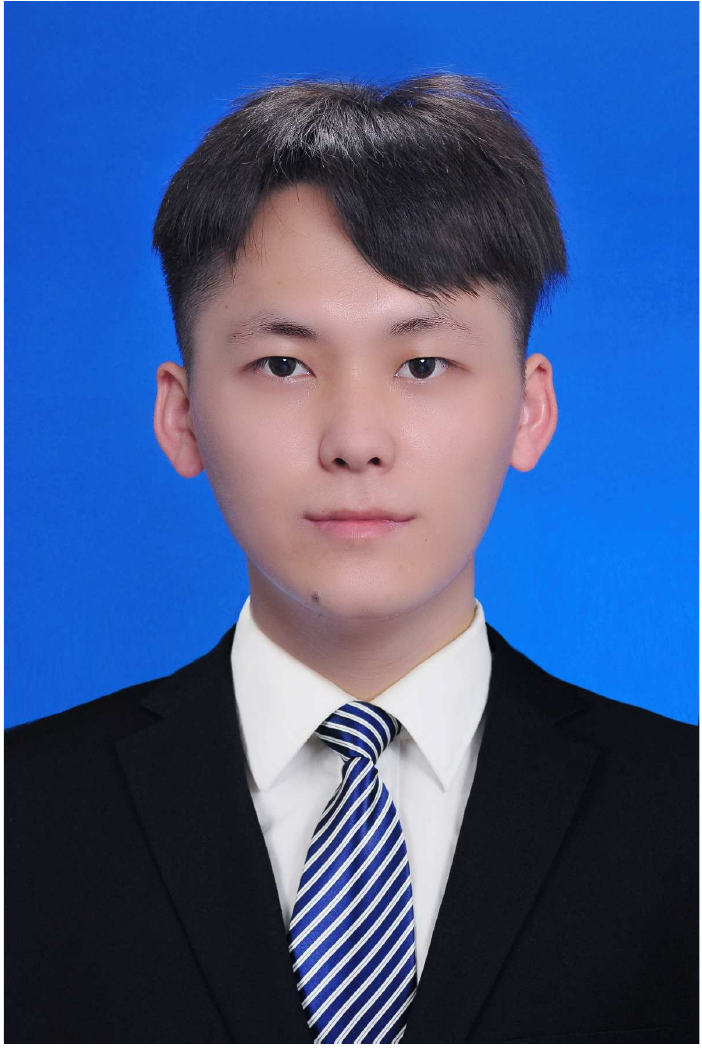}}]{Wenjin Qin}
is currently pursuing a Ph.D. degree with the School of Mathematics and Statistics, Southwest University, Chongqing, China. He received the B.Sc degree in applied statistics 
from Hechi University, Guangxi, China, in 2019, and the M.Sc degree in statistic from Southwest University, Chongqing, China, in 2022. His research interests include deep learning, compressed sensing, low-rank matrix recovery and tensor sparsity.
\end{IEEEbiography}

\begin{IEEEbiography}[{\includegraphics[width=1in,height=1.25in,clip,keepaspectratio]{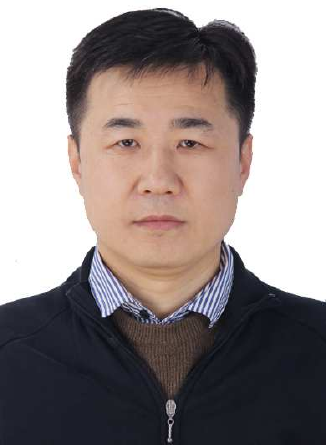}}]{Jianjun Wang}
(Member, IEEE)
received the B.Sc degree in mathematical education from Ningxia University in
2000 and the M.Sc degree in fundamental mathematics in 2003 from Ningxia University, China,
and Ph.D degree in applied mathematics from the Institute for Information and System Science, Xi'an Jiaotong University in 2006.
He is currently a professor in the School of Mathematics and Statistics, Southwest University, Chongqing, China.
His research focus on machine learning, data mining, neural networks and sparse learning.
He is a member of the IEEE.
\end{IEEEbiography}

\begin{IEEEbiography}[{\includegraphics[width=1in,height=1.25in,clip,keepaspectratio]{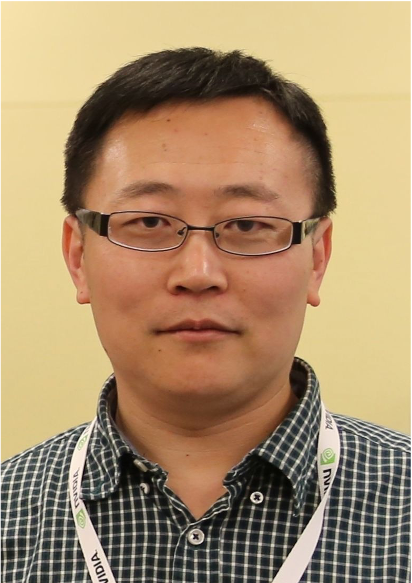}}]{Deyu Meng}
(Member, IEEE) received the B.Sc., M.Sc., and Ph.D. degrees from Xi'an Jiaotong University, Xi'an, China, in 2001, 2004, and 2008, respectively. From 2012 to 2014, he took his two-year sabbatical leave at Carnegie Mellon University, Pittsburgh, PA, USA. He is a Professor with the School of Mathematics and Statistics, Xi'an Jiaotong University, and an Adjunct Professor with the Faculty
of Information Technology, Macau University of Science and Technology, Taipa, Macau, China. His research interests include model-based deep learning, variational networks, and meta learning.
\end{IEEEbiography}




\end{document}